\documentclass{article}

     \PassOptionsToPackage{numbers}{natbib}

     \usepackage[preprint]{neurips_2020}

\usepackage[utf8]{inputenc} 
\usepackage[T1]{fontenc}    
\usepackage{hyperref}       
\usepackage{url}            
\usepackage{booktabs}       
\usepackage{amsfonts}       
\usepackage{nicefrac}       
\usepackage{microtype}      
\usepackage[eulergreek]{sansmath}
\usepackage{amsmath, amssymb, amsthm, enumerate, bm, capt-of,mathtools}
\usepackage[capitalise]{cleveref}
\usepackage{algorithm}  
\usepackage{algorithmicx}  
\usepackage[noend]{algpseudocode}  
\usepackage{siunitx}
\usepackage{placeins}

\usepackage{tikz, tikzscale,pgfplots}
\usetikzlibrary{arrows,shapes,backgrounds,patterns,fadings,matrix,arrows,calc,
	intersections,decorations.markings,
	positioning,arrows.meta}
\pgfplotsset{width = 0.48\textwidth, compat = 1.13, height=5.8cm, grid=major, 
	ticklabel style = {font=\sansmath\sffamily\scriptsize},
	zlabel style = {font = \sf\footnotesize},
	legend style = {font=\sf\scriptsize}, legend cell align = left, 
	title style={yshift=-7pt, font =\sf\footnotesize} }

\DeclareMathOperator{\med}{med}

\DeclareMathOperator*{\argmax}{arg\,max}

\makeatletter
\newcommand{\subalign}[1]{%
	\vcenter{%
		\Let@ \restore@math@cr \default@tag
		\baselineskip\fontdimen10 \scriptfont\tw@
		\advance\baselineskip\fontdimen12 \scriptfont\tw@
		\lineskip\thr@@\fontdimen8 \scriptfont\thr@@
		\lineskiplimit\lineskip
		\ialign{\hfil$\m@th\scriptstyle##$&$\m@th\scriptstyle{}##$\crcr
			#1\crcr
		}%
	}
}
\makeatother

\title{Real-Time Regression with Dividing Local\\ Gaussian Processes}

\author{%
   Armin Lederer \\
   Technical University of Munich \\
   80333 Munich, Germany \\
   \texttt{armin.lederer@tum.de} \\
   \And
   Alejandro Jos\'e Ord\'o\~nez Conejo\\
   Tecnol\'ogico de Costa Rica \\
   30101 Cartago, Costa Rica \\
   \texttt{ajoseoc@gmail.com} \\
   \And
   Korbinian Maier \\
   FRANKA EMIKA GmbH \\
   80797 Munich, Germany \\
   \texttt{korbinian.maier@franka.de} \\
   \And
   Wenxin Xiao \\
   Peking University \\
   100871 Beijing, China \\
   \texttt{xiao.wenxin@foxmail.com} \\
   \And
   Jonas Umlauft \\
   Technical University of Munich \\
   80333 Munich, Germany \\
   \texttt{jonas.umlauft@tum.de} \\
   \And
   Sandra Hirche \\
   Technical University of Munich \\
   80333 Munich, Germany \\
   \texttt{hirche@tum.de}
}

\begin{document}

\maketitle

\begin{abstract}
The increased demand for online prediction 
and the growing availability of large data sets drives the need for 
computationally efficient models. While exact Gaussian process regression shows various 
favorable theoretical properties (uncertainty estimate, unlimited expressive 
power), the poor scaling with respect to the training set size prohibits its application in big data regimes in real-time. 
Therefore, 
this paper proposes dividing local Gaussian processes, which are a novel, 
computationally efficient modeling approach based on Gaussian process regression. Due to 
an iterative, data-driven division of the input space, 
they achieve a sublinear computational complexity in the 
total number of training points in practice, while providing excellent predictive distributions. 
A numerical evaluation on real-world data 
sets shows their advantages over other state-of-the-art methods in terms of accuracy as well as
prediction and update speed.\looseness=-1
\end{abstract}

\section{Introduction}

Recent technological trends enable ever growing storage capacities and declining
costs for sensor hardware, resulting in a significant increase of available data, which 
allows for in-depth analysis and precise modeling of 
various technical systems. As the amount of data increases, it becomes 
inevitable to design methods which scale well to large data sets. 
Especially in control applications, scalability is 
additionally constrained by high update rates and real-time requirements on the prediction. 
These applications include the control of autonomous cars, unmanned aerial 
vehicles~\cite{Andersson2017}, robotic 
manipulators~\cite{Nguyen-Tuong2010},  combustion 
engines~\cite{Lee2017b,Hafner2000} and many others, 
where 
updates rates in the magnitude of~$\SI{e2}{\hertz}$ 
to~$\SI{e4}{\hertz}$ are required. 
In case of predictive control schemes, where possible future trajectories are 
inferred and evaluated, multiple predictions are made for a single control 
command, requiring prediction rates, which are orders of magnitudes 
higher~\cite{Kong2015}.

Parametric learning approaches transfer all information from the data into 
parameters and therefore typically have constant computational complexity of predictions and updates
independent of the number of data points \cite{Nguyen-Tuong2011}. While this generally allows real-time regression, 
it comes at the cost of limited model flexibility and requires expert knowledge. In contrast, non-parametric learning methods such as Gaussian processes 
(GPs) can provide unlimited expressive power~\cite{Rasmussen2006},
but grow in complexity with the number of training points. Therefore, they suffer
from increasing prediction and update times. In order to mitigate this issue, 
several approaches have focused on reducing the complexity of posterior mean computations, 
such as, e.g., inducing point methods \cite{Nguyen-Tuong2010, Schreiter2016} or deterministic 
training conditional approximations \cite{Huber2014, Bijl2017}. 
However, allowing model updates in real-time has
attracted far less attention, and the quality of predictive distributions in online regression has 
barely been addressed.\looseness=-1

The main contribution of this paper is a novel, computationally efficient, GP-based
method for real-time predictions and model updates, called dividing local Gaussian process (DLGP). 
Our approach is based on an online, data-driven division of the 
input space in order to build local GP models. The division of the data during training, 
and the combination of local models for prediction are both
performed with a sublinear computational complexity in practice. 
In a numerical study, we compare our approach with existing 
state-of-the-art 
modeling techniques on real-world data sets with respect to training and 
prediction time as well as prediction error and quality of the predictive distributions. 
The paper is structured as follows:
In \cref{sec:Background}, we discuss the related work in depth and briefly 
introduce GP regression. \Cref{sec:DLGP}  
presents the proposed DLGP model and in \cref{sec:Eval} the 
numerical results are compared. \looseness=-1

\setlength{\textfloatsep}{8pt}
\setlength{\floatsep}{8pt}
\setlength{\abovedisplayskip}{4pt}
\setlength{\belowdisplayskip}{4pt}

\section{Background}
\label{sec:Background}
\subsection{Related Work}

Real-time regression for data-driven models is a challenge that has originally been considered in the context of 
robotics \cite{Nguyen-Tuong2011}, and recently it gains increasing attention in 
control theory~\cite{Umlauft2020} 
and reinforcement learning~\cite{Koller2018}. In early work, this problem has been approached by 
adapting existing
non-parametric methods, such as support vector regression \cite{Ma2003}, to enable real-time learning. 
Due to computational limitations, this approach is difficult to apply in practice. This
lead to the development of dedicated real-time learning methods based on linear regression, which
are typically referred to as locally weighted projection regression (LWPR) 
\cite{Vijayakumar2000, Vijayakumar2005}. LWPR owes its success to reasonable regression quality, 
high update and prediction rates, and its straightforward usage with publicly available software \cite{Klanke2008}. Due to
these reasons, it is commonly applied in robotics \cite{Fagogenis2016, Grandia2018} and control 
\cite{Gao2018} up to today. However, the regression quality of LWPR significantly depends 
on design parameters, which require careful tuning in practice, and therefore contradicts the online
learning paradigm. In order to overcome this issue, Bayesian 
adaptations of LWPR have been proposed, which base on variational inference \cite{Meier2014b, Meier2014}, 
and novel methods using mean field variational Bayesian approximate inference have been developed~\cite{Luts2014}. 
Despite of their theoretical advantages, these methods can barely be found in applications.\looseness=-1

In recent years, Gaussian process approximations have experienced
a great success in online regression research due to the straightforward update rules 
originating from Bayes theory. Due to the high computational complexity of exact updates, 
a large variety of strategies has been developed to mitigate this issue. Originally 
developed for large data sets \cite{Snelson2007}, sparse methods have been among the 
first to be applied to real-time learning \cite{Csato2002}. Some methods use deterministic 
training conditional approximations with active subsets of the training data, which are determined using different metrics 
\cite{Csato2002, Nguyen-Tuong2010, Schreiter2016, Koppel}, while others achieve sparsity 
by using inducing points chosen online \cite{Huber2014, Bijl2017, Le2017}, or through 
compactified covariance functions resulting in sparse cholesky factors 
\cite{Ranganathan2008, Ranganathan2011}. Alternative approaches rely on variational 
inference based on the free energy approximation \cite{Bui2017} or solve variational 
inference in the reproducing kernel Hilbert space via stochastic mirror descent \cite{Cheng2016} to deal with streaming 
data online. Although these approaches can achieve a small prediction error with sufficiently many 
inducing points, their updates for new training data are rather slow. When choosing a small number 
of inducing points to reduce the update time, the prediction performance of these methods typically
drops significantly. This is due to the fact that a lot of information from the training data cannot be included into the model, 
which significantly limits the asymptotic performance of such approaches. Using 
explicit features and parametric learning can mitigate these problems. In particular, when offline data 
is available, the features can be fitted to this data using neural networks \cite{Harrison2018} 
or least squares \cite{Camoriano2016}. Conversely, without any offline data, random trigonometric features 
with strong theoretical guarantees can be easily determined using Bochner's theorem 
\cite{Rahimi2008}, leading to the method's name sparse spectrum GP \cite{Gijsberts2013}. 
These methods are known to suffer from overfitting \cite{Gal2015} and their posterior 
variances are overconfident \cite{Liu2018b}. Furthermore, the posterior mean and variance will 
be periodic functions, such that the variance might collapse far from any training samples 
\cite{vanderWilk2018} leading to overconfident predictions. In contrast to sparse likelihood approximations 
and kernel approximations, Local GP methods \cite{Nguyen-Tuong2009a} follow a different idea: By 
combining the predictions of multiple GP models, the overall computational complexity is reduced. 
The usage of 
multiple GPs is typically exploited by spatially separating them, which is usually
referred to as local GPs \cite{Nguyen-Tuong2008, Nguyen-Tuong2009}, although random training
point assignment to models also allows real-time regression \cite{Xiao2013}. A similar effect
can be achieved using a single GP with moving window to select training samples~\cite{Meier2016}. 
Since the complexity of exact inference in local models still leads to
quickly growing computational complexity, this is typically avoided by using sparse local GPs, 
inheriting many of the disadvantages. Therefore, we pursue a different 
approach by dividing data sets and retraining the local models when they become too large. Thereby, 
we reduce computational complexity of predictions and updates, without suffering from the 
disadvantages of sparse GP approximations.
\looseness=-1

\subsection{Gaussian Processes Regression}

A Gaussian process~$\mathcal{GP}$ is the generalization of a Gaussian distribution, and bases on the 
assumption that any finite collection of random variables~$y_i\in\mathbb{R}$
follows a joint Gaussian distribution. This joint Gaussian distribution is uniquely defined by the
prior mean, which is frequently set to~$0$, and a covariance function~$k:\mathbb{R}^d\times\mathbb{R}^d\rightarrow\mathbb{R}$
\cite{Rasmussen2006}. The observations~$y_i$ can be considered as measurements of a sample 
function~$f:\mathbb{R}^d\rightarrow\mathbb{R}$ of the GP distribution and are typically perturbed
by zero mean Gaussian noise with variance~$\sigma_n^2\in\mathbb{R}_{+,0}$. We concatenate~$N$ 
input training samples~$\bm{x}_i$ and output observations~$y_i$ into a matrix~$\bm{X}$ and a vector 
$\bm{y}$, which represent the training data set~$\mathbb{D}$. Furthermore, we define the elements 
of the GP kernel matrix 
$\bm{K}(\bm{X},\bm{X})$ as~$K_{ij}=k(\bm{x}_i,\bm{x}_j)$ and define the elements of the 
kernel vector~$\bm{k}(\bm{X},\bm{x})$ accordingly. Based on these definitions, we can represent
the GP model efficiently as
\begin{align}
\bm{L}&=\mathrm{cholesky}(\bm{K}(\bm{X},\bm{X})+\sigma_n^2\bm{I})&
\bm{\alpha}&=\bm{L}^T\setminus(\bm{L}\setminus\bm{y})
\label{eq:GP model}
\end{align}
where "$\setminus$" denotes the forward and backward substitution, respectively, such 
that~$\mathcal{O}(N^3)$ and~$\mathcal{O}(N^2)$ operations are required for the computation of $\bm{L}$ and
$\bm{\alpha}$~\cite{Rasmussen2006}, respectively. Then, the posterior GP distribution~$p_{\mathcal{GP}}(f(\bm{x})|\bm{x},\mathbb{D})=\mathcal{N}(\mu(\bm{x}),\sigma^2(\bm{x}))$ 
at a test point~$\bm{x}$ can be computed using
\begin{align}
\mu(\bm{x})&=\bm{k}(\bm{x},\bm{X})\bm{\alpha}&
\bm{v}&=\bm{L}\setminus \bm{k}(\bm{X},\bm{x})&
\sigma^2(\bm{x})&=k(\bm{x},\bm{x})-\bm{v}^T\bm{v},
\label{eq:GP pred}
\end{align}
which requires~$\mathcal{O}(N)$ and~$\mathcal{O}(N^2)$ calculations
for the posterior mean and variance, respectively \cite{Rasmussen2006}.\looseness=-1

\section{Dividing Local Gaussian Processes}
\label{sec:DLGP}
While existing local GP approaches for real-time learning base on the principle that 
all local models have the same spatial extension in the input domain, our proposed 
DLGP approach follows a different paradigm. Starting from a single, global model, local models
are iteratively generated by dividing existing models. This is efficiently performed by sampling the data set, 
to which  each training sample is assigned, from localizing random distributions. We explain this 
iterative binary tree construction using random data assignment 
in detail in \cref{subsec:random trans}. 
In \cref{subsec:pred}, the combination of tree structure and localizing probability functions is exploited to 
derive analytical predictive distributions of the DLGP model.\looseness=-1

\subsection{Binary Tree Construction Using Probabilistic Training Data Assignment}
\label{subsec:random trans}

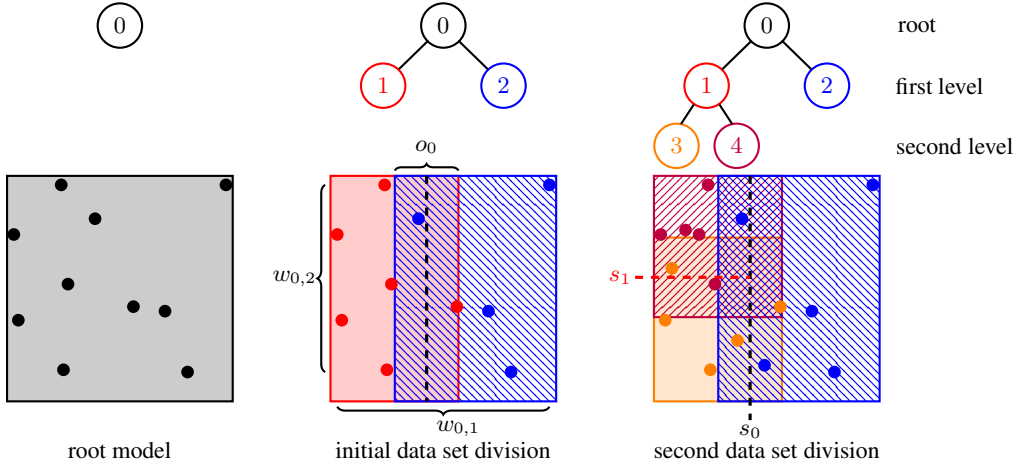
\begin{figure}
	\vspace{1.5mm}
	\tikzstyle{block}=[draw, inner sep=2, outer sep=0, minimum width=2em, 
	rounded corners, thick,  align=center]
	\centering  
	\begin{tikzpicture}[node distance=2cm, 	auto,>=latex',
	every text node part/.style={align=center},
	thick,scale=1, every node/.style={scale=1,font=\small}]
	\def\dis{1.3}
	\def\width{3}
	\def\rad{0.07cm}
	
	\node at (3*\width+2*\dis+0.5,\width+2) {root};	
	\node at (3*\width+2*\dis+0.8,\width+1.2) {first level};	
	\node at (3*\width+2*\dis+1,\width+0.4) {second level};	

	\node[circle,draw] at (\width/2,\width+2) {$0$};	
	\node at (\width/2,-0.65) {root model};	
	
	\filldraw[fill=black!20,draw=black] (0,0) rectangle (\width,\width);
	\filldraw[fill=black,draw=black] (0.05*\width,0.36*\width) circle (\rad);
	\filldraw[fill=black,draw=black] (0.03*\width,0.74*\width) circle (\rad);
	\filldraw[fill=black,draw=black] (0.27*\width,0.52*\width) circle (\rad);
	\filldraw[fill=black,draw=black] (0.56*\width,0.42*\width) circle (\rad);
	\filldraw[fill=black,draw=black] (0.97*\width,0.96*\width) circle (\rad);
	\filldraw[fill=black,draw=black] (0.39*\width,0.81*\width) circle (\rad);
	\filldraw[fill=black,draw=black] (0.24*\width,0.96*\width) circle (\rad);
	\filldraw[fill=black,draw=black] (0.70*\width,0.40*\width) circle (\rad);
	\filldraw[fill=black,draw=black] (0.80*\width,0.13*\width) circle (\rad);
	\filldraw[fill=black,draw=black] (0.25*\width,0.14*\width) circle (\rad);

	\node [circle,draw] at (3*\width/2+\dis,\width+2) {$0$};	
	\node [circle,draw,red] at (3*\width/2+1*\dis-0.8,\width+1.2) {$1$};
	\node [circle,draw,blue] at (3*\width/2+1*\dis+0.8,\width+1.2) {$2$};
	\draw (3*\width/2+\dis-0.2,\width+2-0.2) -- (3*\width/2+1*\dis-0.8+0.21,\width+1.2+0.22);
	\draw (3*\width/2+\dis+0.2,\width+2-0.2) -- (3*\width/2+1*\dis+0.8-0.21,\width+1.2+0.22);
	\node at (3*\width/2+\dis,-0.65) {initial data set division};

	\filldraw[fill=red!20,draw=red] (0+\width+1*\dis,0) rectangle (1.567*\width+1*\dis,\width);
	\fill[pattern=north west lines,pattern color=blue,draw=blue] (1.285*\width+1*\dis,0) rectangle (2*\width+1*\dis,\width);
	
	\filldraw[fill=red,draw=red] (\width+1*\dis+0.05*\width,0.36*\width) circle (\rad);
	\filldraw[fill=red,draw=red] (\width+1*\dis+0.03*\width,0.74*\width) circle (\rad);
	\filldraw[fill=red,draw=red] (\width+1*\dis+0.25*\width,0.14*\width) circle (\rad);
	\filldraw[fill=red,draw=red] (\width+1*\dis+0.56*\width,0.42*\width) circle (\rad);	
	\filldraw[fill=red,draw=red] (\width+1*\dis+0.27*\width,0.52*\width) circle (\rad);
	\filldraw[fill=red,draw=red] (\width+1*\dis+0.24*\width,0.96*\width) circle (\rad);
	
	\filldraw[fill=blue,draw=blue] (\width+1*\dis+0.39*\width,0.81*\width) circle (\rad);
	\filldraw[fill=blue,draw=blue] (\width+1*\dis+0.97*\width,0.96*\width) circle (\rad);		
	\filldraw[fill=blue,draw=blue] (\width+1*\dis+0.70*\width,0.40*\width) circle (\rad);
	\filldraw[fill=blue,draw=blue] (\width+1*\dis+0.80*\width,0.13*\width) circle (\rad);
	
	\draw[dashed, black,very thick] (1.426*\width+\dis,\width) -- (1.426*\width+\dis,-0.0);

	\draw [decorate, decoration={brace,amplitude=3pt,mirror,raise=2pt}, yshift=0pt]
	(1.03*\width+\dis,0) -- (1.97*\width+\dis,0) node [black,midway,xshift=0.2cm,yshift=-0.6cm] {$w_{0,1}$};
	\draw [decorate, decoration={brace,amplitude=3pt,mirror,raise=2pt}, yshift=0pt,black]
	(1*\width+\dis,0.96*\width) -- (1*\width+\dis,0.13*\width) node[black,midway,xshift=-0.9cm,yshift=-0.0cm,align=center] {$w_{0,2}$};
	\draw [decorate, decoration={brace,amplitude=2.5pt,mirror,raise=2pt}, yshift=0pt,black]
	(1.567*\width+\dis,\width) -- (1.285*\width+\dis,\width) node [black,midway,xshift=0.0cm,yshift=0.6cm] {\textcolor{black}{
			$o_{0}$}};

	\node [circle,draw] at (5*\width/2+2*\dis,\width+2) {$0$};
	\node [circle,draw,red] at (5*\width/2+2*\dis-0.8,\width+1.2) {$1$};
	\node [circle,draw,blue] at (5*\width/2+2*\dis+0.8,\width+1.2) {$2$};
	\node [circle,draw,orange] at (5*\width/2+2*\dis-1.2,\width+0.4) {$3$};
	\node [circle,draw,purple] at (5*\width/2+2*\dis-0.4,\width+0.4) {$4$};
	\draw (5*\width/2+2*\dis-0.2,\width+2-0.2) -- (5*\width/2+2*\dis-0.8+0.21,\width+1.2+0.22);
	\draw (5*\width/2+2*\dis+0.2,\width+2-0.2) -- (5*\width/2+2*\dis+0.8-0.21,\width+1.2+0.22);
	\draw (5*\width/2+2*\dis-0.8-0.17,\width+1.2-0.25) -- (5*\width/2+2*\dis-1.2+0.06,\width+0.4+0.29);
	\draw (5*\width/2+2*\dis-0.8+0.17,\width+1.2-0.25) -- (5*\width/2+2*\dis-0.4-0.06,\width+0.4+0.29);
	\node at (5*\width/2+2*\dis,-0.65) {second data set division};
	
	\filldraw[fill=orange!20,draw=orange] (0+2*\width+2*\dis,0) rectangle (2.567*\width+2*\dis,0.7265*\width);
	\fill[pattern color=purple,draw=purple,pattern=north east lines] (0+2*\width+2*\dis,0.3735*\width) rectangle (2.567*\width+2*\dis,\width);
	\fill[pattern=north west lines,pattern color=blue,draw=blue] (2.285*\width+2*\dis,0) rectangle (3*\width+2*\dis,\width);
	\filldraw[fill=blue,draw=blue] (2*\width+2*\dis+0.39*\width,0.81*\width) circle (\rad);
	\filldraw[fill=blue,draw=blue] (2*\width+2*\dis+0.97*\width,0.96*\width) circle (\rad);		
	\filldraw[fill=blue,draw=blue] (2*\width+2*\dis+0.70*\width,0.40*\width) circle (\rad);
	\filldraw[fill=blue,draw=blue] (2*\width+2*\dis+0.80*\width,0.13*\width) circle (\rad);
	\filldraw[fill=blue,draw=blue] (2*\width+2*\dis+0.49*\width,0.16*\width) circle (\rad);
	
	\filldraw[fill=purple,draw=purple] (2*\width+2*\dis+0.24*\width,0.96*\width) circle (\rad);
	\filldraw[fill=purple,draw=purple] (2*\width+2*\dis+0.03*\width,0.74*\width) circle (\rad);
	\filldraw[fill=purple,draw=purple] (2*\width+2*\dis+0.14*\width,0.76*\width) circle (\rad);
	\filldraw[fill=purple,draw=purple] (2*\width+2*\dis+0.20*\width,0.74*\width) circle (\rad);
	
	\filldraw[fill=orange,draw=orange] (2*\width+2*\dis+0.05*\width,0.36*\width) circle (\rad);
	\filldraw[fill=orange,draw=orange] (2*\width+2*\dis+0.25*\width,0.14*\width) circle (\rad);
	\filldraw[fill=orange,draw=orange] (2*\width+2*\dis+0.37*\width,0.27*\width) circle (\rad);
	\filldraw[fill=orange,draw=orange] (2*\width+2*\dis+0.08*\width,0.59*\width) circle (\rad);
	\filldraw[fill=orange,draw=orange] (2*\width+2*\dis+0.56*\width,0.42*\width) circle (\rad);	
	\filldraw[fill=purple,draw=purple] (2*\width+2*\dis+0.27*\width,0.52*\width) circle (\rad);

	\draw[dashed, black,very thick] (2.426*\width+2*\dis,\width) -- (2.426*\width+2*\dis,-0.25);	
	\draw[dashed, red,very thick] (2.*\width+2*\dis-0.25,0.55*\width) -- (2.426*\width+2*\dis,0.55*\width);
	
	\node at (2.426*\width+2*\dis,-0.4) {$s_0$};
	\node[red] at (2.*\width+2*\dis-0.45,0.55*\width) {$s_1$};
	\end{tikzpicture}\vspace{-0.3cm}
	\caption{Iterative model tree construction and corresponding layout of the input space: 
		active regions and training samples belonging to the same node are 
		depicted in the same color.\looseness=-1
		}
	\label{fig:tree}
\end{figure}

Since we consider the problem of real-time regression, we have to deal with streaming data, i.e., sequentially
arriving data samples. 
Therefore, we iteratively construct a model, starting 
with a single data set~$\mathbb{D}_0=\emptyset$. This data set constitutes the root node~$0$ 
of a binary tree~$T_{\mathrm{DLGP}}$, as depicted in \cref{fig:tree}. The incoming training data is added 
to the data set~$\mathbb{D}_0$, and each new data point can be 
efficiently included into the GP model \eqref{eq:GP model} using rank one updates, which exhibit 
quadratic complexity~\cite{Nguyen-Tuong2009}. When the number of training samples reaches 
a prescribed threshold~$\bar{N}$, 
we extend the tree~$T_{\mathrm{DLGP}}$ by growing leaf nodes~$1,2$ with data 
sets~$\mathbb{D}_1$,~$\mathbb{D}_2$ as children of the root node~$0$, as shown in the center of \cref{fig:tree}. 
In order to distribute the data efficiently to the sets~$\mathbb{D}_1$,~$\mathbb{D}_2$, we define 
a function~$p_0:\mathbb{R}^d\rightarrow[0,1]$, 
which returns the probability of an assignment of a point $\bm{x}\in\mathbb{R}^d$ to the set~$\mathbb{D}_1$, i.e.,
$P(\bm{x}\in\mathbb{D}_1|\bm{x})=p_0(\bm{x})$. 
We determine the probability~$p_0(\bm{x})$ for each data pair 
$(\bm{x},y)$ in~$\mathbb{D}_0$, and sample the child node~$i$ from the corresponding 
Bernoulli distributions. 
After the data set division, we compute the local GP models~\eqref{eq:GP model} for both data 
sets, which generally has a complexity of~$\mathcal{O}(\bar{N}^3)$~\cite{Rasmussen2006}. 
Note that the root node contains neither data nor 
a local GP model after the data set division, but instead encodes the structure of the 
data distribution using the function~$p_0(\cdot)$. Therefore,~$p_0(\cdot)$ is a crucial design
choice of the DLGP algorithm. While we propose to use a function~$p_0(\cdot)$ which 
causes a spatial division of the input space to exploit the locality of models during prediction as explained in \cref{subsec:pred}, 
arbitrary choices are possible in general. An example of a spatial data set division is depicted 
in the center of \cref{fig:tree}. \looseness=-1

After the initial data set division, we continue to assign the streaming data pairs~$(\bm{x},y)$ 
to the sets~$\mathbb{D}_1$,~$\mathbb{D}_2$ by sampling from the Bernoulli distributions 
with parameters~$p_0(\bm{x})$. When either 
of the sets~$\mathbb{D}_1$,~$\mathbb{D}_2$ reaches its data capacity limit~$\bar{N}$, we define 
a new function~$p_i(\cdot)$,~$i=1,2$, which induces the conditional probability given the parent node, e.g.,
$P(\bm{x}\in\mathbb{D}_3|\bm{x}\in\mathbb{D}_1,\bm{x})=p_1(\bm{x})$. Based on this conditional probability,
we repeat the division process as explained for the root node. Therefore, we add another level to the 
binary tree as shown on the right-hand side of \cref{fig:tree}. For further training data assignment, it is necessary 
to iteratively determine a branch of the binary tree using random transitions based on Bernoulli distributions 
with probability parameters~$p_i(\bm{x})$ until a leaf node
is reached, as outlined in \cref{alg:DLGP updates}. 
Due to this random branch sampling, the computational complexity of updating the DLGP model
strongly depends on the structure of the binary tree~$T_{\mathrm{DLGP}}$, which in turn is 
a result of the distribution and the order of the input training data~$\bm{x}$. This leads to 
a worst case complexity of~$\mathcal{O}(N+\bar{N}^3)$ for model updates, although in many cases,
e.g., for uniformly distributed input training samples, we can achieve
a lower complexity of~$\mathcal{O}(\log(N)+\bar{N}^3)$ with a balanced tree.
\looseness=-1

\begin{algorithm}[t!]
\caption{Updating the DLGP model}
\label{alg:DLGP updates}
\begin{algorithmic}[1]
\Function{update}{$T_{\mathrm{DLGP}}$,~$\bm{x},y$}
\State $i\gets T_{\mathrm{DLGP}}.$\Call{root}{ }
\While{$\neg$\Call{isLeaf}{$i$}}\Comment{random branch sampling}
	\State~$i\gets i$.\Call{getChild}{\textsc{drawBernoulli}($p_i(\bm{x})$)}
\EndWhile
\If{$|\mathbb{D}_i|=\bar{N}$} \Comment{decision about data set division}
	\State~$i$.\Call{generateChildren}{ }
	\For{\textbf{each}~$(\bm{x}',y')\in\mathbb{D}_i$} \Comment{random training set division}
		\State~$j\gets i$.\Call{getChild}{\textsc{drawBernoulli}($p_i(\bm{x}')$)}
		\State~$j$.\Call{addToDataSet}{$\bm{x}',y'$}
	\EndFor
	\State~$i.$\Call{LeftChild}{ }.\Call{ComputeLocalGP}{ } \Comment{re-computation of the GP models \eqref{eq:GP model}}
	\State~$i.$\Call{RightChild}{ }.\Call{ComputeLocalGP}{ }
	\State~$i \gets i$.\Call{getChild}{\textsc{drawBernoulli}($p_i(\bm{x})$)}
\EndIf
\State~$i$.\Call{addToDataSet}{$\bm{x},y$} \Comment{random assignment of new data pair}
\State~$i$.\Call{UpdateLocalGP}{ } \Comment{update of the GP model \eqref{eq:GP model}}

\State \Return~$T_{\mathrm{DLGP}}$
\EndFunction
\end{algorithmic}
\end{algorithm}

\subsection{Predictive Distribution}
\label{subsec:pred}

Since the training data is assigned through random sampling, predictive distributions of a DLGP
can be calculated for test points without any further approximation. Due to the binary tree structure and the fact that 
all local models are in leaf nodes, this computation is straightforward: given the binary tree, we determine 
the set of leaf nodes~$\mathbb{I}$. For each leaf~$j\in\mathbb{I}$ with depth~$\nu_j$, we compute the marginal
probability~$\tilde{p}_j(\bm{x})=P(\mathbb{D}_{j} |\bm{x})$ 
of the leaf node by multiplying the conditional probabilities~$p_i(\bm{x})$ along its
branch, i.e., \looseness=-1
\begin{align}
\label{eq:globProb}
	\tilde{p}_j(\bm{x})=\prod\limits_{i=1}^{\nu_j} p_{\left\lfloor\!\frac{j+1}{2^i}\!\right\rfloor-1}(\bm{x}).
\end{align}
By employing the definition of conditional probabilities, we obtain the predictive distribution
\begin{align}
	p_{\mathrm{DLGP}}(f(\bm{x})|\bm{x},\bm{X},\bm{y})=\sum\limits_{j\in\mathbb{I}}\tilde{p}_j(\bm{x}) p_{\mathcal{GP}\!_j}(f(\bm{x})|\bm{x},\mathbb{D}_j) ,
\end{align}
where~$p_{\mathcal{GP}\!_j}(f(\bm{x})|\bm{x},\mathbb{D}_j)$ are the distributions of the local Gaussian 
processes~$\mathcal{GP}_j$ with posterior mean~$\mu_j(\bm{x})$ and variance~$\sigma_j^2(\bm{x})$, 
which can be computed with a complexity of~$\mathcal{O}(\bar{N})$ and 
$\mathcal{O}(\bar{N}^2)$, respectively, using \eqref{eq:GP pred}. The predictions 
of all local models are independent of each other, and therefore allow efficient parallelization. 
Moreover, from an ensemble point of view, the mean and variance functions of the predictive distribution 
$p_{\mathrm{DLGP}}(f(\bm{x})|\bm{x},\bm{X},\bm{y})$ straightforwardly follow as 
\begin{align}
\label{eq:DLGP_mean}
	\mu_{\mathrm{DLGP}}(\bm{x})&=\sum\limits_{j\in\mathbb{I}}\tilde{p}_j(\bm{x})\mu_j(\bm{x})\\
	\sigma_{\mathrm{DLGP}}^2(\bm{x})&=\sum\limits_{j\in\mathbb{I}}\tilde{p}_j(\bm{x})\left(\sigma_j^2(\bm{x})+\mu_j^2(\bm{x})\right)-\mu_{\mathrm{DLGP}}^2(\bm{x}).
\end{align}
Therefore, DLGPs provide a stochastically sound framework for the computation of 
the predictive mean and variance function, and do not require further approximation like similar approaches such as, 
e.g., \cite{Nguyen-Tuong2009a, Nguyen-Tuong2009}. While these local GP methods can set an 
upper bound on the number of models, which are used for prediction, this number
grows indefinitely in DLGPs due to the continuous division of the local 
data sets. Since the computational complexity of a prediction grows with the number of local models used for
prediction, we can trivially bound it by~$\mathcal{O}(N\bar{N})$ for the mean and~$\mathcal{O}(N\bar{N}^2)$ 
for the variance prediction.\looseness=-1

In order to take advantage of the data set division in the predictions as well, we want the probabilities~$p_i(\cdot)$ to spatially 
divide the input domain, such that in large regions of the input space only a single model is active by having a positive marginal 
probability $\tilde{p}_j(\bm{x})$, 
while there is merely a small overlapping region, in which multiple models are active. For this purpose, we employ 
probability functions~$p_i(\cdot)$, which split the dimension with largest spread of the training data into two halves. Since sigmoid functions 
never reach~$0$ or~$1$, the described behavior is realized using saturating linear functions\looseness=-1
\begin{align}
p_i(\bm{x})=&\begin{cases}
0&\text{if } x_{j_i}< s_i-\frac{o_i}{2}\\
\frac{x_{j_i}-s_i}{o_i}+\frac{1}{2}& \text{if } s_i-\frac{o_i}{2}\leq x_{j_i}\leq s_{i}+\frac{o_i}{2}\\
1& \text{if } s_i+\frac{o_i}{2}< x_{j_i},
\end{cases}
\end{align}
where~$j_i$ denotes the orthogonal dimension to the nominal dividing hyperplane,~$s_i$ is the position of 
the nominal dividing hyperplane, and~$o_i$ is the size
of the overlapping region. The effect of these parameters on 
the active region of the models is depicted in \cref{fig:tree}. In order to determine 
the parameters for each conditional probability~$p_i(\cdot)$, 
we compute the width vector~$\bm{w}_i$, whose elements are defined as
$w_{i,j}=(\max_{\bm{x}\in\mathbb{D}_i} x_{j} -\min_{\bm{x}\in\mathbb{D}_i} x_{j})$. Based on
this vector, we determine the division dimension~$j_i\!=\!\argmax_{j=1,\ldots,d} w_{i,j}$, the position of the 
dividing hyperplane~$s_{i}=\sum_{\bm{x}\in\mathbb{D}_i}x_{j_i}/\bar{N}$
and the size of the overlapping region as ~$o_i=\theta w_{i_j}$, where~$\theta\in\mathbb{R}_+$ is
the overlap ratio. A detailed evaluation
of the effect of these definitions on the performance of the DLGP method is 
provided in \cref{sec:further Info}.\looseness=-1

Due to the spatial separation of the models, the complexity of predictions significantly reduces. In fact, 
when the regions, in which local models are active, have similar extensions and the overlap ratio~$\theta$ is chosen sufficiently small, 
it is straightforward to see that there exists a number~$N_{\max}(\theta)$ such that the number of active models is bounded 
by~$2^d$ (number of corners of a~$d$-dimensional hypercube) for all~$N\leq N_{\max}(\theta)$. Therefore, the efficiency of 
predictions can be improved by recursively following the paths towards leaf nodes, such that the marginal probabilities~$\tilde{p}_j(\bm{x})$
of all descendants can be immediately set to~$0$ once a conditional probability~$p_i(\bm{x})=0$ is encountered. Thereby, only 
$2^d$ marginal probabilities~$\tilde{p}_j(\bm{x})$ must be computed, each of which requires multiplying $\mathcal{O}(\log(N))$ conditional 
probabilities in a branch of a balanced binary tree. Therefore, the complexity of predictions reduces to an~$\mathcal{O}(2^d(\log(N)+\bar{N}))$ 
and~$\mathcal{O}(2 ^d(\log(N)+\bar{N}^2))$ behavior of the mean and variance computations for~$N\leq N_{\max}(\theta)$ in 
a balanced tree.\looseness=-1

\section{Numerical Evaluation}
\label{sec:Eval}
In order to demonstrate the computational efficiency and the prediction performance of DLGPs, 
we compare them to several state-of-the-art online regression approaches on real world, real-time learning
data sets. In \cref{subsec:setup}, we briefly introduce the used data sets as well as the comparison methods. 
The results of the numerical experiment are provided in \cref{subsec:results}.

\subsection{Setup}
\label{subsec:setup}

For efficiency and performance comparison, we compare the following state-of-the-art 
real-time GP regression approaches and LWPR:
\begin{itemize}
	\item our DLGP method with a fraction of the overlapping region~$\theta=0.05$ and 
	a maximum number of~$\bar{N}=100$ points per local model\looseness=-1
	\item our DLGP method with~$\theta=0.05$ and~$\bar{N}=500$
	\item local GPs \cite{Nguyen-Tuong2009} with a maximum of~$\bar{N}=500$ training samples per model, a threshold~$\bar{w}$, 
	such 	that approximately~$30$ local models are generated, and the insertion of training points based on the information gain criterion\footnote{we used code from \url{https://www.ias.informatik.tu-darmstadt.de/Miscellaneous/}\looseness=-1}
	\item the SONIG algorithm \cite{Bijl2017}, which is an online FITC approach, with inducing input point distance~$4$ and small training 
	input	noise variance of~$10^{-6}$, since this algorithm is designed for noisy training inputs\footnote{we used code from \url{https://github.com/HildoBijl/SONIG/}}
	\item the deterministic training point conditional (DTC) approximation with the maximum error criterion for insertion and 
	deletion of training data from the active set \cite{Schreiter2016} combined with an straightforward adaptation of the 
	sparse online GP algorithm~\cite{Csato2002}; the threshold for insertion is set to~$25\%$ of the standard deviation of the 
	target values corresponding to a joint\looseness=-1
	\item the incremental sparse spectrum GP (I-SSGP) approach \cite{Gijsberts2013}, in which the covariance function 
	of the GP is approximated using~$200$ sinusoidal random features
  	\item the LWPR toolbox \cite{Vijayakumar2000} with initial distance metrics~$0.5\bm{I}$, initial learning rate 
  	of~$0.5$ and~$0.3$ as weight activation 
  	threshold\footnote{we used code from \url{http://wcms.inf.ed.ac.uk/ipab/slmc/research/software-lwpr/}}
  	\item exact GP regression computed using blackbox matrix-matrix inference parallelized on GPUs~\cite{Wang2019a} as a baseline 
	for the prediction performance; 
	the precision of the employed conjugate gradient algorithm is set to~~$0.01$\footnote{we used code from \url{https://github.com/cornellius-gp/gpytorch/}}
\end{itemize}
We compare these methods in two different scenarios for learning of the inverse dynamics of robotic  manipulators, 
which is a common real-time learning problem. In the first scenario, we evaluate the performance on the 
SARCOS data set\footnote{data available at \url{http://www.gaussianprocess.org/gpml/data/}}, which consists 
of ~$44484$ training points with~$d=21$ dimensional inputs and~$7$ dimensional targets corresponding 
to the joints of the robot, which are learned independently. A subset of~$4449$ samples of the training data is 
used as test set to analyze the capabilities of the real-time learning methods to represent nonlinear functions
and monitor the learning progress.
We use GPs with squared exponential kernels, and the hyperparameters are determined a priori based 
on log-likelihood maximization with the GPyTorch toolbox \cite{Gardner2018} using~$100$ iterations of 
conjugate gradient optimization.  After this initial hyperparameter optimization, we keep them constant, 
and we investigate the learning progress after adding new data points at~$100$ uniformly spaced numbers of 
training samples in the interval~$[100,44484]$. For each joint, we determine the 
normalized mean square error (nMSE) of the evaluation on the test data following the definition of 
\cite{Nguyen-Tuong2009}. Moreover, we determine the negative log-likelihood (NLL) averaged over the 
individual test predictions, whenever the learned model provides a predictive distribution (all except local GPs and LWPR). 
This allows to 
investigate the quality of the posterior variance for determining the model uncertainty. Finally, we measure 
the average update and prediction times, for which we only take the computations of the mean function \eqref{eq:DLGP_mean}
into account 
in order to allow a fair comparison between methods with and without predictive distributions.\looseness=-1

In the second scenario, we focus on the more realistic real-time learning problem of iterating between updates
with a single data point and predictions of the next target value as proposed in \cite{Gijsberts2013}.
For this scenario, we employ the KUKA flask pushing data 
set\footnote{data available at \url{https://robotics.com.de/ds/}}, which contains data of a KUKA 
robot arm pushing around flasks with different fill levels \cite{ Rueckert2017}. We follow the approach proposed in \cite{Gijsberts2013}
and determine the hyperparameters based on~$16940$ measurements of experiments with a fill
level of~$300$ml, while the methods are evaluated with a data set of~$112761$ 
samples of experiments with fill levels of~$200$ml and~$400$ml. Due to these different fill levels in the 
offline and online data, as well as changing fill levels in the online data, this data set presents a 
challenging real-time learning problem. Each training pair consists of 
a~$d=15$ dimensional input and~$5$ target values corresponding to joints of the robotic manipulator, which 
are learned independently with the different methods. The prediction performance is evaluated based 
on the nMSE, where the mean is taken over all predictions,  and we analogously 
determine the average of the NLL to analyze the quality of the predictive distributions.\looseness=-1

\subsection{Results}
\label{subsec:results}

\begin{figure}
	\pgfplotsset{width=85\textwidth /100, compat = 1.13, 
		height =64\textwidth /100, grid= major, 
		legend cell align = left, ticklabel style = {font=\scriptsize},
		every axis label/.append style={font=\small},
		legend style = {font=\scriptsize},title style={yshift=0pt, font = \small}, 
		every x tick scale label/.style={at={(xticklabel cs:1)},anchor=south west} }
	\begin{minipage}{0.48\textwidth}
	\def\file{plots/Real_Sarcos_long_plot_data.txt}
	\begin{tikzpicture}
	\begin{axis}[grid=none,enlargelimits=false, axis on top,
	xlabel={$N$}, ylabel={nMSE}, 
	xmin=1, xmax = 44484, ymin = 0, ymax =1.4,	
	title = {prediction error}, 
	scaled ticks=false, tick label style={/pgf/number format/fixed}
	]
	\addplot[thick] table[x = steps_DLGP, y = e0_DLGP ]{\file};
	\addplot[thick, dashed] table[x = steps_DLGP500, y = e0_DLGP500 ]{\file};
	\addplot[red] table[x = steps_LGP, y = e0_LGP ]{\file};
	\addplot[red, dashed] table[x = steps_IP, y = e0_IP ]{\file};
	\addplot[blue] table[x = steps_Sparse,y = e0_Sparse ]{\file};
	\addplot[blue, dashed] table[x = steps_ISSGP,y = e0_ISSGP ]{\file};
	\addplot[green] table[x = steps_lwpr,y = e0_lwpr ]{\file};
	\addplot[green, dashed] table[x = steps_DLGP,y = e0_GPexact ]{\file};
	\end{axis}
	\end{tikzpicture}
	\end{minipage}\hspace{-1.4cm}
	\begin{minipage}{0.48\textwidth}
	\def\file{plots/Real_Sarcos_longNLL_plot_data.txt}
	\begin{tikzpicture}
	\begin{axis}[grid=none,enlargelimits=false, axis on top,
	xlabel={$N$}, ylabel={NLL},
	xmin=1, xmax = 44484, ymin = 0, ymax = 80,	
	legend style={at={(1.02,0.5)},anchor=west},
	title = {negative log-likelihood}, 
	scaled ticks=false, tick label style={/pgf/number format/fixed}	]
	\addplot[thick] table[x = steps_Sparse, y = aNLL0_DLGP  ]{\file};
	\addlegendentry{DLGP$_{100}$};
	\addplot[thick, dashed] table[x = steps_Sparse, y = aNLL0_DLGP500  ]{\file};
	\addlegendentry{DLGP$_{500}$};
	
	\addlegendimage{line legend, red};
	\addlegendentry{local GPs \cite{Nguyen-Tuong2009}};
	\addplot[red,dashed] table[x = steps_Sparse, y =aNLL0_IP  ]{\file};
	\addlegendentry{SONIG \cite{Bijl2017}};
	\addplot[blue] table[x = steps_Sparse, y = aNLL0_Sparse  ]{\file};
	\addlegendentry{DTC \cite{Schreiter2016}};
	\addplot[blue,dashed] table[x = steps_Sparse, y = aNLL0_ISSGP  ]{\file};
	\addlegendentry{I-SSGP \cite{Gijsberts2013}};
	\addlegendimage{line legend, green};
	\addlegendentry{LWPR \cite{Vijayakumar2000}};
	\addplot[green,dashed] table[x = steps_Sparse, y = aNLL0_GPexact ]{\file};
	\addlegendentry{exact GP  \cite{Wang2019a}};
	\end{axis}
	\end{tikzpicture}
	\end{minipage}
	\vspace{-0.2cm}
	\caption{Left: The nMSE of the DLGPs  is comparable to state-of-the-art online
	regression methods for small and medium numbers of training samples, but keeps decreasing when other methods 
	effectively stop learning. Right: DLGPs provide high quality predictive distributions similar to exact GPs 
	for joint 1 of the SARCOS data set, while this is an issue in existing approaches. }
	\label{fig:e}
\end{figure}
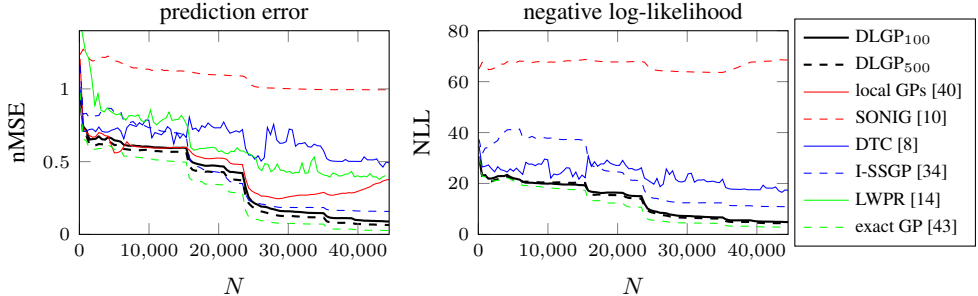
\begin{table}[!t]
	\centering
	\caption{Prediction performance (nMSE with  NLL in brackets, where available)  for the SARCOS data set after observation of all training samples with the results from exact GP regression as baseline from an off-line method;~$J_i$ denotes the~$i$-th joint.}
	\resizebox{\textwidth}{!}{%
	\begin{tabular}{l c c c c c c c || c }
		 \hline 
		& DLGP$_{100}$ & DLGP$_{500}$ & local GPs & SONIG & DTC  & I-SSGP & LWPR & exact GP \\ \hline 
		$J_1$ & 0.08(4.8) & \textbf{0.07(4.3)} & 0.38 & 0.99(68.6)  & 0.49(17.4) & 0.16(10.9) & 0.39 &0.03(2.8)\\
		$J_2$ & 0.12(4.0) & \textbf{0.10(3.7)} & 0.66 & 1.94(81.2) & 0.87(17.6) & 0.30(11.2) & 0.25 & 0.04(2.5)\\
		$J_3$ & 0.06(2.2) & \textbf{0.05(2.1)} & 0.29 & 1.21(36.5)  & 0.52(6.0) & 0.11(3.2) & 0.16 & 0.02(1.9)\\
		$J_4$ & 0.06(2.4) & \textbf{0.03(2.2)} & 0.54 & 2.25(119.3) & 0.36(7.3) & 0.13(5.2) & 0.26 & 0.01(1.8)\\
		$J_5$ & \textbf{0.01(-1.0)} & 0.02(-1.0)& 0.24& 1.19(2.4) & 0.85(11.0) & 0.33(27.6) & 0.17 & 0.007(-0.2)\\
		$J_6$ & \textbf{0.01(-0.5)} & 0.01(0.0)& 0.18& 0.92(2.5)  & 0.51(0.1) & 0.15(64.6) & 0.4 & 0.008(0.2)\\
		$J_7$ & 0.04(1.5) & \textbf{0.03(1.5)} & 0.39 & 1.56(5.2) & 0.32(1.8) & 0.11(1.6) & 0.17 & 0.01(1.6)\\
		\rule{0pt}{3ex}  
	\end{tabular}
	}\vspace{-0.5cm}
	\label{tab:err_SARCOS}
\end{table}
The prediction error and negative log-likelihood development of the different online regression methods for joint 1
of the SARCOS data set are depicted in \cref{fig:e}.\footnote{Detailed simulation results for both scenarios and all joints can be found 
in \cref{app:additional results}.} As shown at the left hand-side for the regression error, both DLGPs outperform 
existing online regression methods regarding the regression error for large numbers of training samples, and 
exhibit at least comparable performance for small and medium training set sizes. Only the exact GP provides
a better performance, since it is an offline learning method, and merely serves as baseline to demonstrate the high 
regression quality of the DLGP approach. Although the local GP approach \cite{Nguyen-Tuong2009} as 
well as the I-SSGP method~\cite{Gijsberts2013} exhibit a learning behavior comparable to DLGPs at the beginning, 
only DLGPs are capable of learning from all the data, while other methods stop prematurely due to limited 
expressiveness. The NLL curves, which are depicted at the right hand-side of \cref{fig:e}, exhibit a 
similar behavior: both DLGPs results are close to the exact GP, while the I-SSGP curve exhibits a far larger offset and 
the NLL of other methods is barely affected by additional data. 
These results generally repeat for the other joints, as 
summarized by \cref{tab:err_SARCOS}, which displays the results after observation of all training samples. 
The excellent predictive distributions of DLGPs 
are especially remarkable, since many GP approximations are known to suffer from bad posterior 
variance estimates \cite{Quinonero-candela2005}. For joints~$6$ and~$7$, the NLL of the 
DLGPs is even lower than the corresponding values of the exact GP, which is a consequence of the localizing data set division, 
such that only training samples close to a test point are employed for prediction. However, when this improvement over 
exact GPs occurs, it has the unintuitive side effect that an increase in~$\bar{N}$ causes an almost negligible 
prediction performance deterioration.\looseness=-1

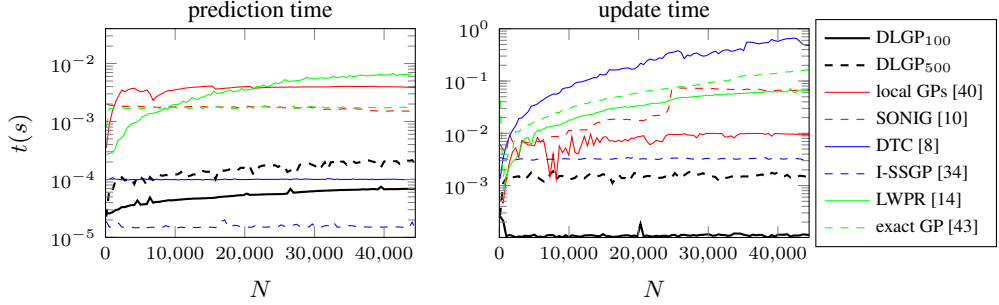
\begin{figure}

	\pgfplotsset{width=85\textwidth /100, compat = 1.13, 
		height =65\textwidth /100, grid= major, 
		legend cell align = left, ticklabel style = {font=\scriptsize},
		every axis label/.append style={font=\small},
		legend style = {font=\scriptsize},title style={yshift=0pt, font = \small}, 
		every x tick scale label/.style={at={(xticklabel cs:1)},anchor=south west} }
	\begin{minipage}{0.48\textwidth}
	\def\file{plots/Real_Sarcos_long_plot_data.txt}
	\begin{tikzpicture}
	\begin{semilogyaxis}[grid=none,enlargelimits=false, axis on top,
	xlabel={$N$}, ylabel={$t(s)$}, 
	xmin=1, xmax = 44484, ymin = 0.00001, ymax = 0.04,	
	title = {prediction time}, scaled ticks=false, tick label style={/pgf/number format/fixed}
	]
	\addplot[thick] table[x = steps_DLGP, y = t_pred0_DLGP ]{\file};
	\addplot[thick, dashed] table[x = steps_DLGP500, y = t_pred0_DLGP500 ]{\file};
	\addplot[red] table[x = steps_LGP, y = t_pred0_LGP ]{\file};
	\addplot[red,dashed] table[x = steps_IP, y = t_pred0_IP ]{\file};
	\addplot[blue] table[x = steps_Sparse,y = t_pred0_Sparse ]{\file};
	\addplot[blue,dashed] table[x = steps_ISSGP,y = t_pred0_ISSGP ]{\file};
	\addplot[green] table[x = steps_lwpr,y = t_pred0_lwpr ]{\file};
	\addplot[green,dashed] table[x = steps_DLGP,y = t_pred0_GPexact ]{\file};
	\end{semilogyaxis}
	\end{tikzpicture}
	\end{minipage}\hspace{-1.cm}
	\begin{minipage}{0.48\textwidth}
	\def\file{plots/Real_Sarcos_long_plot_data.txt}
	\begin{tikzpicture}
	\begin{semilogyaxis}[grid=none,enlargelimits=false, axis on top,
	xlabel={$N$},
	xmin=1, xmax = 44484, ymin = 0, ymax = 1,
	legend style={at={(1.02,0.5)},anchor=west},
	title = {update time}, scaled ticks=false, tick label style={/pgf/number format/fixed}
	]
	\addplot[thick] table[x = steps_DLGP, y = t_update0_DLGP ]{\file};
	\addlegendentry{DLGP$_{100}$};
	\addplot[thick, dashed] table[x = steps_DLGP500, y = t_update0_DLGP500 ]{\file};
	\addlegendentry{DLGP$_{500}$};
	\addplot[red] table[x = steps_LGP, y = t_update0_LGP ]{\file};
	\addlegendentry{local GPs \cite{Nguyen-Tuong2009}};
	\addplot[red,dashed] table[x = steps_IP, y = t_update0_IP ]{\file};
	\addlegendentry{SONIG \cite{Bijl2017}};
	\addplot[blue] table[x = steps_Sparse,y = t_update0_Sparse ]{\file};
	\addlegendentry{DTC \cite{Schreiter2016}};
	\addplot[blue, dashed] table[x = steps_ISSGP,y = t_update0_ISSGP ]{\file};
	\addlegendentry{I-SSGP \cite{Gijsberts2013}};
	\addplot[green] table[x = steps_Sparse,y = t_update0_lwpr ]{\file};
	\addlegendentry{LWPR \cite{Vijayakumar2000}};
	\addplot[green,dashed] table[x = steps_DLGP,y = t_update0_GPexact ]{\file};
	\addlegendentry{exact GP  \cite{Wang2019a}};
	\end{semilogyaxis}
	\end{tikzpicture}
	\end{minipage}
	\vspace{-0.2cm}
	\caption{Left: The prediction time of DLGPs  increases logarithmitcally on the SARCOS data set (joint 1), and remains 
	smaller than for many existing methods. 
	Right: The update time of DLGPs is almost constant due to a slow logarithmic growth, such that the 
	DLGP method is significantly faster than all state-of-the-art methods.}
	\label{fig:t_update}
\end{figure}

Since small numbers~$\bar{N}$ are preferable due to lower computational complexity, 
this slight deterioration is not a critical disadvantage. The effect of different values for~$\bar{N}$ 
can be seen at the left-hand side of \cref{fig:t_update}, which depicts the average computation 
time of predictions for the different methods. The higher value of~$\bar{N}$ slows down the prediction, but
the prediction times of both DLGPs grow slowly and in fact exhibit a logarithmic dependency on the 
number of training samples~$N$. Although other methods such as the online GP with DTC approximation \cite{Schreiter2016}
and the I-SSGP \cite{Gijsberts2013} have constant prediction times, DLGPs are only slightly slower and could be further 
sped up by reducing~$\bar{N}$ or reducing the overlap ratio~$\theta$. 
Moreover, both DLGPs significantly outperform the state-of-the-art method regarding 
the average update time, as illustrated at the right-hand side of \cref{fig:t_update}. While other methods exhibit a 
linear growth in update time, it grows logarithmically for the DLGPs, which almost seems constant due to the small 
slope. Thereby, DLGPs achieve even lower update times than methods with constant update complexity such as I-SSGP, 
and allow predictions \textbf{and} updates at the high rates necessary for real-time learning.

When applying the methods to the real-time learning problem in our second scenario, similar results
for the prediction errors and predictive distributions can be observed, as summarized in \cref{tab:err_KUKA}. However, 
the advantages of local training sets can be seen even more strongly since even the local GP approach 
exhibits better performance than many other approaches. Therefore, the DLGP with~$\bar{N}=100$ exhibits slightly
better performance than our second DLGP with higher value of~$\bar{N}$. Despite of this unintuitive behavior, 
both DLGPs outperform all other methods in terms of the nMSE and NLL. Therefore, DLGPs are particularly 
suited for real-time regression problems due to the high prediction accuracy, trustworthy predictive distributions
as well as low update and prediction times.\looseness=-1

\begin{table}
	\centering
	\caption{Prediction performance (nMSE and NLL in brackets) for the KUKA flask pushing data set after observation of all 
	training samples; exact GP regression cannot be evaluated due to exceedingly high computation time;~$J_i$ denotes the~$i$-th joint.}
	\begin{tabular}{l c c c c c c c c}
		 \hline 
		& DLGP$_{100}$& DLGP$_{500}$ & local GPs & SONIG & DTC  & I-SSGP & LWPR\\ \hline 
		$J_1$ & \textbf{0.04(-2.8)} & 0.04(-2.8)  & 0.08 & 0.81(-0.3) & 0.82(28.3) & 0.13(5.4) & 0.34\\
		$J_2$ & \textbf{0.06(-2.8)} & 0.06(-2.8)  & 0.09 & 0.84(-0.5) & 0.95(19.0) & 0.16(4.3) & 0.50\\
		$J_3$ & \textbf{0.09(-2.9)} & 0.11(-2.7)  & 0.31 & 0.87(-0.9) & 0.89(23.8) & 0.24(5.1) & 0.53\\
		$J_4$ & \textbf{0.04(-2.8)} & 0.04(-2.7) & 0.13 & 0.73(0.2) & 0.73(20.3) & 0.12(11.1) & 0.32\\ 
		$J_5$ & \textbf{0.07(-2.8)} & 0.08(-2.7) & 0.08 & 1.00(0.9) & 0.98(14.6) & 0.24(17.7) & 0.69\\
		\rule{0pt}{3ex}   
	\end{tabular}\vspace{-0.5cm}
	\label{tab:err_KUKA}
\end{table}

\section{Conclusion}
\label{sec:Conclusion}
This paper presents a novel online regression method, which bases on the division of local
Gaussian process models. It allows to preform updates for new incoming data points and predictions  
with logarithmic complexity in practice. Our numerical evaluation shows, that the accuracy is higher than for most 
existing methods and particularly shows advantages for large data sets. Moreover, the predictive distributions
are more reliable and update times are significantly below the state of the art.\looseness=-1

\section*{Broader Impact}

The major positive outcome of the presented research lies 
in the development of an online applicable regression method for fast processes in sequential decision making 
problems. The proposed method neither provides
direct benefits to anybody, nor does it put anybody at 
disadvantage. There is no immediate consequence of 
failure for online regression, and similarly, biases are 
not actively exploited.

\bibliographystyle{IEEEtran}
\bibliography{myBib}

\appendix

\section{Hardware and Software used in the Simulations}

All simulations were performed on a computer with an Intel(R) Core(TM) i9-9900X CPU and 
$128$GB RAM. The GPU computations of exact GP regression were performed on two GeForce RTX 2080 Ti and
one NVIDIA TITAN V GPUs. DLGPs, local GPs, SONIG, sparse GPs with DTC approximation and incremental
sparse spectrum GPs were implemented using MATLAB R2019a, while for exact GPs a Python implementation was employed
and LWPR is based on a C++ implementation.

\section{Further Information on the DLGP algorithm}
\label{sec:further Info}

This section provides details about the available parameters, the chosen values in the simulations and their influence
on the performance of DLGP. In general and if not stated differently, the default parameters for a DLGP run are set to
$\theta = 0.05$ for the fraction of the overlapping region and $\bar{N} = 100$ for the maximum number of points per local model.
The nominal dividing hyperplane is positioned using the mean of the contained data points. \\
In the following, each data set for evaluating the impact of a parameter is generated by performing 100 Monte Carlo simulations
in order to eliminate probabilistic effects of the random data assignment. Each 
figure depicts the mean (full lines) and standard deviation (shaded areas) of the Monte-Carlo simulations. 
The simulation scenario here is the same as the first one described in \cref{subsec:setup}. While the data set is the same, the computation times are measured based on the mean and variance predictions,
and thus, show increased values compared to the main article.

\subsection{Position of the nominal dividing hyperplane}\label{sec:divMethod}

As described in \cref{subsec:random trans}, the position of the dividing hyperplane influences the data partitioning
and has to be chosen such that the two resulting leaf nodes contain an approximately equal amount of points in order to achieve a
balanced binary tree. Hence, natural ways of calculating $s_i$ are

\begin{align}
s_i &= \med\limits_{\bm{x}\in\mathbb{D}_i}(x_{j_i}) \quad \forall x_{j_i} \in \mathbb{D} \label{eq:simedian} \\
s_i &= \frac{1}{\bar{N}} \sum_{\bm{x}\in\mathbb{D}_i} x_{j_i} \label{eq:simean} \\
s_i &= \frac{1}{2} \left( \max_{\bm{x}\in\mathbb{D}_i}{x_{j_i} } - \min_{\bm{x}\in\mathbb{D}_i} x_{j_i}\right), \label{eq:sidist}
\end{align}

where~\eqref{eq:simedian} and~\eqref{eq:simean} use the median and mean, respectively, of all $\bm{x}\in\mathbb{D}_i$ in the $j_i$-th dimension, 
and~\eqref{eq:sidist} is based on half the distance between minimum and maximum of the data points. In the main article, the
calculations and derivations use the strategy based on the mean value from~\eqref{eq:simean}.

\begin{figure}[p]
	\pgfplotsset{width=110\textwidth /100, compat = 1.13,
		height =100\textwidth /100, grid= major,
		legend cell align = left, ticklabel style = {font=\scriptsize},
		every axis label/.append style={font=\small},
		legend style = {font=\scriptsize},title style={yshift=0pt, font = \small},
	}
	
	\begin{minipage}{0.32\textwidth}
		\def\file{plots/div_test.txt}
		\begin{tikzpicture}
		\begin{axis}[grid=none,enlargelimits=false,
		ylabel={$t(s)$}, 
		xmin=0.25, xmax = 3.75, ymin = 0, ymax =6,	
		y label style={yshift=-0.2cm},
		legend style={at={(1.02,0.5)},anchor=west},xtick pos=left,
		title = {total update time}, scaled ticks=false,
		,ybar,xtick={1.0,2,3.0},xticklabels={median, mean, minmax}
		]
		\addplot[fill=black!25,point meta=y,error bars/.cd,y explicit, y dir=both,] table[x = divMethod, y = t_update_mean,y error=t_update_error  ]{\file};
		\end{axis}
		\end{tikzpicture}
	\end{minipage}\hspace{-.2cm}
	\begin{minipage}{0.32\textwidth}
		\center
		\def\file{plots/div_test.txt}
		\begin{tikzpicture}
		\begin{axis}[grid=none,enlargelimits=false,
		ylabel={$t(s)$}, 
		xmin=0.25, xmax = 3.75, ymin = 0, ymax =1.1,
		y label style={yshift=-0.2cm},
		legend style={at={(1.02,0.5)},anchor=west},xtick pos=left,
		title = {total prediction time}, scaled ticks=true,
		,ybar,xtick={1.0,2,3.0},xticklabels={median,mean,minmax}
		]
		\addplot[fill=black!25,point meta=y,error bars/.cd,y explicit, y dir=both,] table[x = divMethod, y = t_pred_mean,y error=t_pred_error  ]{\file};
		\end{axis}
		\end{tikzpicture}
	\end{minipage}\hspace{-.0cm}
	\begin{minipage}{0.32\textwidth}
		\center
		\def\file{plots/div_test.txt}
		\begin{tikzpicture}
		\begin{axis}[grid=none,enlargelimits=false,
		ylabel={$O_D(\%)$},
		xmin=0.25, xmax = 3.75, ymin = 0, ymax =0.18,	
		y label style={yshift=-0.2cm},
		legend style={at={(1.02,0.5)},anchor=west},xtick pos=left,
		title = {$\%$ of points in overlapping region}, scaled ticks=false, tick label style={/pgf/number format/fixed},
		,ybar,xtick={1.0,2,3.0},xticklabels={median,mean,minmax}
		]
		\addplot[fill=black!25,point meta=y,error bars/.cd,y explicit, y dir=both,] table[x = divMethod, y = overlapRatio_mean,y error=overlapRatio_error  ]{\file};
		\end{axis}
		\end{tikzpicture}
	\end{minipage}
	
	\begin{minipage}{0.32\textwidth}
		\def\file{plots/div_test.txt}
		\begin{tikzpicture}
		\begin{axis}[grid=none,enlargelimits=false,
		ylabel={nMSE}, 
		xmin=0.25, xmax = 3.75, ymin = 0, ymax =0.1,
		y label style={yshift=-0.2cm},
		legend style={at={(1.02,0.5)},anchor=west},xtick pos=left,
		title = {prediction error}, scaled ticks=false, tick label style={/pgf/number format/fixed},
		,ybar,xtick={1,2,3},xticklabels={median,mean,minmax}
		]
		\addplot[fill=black!25,point meta=y,error bars/.cd,y explicit, y dir=both,] table[x = divMethod, y = error_mean,y error=error_error  ]{\file};
		\end{axis}
		\end{tikzpicture}
	\end{minipage}\hspace{-.0cm}
	\begin{minipage}{0.32\textwidth}
		\center
		\def\file{plots/div_test.txt}
		\begin{tikzpicture}
		\begin{axis}[grid=none,enlargelimits=false,
		ylabel={$NLL$},
		xmin=0.25, xmax = 3.75, ymin = 0, ymax =5.5,
		y label style={yshift=-0.2cm},
		legend style={at={(1.02,0.5)},anchor=west},xtick pos=left,
		title = {negative log-likelihood}, scaled ticks=true,
		,ybar,xtick={1,2,3},xticklabels={median,mean,minmax}
		]
		\addplot[fill=black!25,point meta=y,error bars/.cd,y explicit, y dir=both,] table[x = divMethod, y = nllMeans_mean,y error=nllMeans_error  ]{\file};
		\end{axis}
		\end{tikzpicture}
	\end{minipage}\hspace{-.0cm}
	\begin{minipage}{0.32\textwidth}
		\center
		\def\file{plots/div_test.txt}
		\begin{tikzpicture}
		\begin{axis}[grid=none,enlargelimits=false,
		ylabel={$\#$divisions}, 
		xmin=0.25, xmax = 3.75, ymin = 0, ymax =950,
		y label style={yshift=-0.2cm},
		legend style={at={(1.02,0.5)},anchor=west},xtick pos=left,
		title = {number of divisions}, scaled ticks=false, tick label style={/pgf/number format/fixed},
		,ybar,xtick={1,2,3},xticklabels={median,mean,minmax}
		]
		\addplot[fill=black!25,point meta=y,error bars/.cd,y explicit, y dir=both,] table[x = divMethod, y = divCount_mean,y error=divCount_error  ]{\file};
		\end{axis}
		\end{tikzpicture}
	\end{minipage}
	
	\caption{Comparison of strategies to obtain position of dividing hyperplane $s_i$, where the methods are mean, median
		and half distance between min and max (minmax).}
	\label{fig:divMethod}
\end{figure}
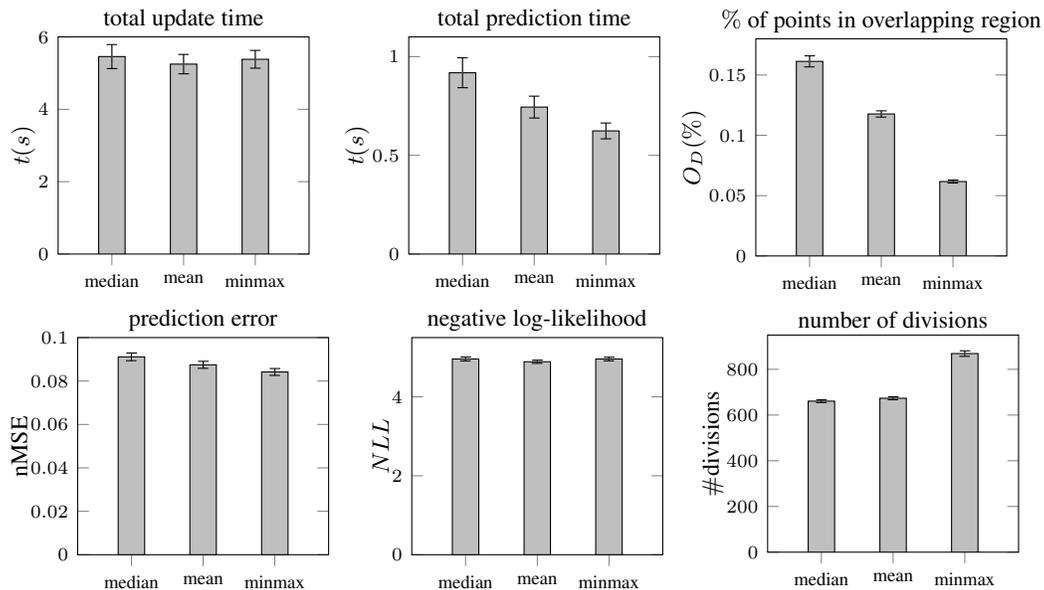

\begin{figure}[p]
	\pgfplotsset{width=85\textwidth /100, compat = 1.13,
		height =74\textwidth /100, grid= major,
		legend cell align = left, ticklabel style = {font=\scriptsize},
		every axis label/.append style={font=\small},
		legend style = {font=\scriptsize},title style={yshift=0pt, font = \small},
		every x tick scale label/.style={at={(xticklabel cs:1)},anchor=south west} }
	\begin{minipage}{0.48\textwidth}
		\center
		\def\file{plots/div_Method_examples.txt}
		\begin{tikzpicture}
		\begin{axis}[grid=none,enlargelimits=false, axis on top,
		xlabel={$N$}, ylabel={nMSE}, 
		xmin=0, xmax = 44484,
		title = {error},  tick label style={/pgf/number format/fixed}
		]
		\addplot[blue!20,fill=blue!20] table[x = Ns_long, y = error_error_med  ]{\file};
		\addplot[black!20,fill=black!20] table[x = Ns_long, y = error_error_med  ]{\file};
		\addplot[green!20,fill=green!20] table[x = Ns_long, y = error_error_med  ]{\file};
		\addplot[blue] table[x = Ns, y = error_mean_med  ]{\file};
		\addplot[thick] table[x = Ns, y =  error_mean_mean  ]{\file};
		\addplot[green] table[x = Ns, y =  error_mean_mm  ]{\file};
		\end{axis}
		\end{tikzpicture}
	\end{minipage}	\hspace{-0.5cm}
	\begin{minipage}{0.48\textwidth}
		\center
		\def\file{plots/div_Method_examples.txt}
		\begin{tikzpicture}
		\begin{axis}[grid=none,enlargelimits=false, axis on top,
		xlabel={$N$}, ylabel={NLL}, 
		xmin=0, xmax = 44484,
		legend style={at={(1.02,0.5)},anchor=west},
		title = {negative log-likelihood},  tick label style={/pgf/number format/fixed}
		]
		\addplot[blue!20,fill=blue!20] table[x = Ns_long, y = nllMeans_error_med  ]{\file};
		\addplot[black!20,fill=black!20] table[x = Ns_long, y = nllMeans_error_mean  ]{\file};
		\addplot[green!20,fill=green!20] table[x = Ns_long, y = nllMeans_error_mm  ]{\file};
		\addplot[blue] table[x = Ns, y = nllMeans_mean_med  ]{\file};
		\addplot[thick] table[x = Ns, y =  nllMeans_mean_mean  ]{\file};
		\addplot[green] table[x = Ns, y =  nllMeans_mean_mm  ]{\file};
		\end{axis}
		\end{tikzpicture}
	\end{minipage}

	\begin{minipage}{0.48\textwidth}
		\center
		\def\file{plots/div_Method_examples.txt}
		\begin{tikzpicture}
		\begin{axis}[grid=none,enlargelimits=false, axis on top,
		xlabel={$N$}, ylabel={$t(s)$}, 
		xmin=0, xmax = 44484,
		title = {average update time},  tick label style={/pgf/number format/fixed},
		every y tick scale label/.style={at={(yticklabel cs:1)},anchor=south west},
		]
		\addplot[blue!20,fill=blue!20] table[x = Ns_long, y = t_update_error_med ]{\file};
		\addplot[black!20,fill=black!20] table[x = Ns_long, y = t_update_error_mean ]{\file};
		\addplot[green!20,fill=green!20] table[x = Ns_long, y = t_update_error_mm ]{\file};
		
		\addplot[blue] table[x = Ns, y = t_update_mean_med ]{\file};
		\addplot[thick] table[x = Ns, y = t_update_mean_mean ]{\file};
		\addplot[green] table[x = Ns, y = t_update_mean_mm ]{\file};
		\end{axis}
		\end{tikzpicture}
	\end{minipage}	\hspace{-0.5cm}
	\begin{minipage}{0.48\textwidth}
		\center
		\def\file{plots/div_Method_examples.txt}
		\begin{tikzpicture}
		\begin{axis}[grid=none,enlargelimits=false, axis on top,
		xlabel={$N$}, ylabel={$t(s)$}, 
		xmin=0, xmax = 44484,
		legend style={at={(1.02,0.5)},anchor=west},
		title = {average prediction time}, tick label style={/pgf/number format/fixed},
		every y tick scale label/.style={at={(yticklabel cs:1)},anchor=south west},
		]
		\addplot[blue!20,fill=blue!20] table[x = Ns_long, y = t_pred_error_med ]{\file};
		\addplot[black!20,fill=black!20] table[x = Ns_long, y = t_pred_error_mean ]{\file};
		\addplot[green!20,fill=green!20] table[x = Ns_long, y = t_pred_error_mm ]{\file};
		
		\addplot[blue] table[x = Ns, y = t_pred_mean_med ]{\file};
		\addlegendentry{median}
		\addplot[thick] table[x = Ns, y = t_pred_mean_mean ]{\file};
		\addlegendentry{mean}
		\addplot[green] table[x = Ns, y = t_pred_mean_mm ]{\file};
		\addlegendentry{minmax}
		\end{axis}
		\end{tikzpicture}
	\end{minipage}
	\caption{Prediction error, NLL, update and prediction time for joint 1 through the learning process obtained with different
		methods for choosing $s_i$.}
	\label{fig:times_divM}
\end{figure}
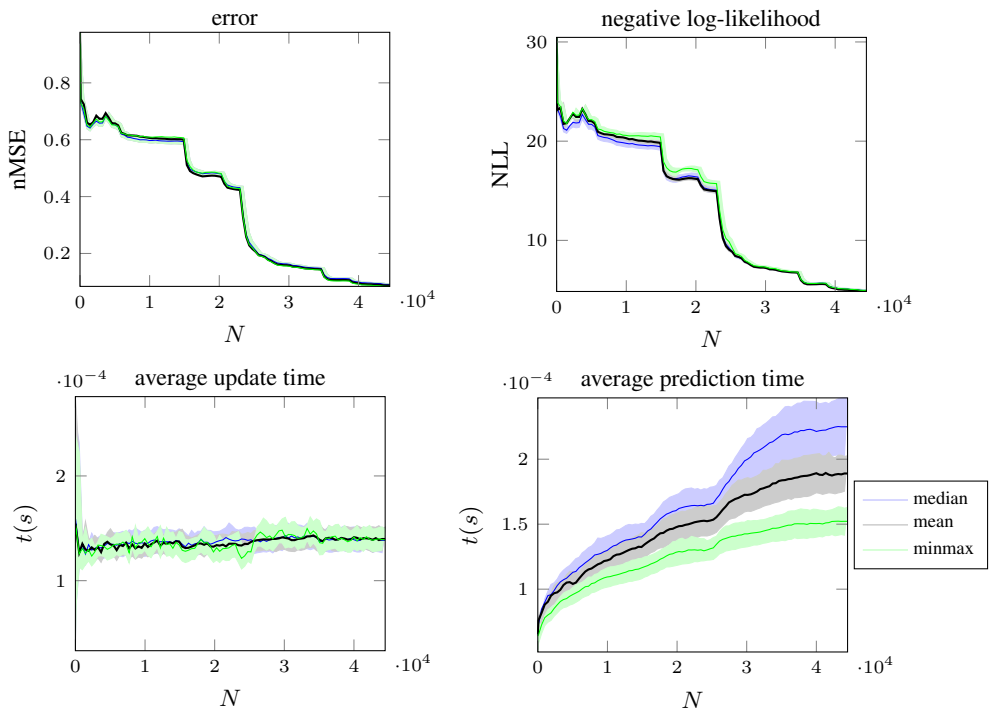

Figure~\ref{fig:divMethod} provides a comparison of these three methods using the overall update and prediction times, normalized mean
squared prediction error,
average negative log-likelihood (NLL), number of tree divisions and the ratio $O_D$. This ratio is defined as the percentage of points,
which lie in the overlapping region during all model divisions. It can be observed that
no method significantly drops off in any of the provided measures. In contrast, the update time, prediction error and negative
log-likelihood only differ in single-digit percentage ranges. However, there is a more substantial effect on the prediction
time as the results obtained with~\eqref{eq:sidist} yield a $31\%$ smaller time than the method from~\eqref{eq:simedian} while
the number of divisions indicates a reverse behavior.
The ratio $O_D$ hereby allows a more indirect insight. While more points in the overlapping region theoretically lead to smoother
functions in that region, less points entail less computational complexity and lower prediction times. This effect can
be seen when comparing the ratio $O_D$ with the prediction times in Figure~\ref{fig:divMethod} as they show a related behavior.
Meanwhile, the number of divisions only has effects on the updating part of the GP and leaves the predictions unchanged.

Additionally, the plots in Figure~\ref{fig:times_divM} show the dependency on the training set size of the same simulation for prediction
error, NLL, average update and prediction times, where especially for the accuracy measured by error, no significant difference
is visible. However, the NLL of the median and minmax approach exhibit higher values than the mean division method, in particular
at low numbers of training samples. While the average update time evolves as discussed above independently of the 
amount of training data, the average prediction time is subject to few 
outliers in the median and distance based methods, with the mean being the smoothest approach.

In summary, the simulation results in Figures~\ref{fig:divMethod} and~\ref{fig:times_divM} allow to conclude that 
the performance cannot be influenced substantially
by choosing a different dividing method, resulting in a robust interface for potential use-cases. However, it still provides
room for tuning the performance in a certain domain if required by the application scenario. If a trade-off between all measures
is required, using the mean~\eqref{eq:simean} suggests the best result in this particular setup.

\subsection{Overlapping factor}

The overlapping factor $\theta$ allows to adjust the smoothness of the transition between two neighboring child sets in the binary tree.
In general, a greater overlapping region is achieved by a larger $\theta$, which yields a smoother model with the drawback of
increased computational complexity of predictions.

Figure~\ref{fig:overlap_evaluation} depicts the influence of the overlapping factor on the same performance measures as in
Section~\ref{sec:divMethod}. It is evident that $\theta$ has no effect on the update time, since the
size of the overlapping region has no direct impact on the computational complexity of updates. Additionally, the almost linear
dependency between the ratio $O_D$ of points in the overlapping region and $\theta$ can be directly deduced.
The higher the overlapping factor, the greater is the overlapping region with more points in it. The prediction performance 
in terms of computation times, error and NLL is affected strongly. The complexity analysis
from \cref{subsec:pred} emerges especially for the total prediction time. With increasing~$\theta$, the
predictions evolve from logarithmic to linear complexity since larger overlapping regions lead to more active local models such
that more local predictions have to be calculated.
As one can see in the NLL and error plots, the accuracy declines with an increasing $\theta$. One reason for this lies in the
constant number of samples $N$ and the increasing number of divisions which leads to many local models with only a few data points.
This suggests that the data assignment does not yield a tree with equally distributed points and a worsening accuracy in
general. 

\begin{figure}[p]
	\pgfplotsset{width=85\textwidth /100, compat = 1.13, 
		height =74\textwidth /100, grid= major, 
		legend cell align = left, ticklabel style = {font=\scriptsize},
		every axis label/.append style={font=\small},
		legend style = {font=\scriptsize},title style={yshift=0pt, font = \small},
		every x tick scale label/.style={at={(xticklabel cs:1)},anchor=south west} }
	\begin{minipage}{0.48\textwidth}
		\center
		\def\file{plots/overlap_examples.txt}
		\begin{tikzpicture}
		\begin{axis}[grid=none,enlargelimits=false, axis on top,
		xlabel={$N$}, ylabel={nMSE}, 
		xmin=0, xmax = 44484, 
		title = {error},  tick label style={/pgf/number format/fixed}
		]
		\addplot[blue!20,fill=blue!20] table[x = Ns_long, y = error_error_001  ]{\file};
		\addplot[black!20,fill=black!20] table[x = Ns_long, y = error_error_005  ]{\file};
		\addplot[green!20,fill=green!20] table[x = Ns_long, y = error_error_03  ]{\file};
		
		\addplot[blue] table[x = Ns, y = error_mean_001  ]{\file};
		\addplot[thick] table[x = Ns, y = error_mean_005  ]{\file};
		\addplot[green] table[x = Ns, y = error_mean_03  ]{\file};
		\end{axis}
		\end{tikzpicture}
	\end{minipage}	\hspace{-0.5cm}
	\begin{minipage}{0.48\textwidth}
		\center
		\def\file{plots/overlap_examples.txt}
		\begin{tikzpicture}
		\begin{axis}[grid=none,enlargelimits=false, axis on top,
		xlabel={$N$}, ylabel={NLL}, 
		xmin=0, xmax = 44484, 
		legend style={at={(1.02,0.5)},anchor=west},
		title = {negative log-likelihood},  tick label style={/pgf/number format/fixed}
		]
		\addplot[blue!20,fill=blue!20] table[x = Ns_long, y = nllMeans_error_001  ]{\file};
		\addplot[black!20,fill=black!20] table[x = Ns_long, y = nllMeans_error_005  ]{\file};
		\addplot[green!20,fill=green!20] table[x = Ns_long, y = nllMeans_error_03  ]{\file};
		
		\addplot[blue] table[x = Ns, y = nllMeans_mean_001  ]{\file};
		\addplot[thick] table[x = Ns, y =  nllMeans_mean_005  ]{\file};
		\addplot[green] table[x = Ns, y =  nllMeans_mean_03  ]{\file};
		\end{axis}
		\end{tikzpicture}
	\end{minipage}

	\begin{minipage}{0.48\textwidth}
		\center
		\def\file{plots/overlap_examples.txt}
		\begin{tikzpicture}
		\begin{axis}[grid=none,enlargelimits=false, axis on top,
		xlabel={$N$}, ylabel={$t(s)$},
		xmin=0, xmax = 44484, 
		title = {average update time},  tick label style={/pgf/number format/fixed},
		every y tick scale label/.style={at={(yticklabel cs:1)},anchor=south west},
		]
		\addplot[blue!20,fill=blue!20] table[x = Ns_long, y = t_update_error_001  ]{\file};
		\addplot[black!20,fill=black!20] table[x = Ns_long, y = t_update_error_005  ]{\file};
		\addplot[green!20,fill=green!20] table[x = Ns_long, y = t_update_error_03  ]{\file};
		
		\addplot[blue] table[x = Ns, y = t_update_mean_001 ]{\file};
		\addplot[thick] table[x = Ns, y = t_update_mean_005 ]{\file};
		\addplot[green] table[x = Ns, y = t_update_mean_03 ]{\file};
		\end{axis}
		\end{tikzpicture}
	\end{minipage}	\hspace{-0.5cm}
	\begin{minipage}{0.48\textwidth}
		\center
		\def\file{plots/overlap_examples.txt}
		\begin{tikzpicture}
		\begin{axis}[grid=none,enlargelimits=false, axis on top,
		xlabel={$N$}, ylabel={$t(s)$}, 
		xmin=0, xmax = 44484,
		legend style={at={(1.02,0.5)},anchor=west},
		title = {average prediction time}, tick label style={/pgf/number format/fixed},
		every y tick scale label/.style={at={(yticklabel cs:1)},anchor=south west},
		]
		\addplot[blue!20,fill=blue!20] table[x = Ns_long, y = t_pred_error_001  ]{\file};
		\addplot[black!20,fill=black!20] table[x = Ns_long, y = t_pred_error_005  ]{\file};
		\addplot[green!20,fill=green!20] table[x = Ns_long, y = t_pred_error_03  ]{\file};
		
		\addplot[blue] table[x = Ns, y = t_pred_mean_001 ]{\file};
		\addlegendentry{$\theta=0.01$}
		\addplot[thick] table[x = Ns, y = t_pred_mean_005 ]{\file};
		\addlegendentry{$\theta=0.05$}
		\addplot[green] table[x = Ns, y = t_pred_mean_03 ]{\file};
		\addlegendentry{$\theta=0.3$}
		\end{axis}
		\end{tikzpicture}
	\end{minipage}
	\caption{Prediction error, NLL, update and prediction time for joint 1 through the learning process obtained with different
		values for the overlapping factor $\theta$.}
	\label{fig:times_theta}
\end{figure}
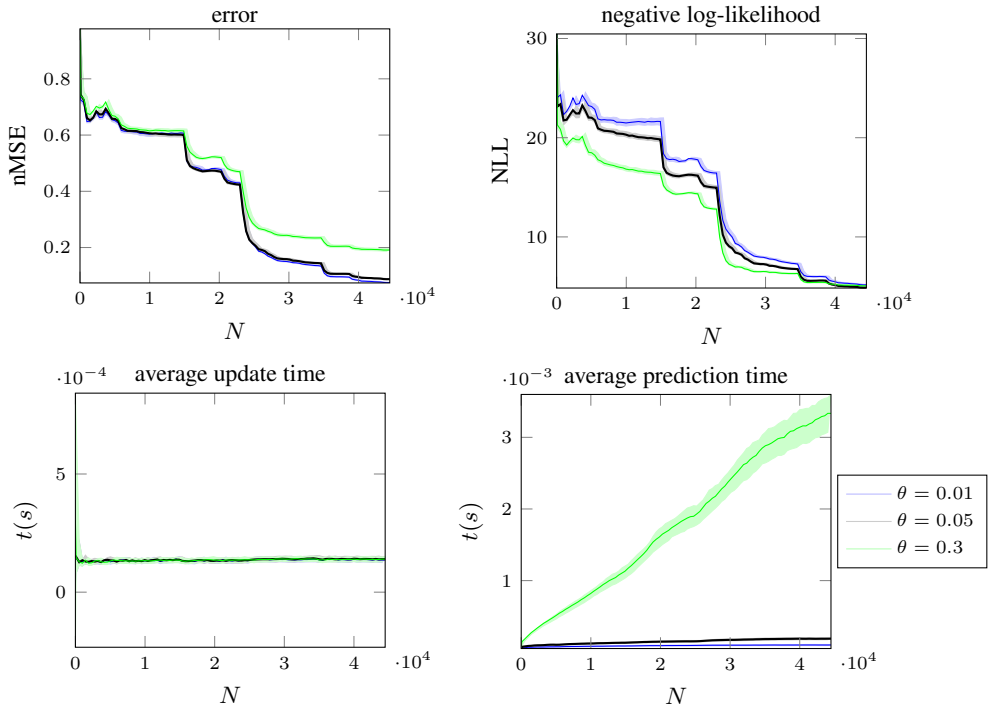

Accordingly, Figure~\ref{fig:times_theta} shows these results depending on the number of training samples for joint~1. 
The NLL for $\theta = 0.3$ starts at a smaller value, but lower values of $\theta$ 
approach a similar performance 
with increasing $N$. In contrast, the errors start on
a similar level, but the differences between the different values of $\theta$ increase.
Finally, looking at the learning process in Figure~\ref{fig:times_theta} consolidates the findings regarding the computation times.
For example, the
choice of $\theta$ does not influence the average update time. For the average prediction time, once again the complexity discussion from the
main article becomes clearly visible. Choosing a sufficiently small $\theta$ such that $N < N_{\max}(\theta)$ leads to a logarithmic
complexity compared to a linear behavior for too large $\theta$. Therefore, choosing reasonable parameters can drastically improve
the performance by eliminating the linear computational complexity.

\begin{figure}[p]
	\pgfplotsset{width=120\textwidth /100, compat = 1.13, 
		height =100\textwidth /100, grid= major, 
		legend cell align = left, ticklabel style = {font=\scriptsize},
		every axis label/.append style={font=\small},
		legend style = {font=\scriptsize},title style={yshift=0pt, font = \small},
		every x tick scale label/.style={at={(xticklabel cs:1)},anchor=south west} }
	\begin{minipage}{0.3\textwidth}
		\center
		\def\file{plots/overlap_test.txt}
		\begin{tikzpicture}
		\begin{axis}[grid=none,enlargelimits=false, axis on top,
		xlabel={$\theta$}, ylabel={$t(s)$}, 
		xmin=0, xmax = 0.5, ymin = 8,
		y label style={yshift=-0.2cm},
		title = {total update time},  tick label style={/pgf/number format/fixed}
		]
		\addplot[fill=black!25] table[x = wo_long, y = t_update_error ]{\file};
		\addplot[thick] table[x = wo, y = t_update_mean ]{\file};
		
		\end{axis}
		\end{tikzpicture}
	\end{minipage}	\hspace{0.2cm}
	\begin{minipage}{0.3\textwidth}
		\center
		\def\file{plots/overlap_test.txt}
		\begin{tikzpicture}
		\begin{axis}[grid=none,enlargelimits=false, axis on top,
		xlabel={$\theta$}, ylabel={$t(s)$},
		xmin=0, xmax = 0.5, ymin =0, 
		y label style={yshift=-0.2cm},
		title = {total prediction time}, tick label style={/pgf/number format/fixed}
		]
		\addplot[fill=black!25] table[x = wo_long, y = t_pred_error ]{\file};
		\addplot[thick] table[x = wo, y = t_pred_mean ]{\file};
		\end{axis}
		\end{tikzpicture}
	\end{minipage}\hspace{0.2cm}
	\begin{minipage}{0.3\textwidth}
		\center
		\def\file{plots/overlap_test.txt}
		\begin{tikzpicture}
		\begin{axis}[grid=none,enlargelimits=false, axis on top,
		xlabel={$\theta$}, ylabel={$O_D(\%)$},
		xmin=0, xmax = 0.5, ymin = 0, 
		y label style={yshift=-0.2cm},
		title = {$\%$ of points in overlapping region}, tick label style={/pgf/number format/fixed}
		]
		\addplot[fill=black!25] table[x = wo_long, y = overlapRatio_error ]{\file};
		\addplot[thick] table[x = wo, y = overlapRatio_mean ]{\file};
		\end{axis}
		\end{tikzpicture}
	\end{minipage}\\
	
	\begin{minipage}{0.3\textwidth}
		\def\file{plots/overlap_test.txt}
		\begin{tikzpicture}
		\begin{axis}[grid=none,enlargelimits=false, axis on top,
		xlabel={$\theta$}, ylabel={nMSE}, 
		xmin=0, xmax = 0.5, ymin =0, 
		y label style={yshift=-0.2cm},
		title = {prediction error},  tick label style={/pgf/number format/fixed}
		]
		\addplot[fill=black!25] table[x = wo_long, y = error_error ]{\file};
		\addplot[thick] table[x = wo, y = error_mean ]{\file};
		\end{axis}
		\end{tikzpicture}
	\end{minipage}	\hspace{0.2cm}
	\begin{minipage}{0.3\textwidth}
		\center
		\def\file{plots/overlap_test.txt}
		\begin{tikzpicture}
		\begin{axis}[grid=none,enlargelimits=false, axis on top,
		xlabel={$\theta$}, ylabel={$NLL$},
		xmin=0, xmax = 0.5, ymin = 4.8, 
		y label style={yshift=-0.2cm},
		title = {negative log-likelihood}, tick label style={/pgf/number format/fixed}
		]
		\addplot[fill=black!25] table[x = wo_long, y = nllMeans_error ]{\file};
		\addplot[thick] table[x = wo, y = nllMeans_mean ]{\file};
		\end{axis}
		\end{tikzpicture}
	\end{minipage}\hspace{0.2cm}
	\begin{minipage}{0.3\textwidth}
		\center
		\def\file{plots/overlap_test.txt}
		\begin{tikzpicture}
		\begin{axis}[grid=none,enlargelimits=false, axis on top,
		xlabel={$\theta$}, ylabel={$\#$divisions},
		xmin=0, xmax = 0.5, 
		y label style={yshift=-0.2cm},
		title = {number of divisions}, tick label style={/pgf/number format/fixed}
		]
		\addplot[fill=black!25] table[x = wo_long, y = divCount_error ]{\file};
		\addplot[thick] table[x = wo, y = divCount_mean ]{\file};
		\end{axis}
		\end{tikzpicture}
	\end{minipage}
	\caption{Influence of varying overlapping factor $\theta$ on the update and prediction time, prediction error, NLL, points
		in overlapping region and number of divisions.}
	\label{fig:overlap_evaluation}
\end{figure}
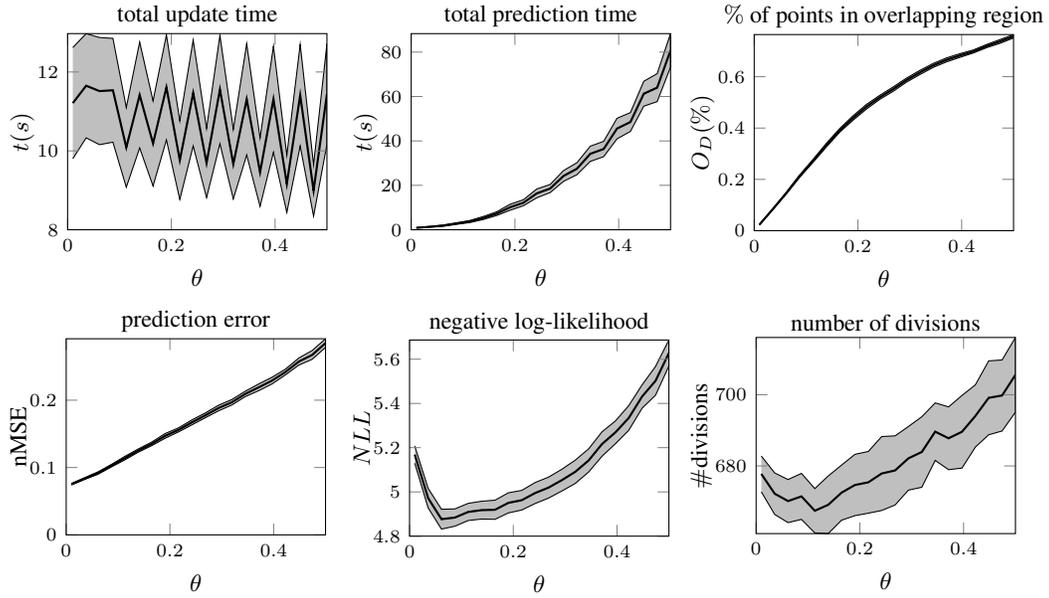

To summarize, the overlapping factor $\theta$ can have a great impact on the performance of predicting new incoming data. 
If $\theta$ is chosen sufficiently small, one can achieve logarithmic computational complexity for calculating
mean and variance prediction, while also resulting in a good regression performance.

\begin{figure}[p]
	\pgfplotsset{width=120\textwidth /100, compat = 1.13, 
		height =100\textwidth /100, grid= major, 
		legend cell align = left, ticklabel style = {font=\scriptsize},
		every axis label/.append style={font=\small},
		legend style = {font=\scriptsize},title style={yshift=0pt, font = \small},
		every x tick scale label/.style={at={(xticklabel cs:1)},anchor=south west} }
	\begin{minipage}{0.3\textwidth}
		\center
		\def\file{plots/pts_test.txt}
		\begin{tikzpicture}
		\begin{axis}[grid=none,enlargelimits=false, axis on top,
		xlabel={$\bar{N}$}, ylabel={$t(s)$}, 
		xmin=0, xmax = 700, ymin = 0,
		y label style={yshift=-0.2cm},
		title = {total update time},  tick label style={/pgf/number format/fixed}
		]
		\addplot[fill=black!25] table[x = pts_long, y = t_update_error ]{\file};
		\addplot[thick] table[x = pts, y = t_update_mean ]{\file};
		\end{axis}
		\end{tikzpicture}
	\end{minipage}	\hspace{0.2cm}
	\begin{minipage}{0.3\textwidth}
		\center
		\def\file{plots/pts_test.txt}
		\begin{tikzpicture}
		\begin{axis}[grid=none,enlargelimits=false, axis on top,
		xlabel={$\bar{N}$}, ylabel={$t(s)$},
		xmin=0, xmax = 700, ymin =0, ymax =9.5,	
		y label style={yshift=-0.2cm},
		title = {total prediction time}, tick label style={/pgf/number format/fixed}
		]
		\addplot[fill=black!25] table[x = pts_long, y = t_pred_error ]{\file};
		\addplot[thick] table[x = pts, y = t_pred_mean ]{\file};
		\end{axis}
		\end{tikzpicture}
	\end{minipage}\hspace{0.2cm}
	\begin{minipage}{0.3\textwidth}
		\center
		\def\file{plots/pts_test.txt}
		\begin{tikzpicture}
		\begin{axis}[grid=none,enlargelimits=false, axis on top,
		xlabel={$\bar{N}$}, ylabel={$O_D(\%)$},
		xmin=0, xmax = 700, ymin = 0, 
		y label style={yshift=-0.2cm},
		title = {$\%$ of points in overlapping region}, tick label style={/pgf/number format/fixed}
		]
		\addplot[fill=black!25] table[x = pts_long, y = overlapRatio_error ]{\file};
		\addplot[thick] table[x = pts, y = overlapRatio_mean ]{\file};
		\end{axis}
		\end{tikzpicture}
	\end{minipage}\\
	
	\begin{minipage}{0.3\textwidth}
		\def\file{plots/pts_test.txt}
		\begin{tikzpicture}
		\begin{axis}[grid=none,enlargelimits=false, axis on top,
		xlabel={$\bar{N}$}, ylabel={nMSE},
		xmin=0, xmax = 700, ymin =0.05,
		y label style={yshift=-0.2cm},
		title = {prediction error},  tick label style={/pgf/number format/fixed}
		]
		\addplot[fill=black!25] table[x = pts_long, y = error_error ]{\file};
		\addplot[thick] table[x = pts, y = error_mean ]{\file};
		\end{axis}
		\end{tikzpicture}
	\end{minipage}	\hspace{0.2cm}
	\begin{minipage}{0.3\textwidth}
		\center
		\def\file{plots/pts_test.txt}
		\begin{tikzpicture}
		\begin{axis}[grid=none,enlargelimits=false, axis on top,
		xlabel={$\bar{N}$}, ylabel={$NLL$},
		xmin=0, xmax = 700, ymin = 4,
		y label style={yshift=-0.2cm},
		title = {negative log-likelihood}, tick label style={/pgf/number format/fixed}
		]
		\addplot[fill=black!25] table[x = pts_long, y = nllMeans_error ]{\file};
		\addplot[thick] table[x = pts, y = nllMeans_mean ]{\file};
		\end{axis}
		\end{tikzpicture}
	\end{minipage}\hspace{0.2cm}
	\begin{minipage}{0.3\textwidth}
		\center
		\def\file{plots/pts_test.txt}
		\begin{tikzpicture}
		\begin{axis}[grid=none,enlargelimits=false, axis on top,
		xlabel={$\bar{N}$}, ylabel={$\#$divisions},
		xmin=0, xmax = 700, ymin = 80, ymax =1000,
		y label style={yshift=-0.2cm},
		title = {number of divisions}, tick label style={/pgf/number format/fixed}
		]
		\addplot[fill=black!25] table[x = pts_long, y = divCount_error ]{\file};
		\addplot[thick] table[x = pts, y = divCount_mean ]{\file};
		\end{axis}
		\end{tikzpicture}
	\end{minipage}
	\caption{Influence of the data point limit $\bar{N}$ on key performance measures.}
	\label{fig:pts_evaluation}
\end{figure}
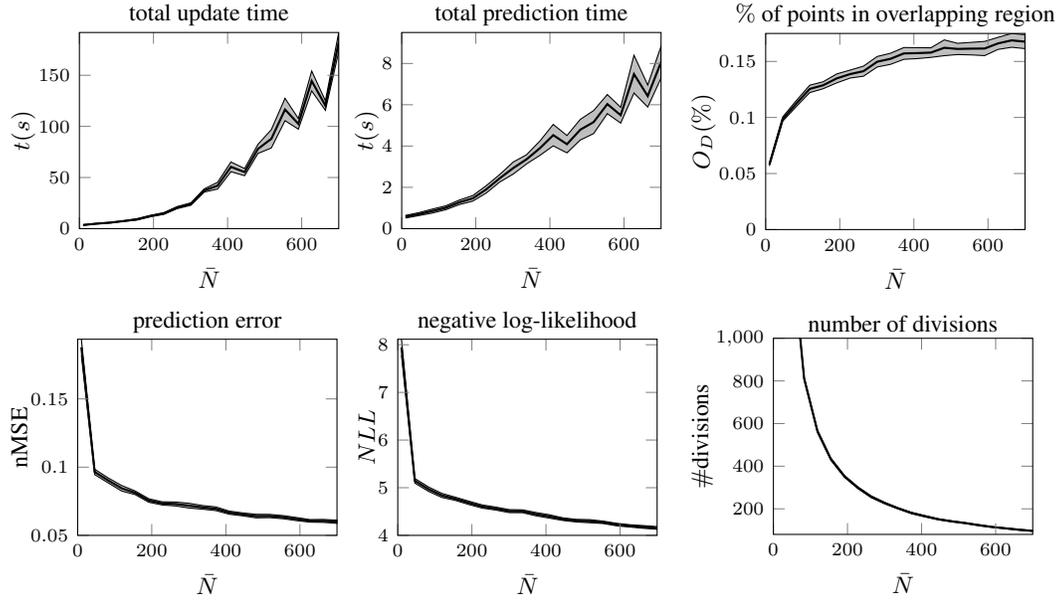

\begin{figure}[p]
	\pgfplotsset{width=85\textwidth /100, compat = 1.13, 
		height =74\textwidth /100, grid= major, 
		legend cell align = left, ticklabel style = {font=\scriptsize},
		every axis label/.append style={font=\small},
		legend style = {font=\scriptsize},title style={yshift=0pt, font = \small},
		every x tick scale label/.style={at={(xticklabel cs:1)},anchor=south west} }
	\begin{minipage}{0.48\textwidth}
		\center
		\def\file{plots/pts_examples.txt}
		\begin{tikzpicture}
		\begin{axis}[grid=none,enlargelimits=false, axis on top,
		xlabel={$N$}, ylabel={nMSE}, 
		xmin=0, xmax = 44484, 
		title = {error},  tick label style={/pgf/number format/fixed}
		]
		\addplot[blue!20, fill=blue!20] table[x = Ns_long, y = error_error_10  ]{\file};
		\addplot[black!20, fill=black!20] table[x = Ns_long, y = error_error_100  ]{\file};
		\addplot[green!20, fill=green!20] table[x = Ns_long, y = error_error_500  ]{\file};
		
		\addplot[blue] table[x = Ns, y = error_mean_10  ]{\file};
		\addplot[thick] table[x = Ns, y = error_mean_100  ]{\file};
		\addplot[green] table[x = Ns, y = error_mean_500  ]{\file};
		\end{axis}
		\end{tikzpicture}
	\end{minipage}	\hspace{-0.5cm}
	\begin{minipage}{0.48\textwidth}
		\center
		\def\file{plots/pts_examples.txt}
		\begin{tikzpicture}
		\begin{axis}[grid=none,enlargelimits=false, axis on top,
		xlabel={$N$}, ylabel={NLL}, 
		xmin=0, xmax = 44484,
		legend style={at={(1.02,0.5)},anchor=west},
		title = {negative log-likelihood},  tick label style={/pgf/number format/fixed}
		]
		\addplot[blue!20, fill=blue!20] table[x = Ns_long, y = nllMeans_error_10  ]{\file};
		\addplot[black!20, fill=black!20] table[x = Ns_long, y = nllMeans_error_100  ]{\file};
		\addplot[green!20, fill=green!20] table[x = Ns_long, y = nllMeans_error_500  ]{\file};
		
		\addplot[blue] table[x = Ns, y = nllMeans_mean_10  ]{\file};
		\addplot[thick] table[x = Ns, y =  nllMeans_mean_100  ]{\file};
		\addplot[green] table[x = Ns, y =  nllMeans_mean_500  ]{\file};
		\end{axis}
		\end{tikzpicture}
	\end{minipage}

	\begin{minipage}{0.48\textwidth}
		\center
		\def\file{plots/pts_examples.txt}
		\begin{tikzpicture}
		\begin{axis}[grid=none,enlargelimits=false, axis on top,
		xlabel={$N$}, ylabel={$t(s)$}, 
		xmin=0, xmax = 44484, 
		title = {average update time},  tick label style={/pgf/number format/fixed},
		every y tick scale label/.style={at={(yticklabel cs:1)},anchor=south west},
		]
		\addplot[blue!20, fill=blue!20] table[x = Ns_long, y = t_update_error_10  ]{\file};
		\addplot[black!20, fill=black!20] table[x = Ns_long, y = t_update_error_100  ]{\file};
		\addplot[green!20, fill=green!20] table[x = Ns_long, y = t_update_error_500  ]{\file};
		
		\addplot[blue] table[x = Ns, y = t_update_mean_10 ]{\file};
		\addplot[thick] table[x = Ns, y = t_update_mean_100 ]{\file};
		\addplot[green] table[x = Ns, y = t_update_mean_500 ]{\file};
		\end{axis}
		\end{tikzpicture}
	\end{minipage}	\hspace{-0.5cm}
	\begin{minipage}{0.48\textwidth}
		\center
		\def\file{plots/pts_examples.txt}
		\begin{tikzpicture}
		\begin{axis}[grid=none,enlargelimits=false, axis on top,
		xlabel={$N$}, ylabel={$t(s)$}, 
		xmin=0, xmax = 44484,
		legend style={at={(1.02,0.5)},anchor=west},
		title = {average prediction time}, tick label style={/pgf/number format/fixed},
		every y tick scale label/.style={at={(yticklabel cs:1)},anchor=south west},
		]
		\addplot[blue!20, fill=blue!20] table[x = Ns_long, y = t_pred_error_10  ]{\file};
		\addplot[black!20, fill=black!20] table[x = Ns_long, y = t_pred_error_100  ]{\file};
		\addplot[green!20, fill=green!20] table[x = Ns_long, y = t_pred_error_500  ]{\file};
		
		\addplot[blue] table[x = Ns, y = t_pred_mean_10 ]{\file};
		\addlegendentry{$\bar{N}=10$};
		\addplot[thick] table[x = Ns, y = t_pred_mean_100 ]{\file};
		\addlegendentry{$\bar{N}=100$};
		\addplot[green] table[x = Ns, y = t_pred_mean_500 ]{\file};
		\addlegendentry{$\bar{N}=500$};
		\end{axis}
		\end{tikzpicture}
	\end{minipage}
	\caption{Prediction error, NLL, update and prediction time for joint 1 through the learning process obtained with different
		values of $\bar{N}$.}
	\label{fig:times_pointlimit}
\end{figure}
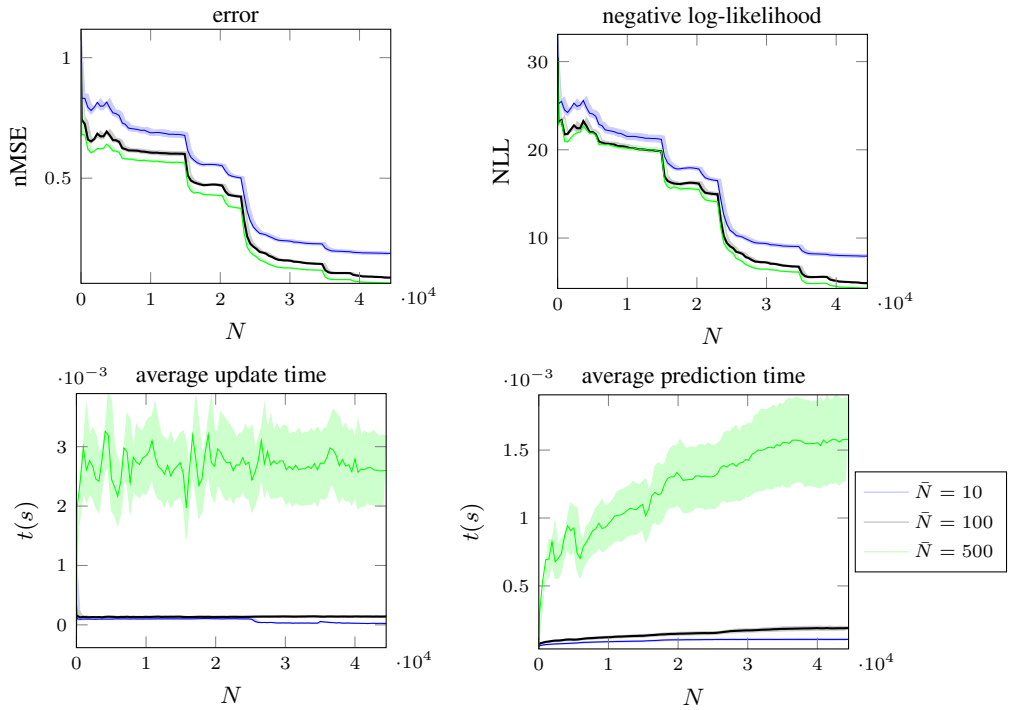

\subsection{Data limit}

The parameter for the data limit $\bar{N}$ determines the amount of data samples per local model at which the division into two
sub-models is performed. Going for a too high value of $\bar{N}$ results in local models requiring a high computational effort 
while a too small
value leads to a large number of local numbers typically causing a decreased regression performance.

For outlining the effects of $\bar{N}$ on the learning process, the impact on prediction error, NLL, average update and prediction times,
the ratio $O_D$ and the number of divisions is evaluated first as depicted in Figure~\ref{fig:pts_evaluation}. It indicates that
the update time is bounded by $\mathcal{O}(\log(N) + \bar{N}^3)$ as discussed in \cref{subsec:random trans}. 
Note that $N$ is constant in this setting. Therefore, for high values of $\bar{N}$, the number of local models 
is becoming small
such that the normally cubic growth is counterbalanced in part. The fact that increasing the data limit leads to fewer local models can 
be observed in the number of divisions, which is almost reciprocal to $\bar{N}$. This effect also influences the average prediction time, 
which initially grows quadratically in $\bar{N}$ causing a quickly growing total prediction time. In contrast to 
the negative effect on the computation times, high values of $\bar{N}$ are generally beneficial for the regression 
error and the quality of the predictive distributions. In fact, one can see the effect of a too small value of $\bar{N}$: 
If there are many local models, the
amount of data points per model is too small, such that no meaningful local model can be learned. Therefore,
a sufficiently high value of $\bar{N}$ is necessary to learn useful local models.

Additionally, Figure~\ref{fig:times_pointlimit} depicts the learning process for joint 1. One can clearly see the implications
of too small and too high data limits $\bar{N}$. First, $\bar{N} = 10$ leads to a significantly worse accuracy as the error and NLL are
constantly higher than for the remaining two simulations with higher values of $\bar{N}$. However, the timing behavior is better in 
accordance with
the findings in Figure~\ref{fig:pts_evaluation}. In contrast, while a much higher $\bar{N} = 500$ shows a good regression
performance, the increased computational effort also increases the average update and prediction times by an order of magnitude.
For $\bar{N} = 100$, one can leverage the advantages and avoid the drawbacks of both too low and too high $\bar{N}$ to some
extent. Both, the accuracy in the error and NLL as well as the average update and prediction times remain in a similar range as the
better performing parameter. Therefore, it stands to reason that $\bar{N}$ is a good choice for a trade-off between good
regression performance and reasonable computational effort.

\section{Additional Simulation Results}
\label{app:additional results}

This section provides the detailed results corresponding to both scenarios outlined in \cref{subsec:setup}. 
For each joint, the update times, prediction times, normalized mean squared error and average log-likelihood are depicted.
While prediction times for the SARCOS data set are determined using mean predictions only, both mean and variance
predictions are computed, when measuring the prediction time for the KUKA flask pushing data set.

\subsection{SARCOS Data Set}

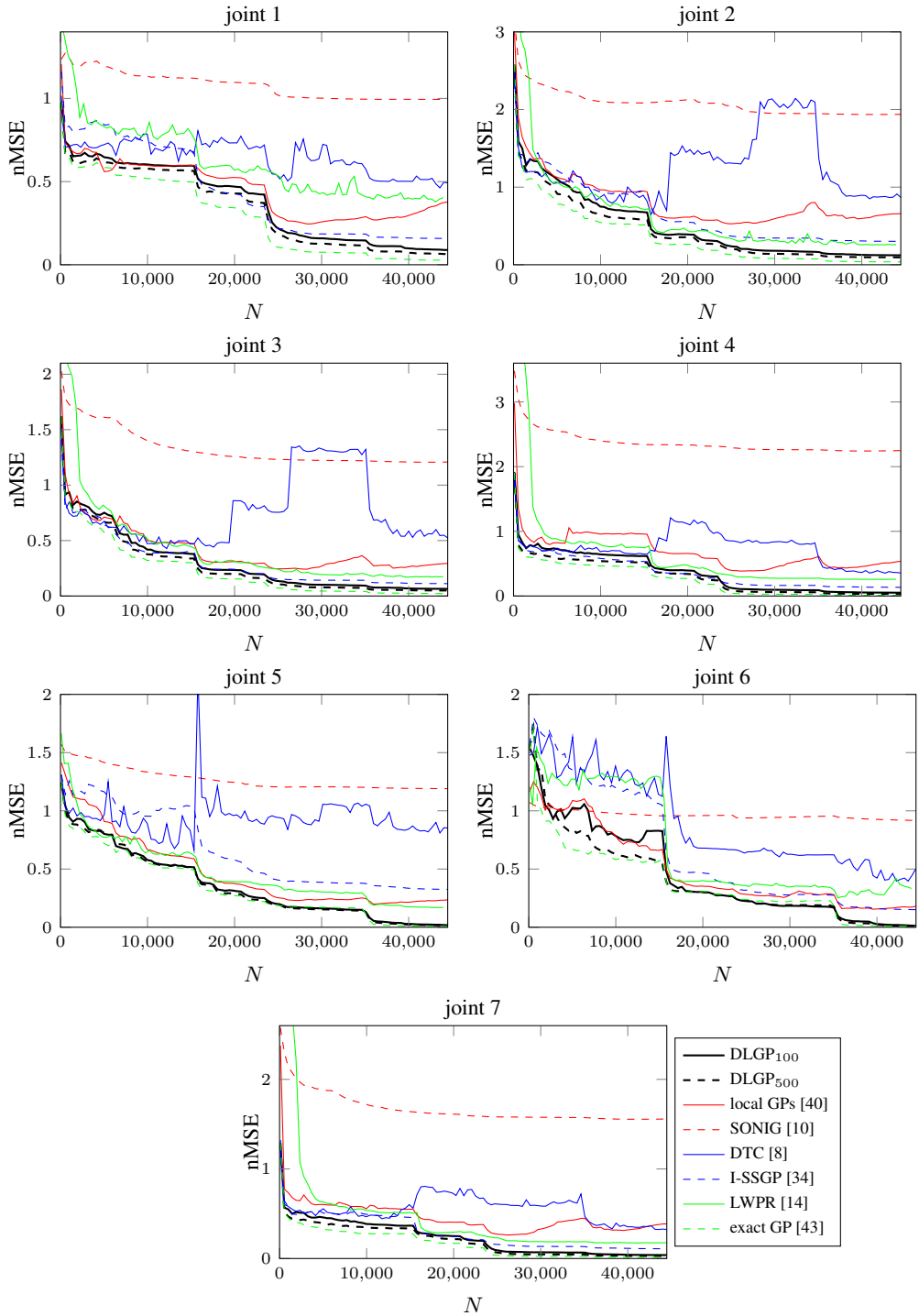
\begin{figure*}[p]
	\pgfplotsset{width=7.2cm, compat = 1.13, 
		height =74\textwidth /100, grid= major, 
		legend cell align = left, ticklabel style = {font=\scriptsize},
		every axis label/.append style={font=\small},
		legend style = {font=\scriptsize},title style={yshift=0pt, font = \small},
		every x tick scale label/.style={at={(xticklabel cs:1)},anchor=south west} }
	
	\center
	
	\begin{minipage}{0.48\textwidth}
		\def\file{plots/Real_Sarcos_long_plot_data.txt}
		\begin{tikzpicture}
		\begin{axis}[grid=none,enlargelimits=false, axis on top,
		xlabel={$N$}, ylabel={nMSE}, 
		xmin=1, xmax = 44484, ymin = 0, ymax =1.4,	
		y label style={yshift=-0.2cm},
		title = {joint 1},
		scaled ticks=false, tick label style={/pgf/number format/fixed}
		]
		\addplot[thick] table[x = steps_DLGP, y = e0_DLGP ]{\file};
		\addplot[thick, dashed] table[x = steps_DLGP500, y = e0_DLGP500 ]{\file};
		\addplot[red] table[x = steps_LGP, y = e0_LGP ]{\file};
		\addplot[red,dashed] table[x = steps_IP, y = e0_IP ]{\file};
		\addplot[blue] table[x = steps_Sparse,y = e0_Sparse ]{\file};
		\addplot[blue,dashed] table[x = steps_ISSGP,y = e0_ISSGP ]{\file};
		\addplot[green] table[x = steps_lwpr,y = e0_lwpr ]{\file};
		\addplot[green,dashed] table[x = steps_DLGP,y = e0_GPexact ]{\file};
		\end{axis}
		\end{tikzpicture}
	\end{minipage}
	\begin{minipage}{0.48\textwidth}
		\def\file{plots/Real_Sarcos_long_plot_data.txt}
		\begin{tikzpicture}
		\begin{axis}[grid=none,enlargelimits=false, axis on top,
		xlabel={$N$}, ylabel={nMSE},
		xmin=1, xmax = 44484, ymin = 0, ymax =3,
		y label style={yshift=-0.2cm},
		title = {joint 2},
		scaled ticks=false, tick label style={/pgf/number format/fixed}
		]
		\addplot[thick] table[x = steps_DLGP, y = e1_DLGP ]{\file};
		\addplot[thick, dashed] table[x = steps_DLGP500, y = e1_DLGP500 ]{\file};
		\addplot[red] table[x = steps_LGP, y = e1_LGP ]{\file};
		\addplot[red,dashed] table[x = steps_IP, y = e1_IP ]{\file};
		\addplot[blue] table[x = steps_Sparse,y = e1_Sparse ]{\file};
		\addplot[blue,dashed] table[x = steps_ISSGP,y = e1_ISSGP ]{\file};
		\addplot[green] table[x = steps_lwpr,y = e1_lwpr ]{\file};
		\addplot[green,dashed] table[x = steps_DLGP,y = e1_GPexact ]{\file};
		\end{axis}
		\end{tikzpicture}
	\end{minipage}
	
	\begin{minipage}{0.48\textwidth}
		\def\file{plots/Real_Sarcos_long_plot_data.txt}
		\begin{tikzpicture}
		\begin{axis}[grid=none,enlargelimits=false, axis on top,
		xlabel={$N$}, ylabel={nMSE}, 
		xmin=1, xmax = 44484, ymin = 0, ymax =2.1,	
		y label style={yshift=-0.2cm},
		title = {joint 3},
		scaled ticks=false, tick label style={/pgf/number format/fixed}
		]
		\addplot[thick] table[x = steps_DLGP, y = e2_DLGP ]{\file};
		\addplot[thick, dashed] table[x = steps_DLGP500, y = e2_DLGP500 ]{\file};
		\addplot[red] table[x = steps_LGP, y = e2_LGP ]{\file};
		\addplot[red,dashed] table[x = steps_IP, y = e2_IP ]{\file};
		\addplot[blue] table[x = steps_Sparse,y = e2_Sparse ]{\file};
		\addplot[blue,dashed] table[x = steps_ISSGP,y = e2_ISSGP ]{\file};
		\addplot[green] table[x = steps_lwpr,y = e2_lwpr ]{\file};
		\addplot[green,dashed] table[x = steps_DLGP,y = e2_GPexact ]{\file};
		\end{axis}
		\end{tikzpicture}
	\end{minipage}
	\begin{minipage}{0.48\textwidth}
		\def\file{plots/Real_Sarcos_long_plot_data.txt}
		\begin{tikzpicture}
		\begin{axis}[grid=none,enlargelimits=false, axis on top,
		xlabel={$N$}, ylabel={nMSE}, 
		xmin=1, xmax = 44484, ymin = 0, ymax =3.6,
		y label style={yshift=-0.2cm},
		title = {joint 4},
		scaled ticks=false, tick label style={/pgf/number format/fixed}
		]
		\addplot[thick] table[x = steps_DLGP, y = e3_DLGP ]{\file};
		\addplot[thick, dashed] table[x = steps_DLGP500, y = e3_DLGP500 ]{\file};
		\addplot[red] table[x = steps_LGP, y = e3_LGP ]{\file};
		\addplot[red,dashed] table[x = steps_IP, y = e3_IP ]{\file};
		\addplot[blue] table[x = steps_Sparse,y = e3_Sparse ]{\file};
		\addplot[blue,dashed] table[x = steps_ISSGP,y = e3_ISSGP ]{\file};
		\addplot[green] table[x = steps_lwpr,y = e3_lwpr ]{\file};
		\addplot[green,dashed] table[x = steps_DLGP,y = e3_GPexact ]{\file};
		\end{axis}
		\end{tikzpicture}
	\end{minipage}
	
	\begin{minipage}{0.48\textwidth}
		\def\file{plots/Real_Sarcos_long_plot_data.txt}
		\begin{tikzpicture}
		\begin{axis}[grid=none,enlargelimits=false, axis on top,
		xlabel={$N$}, ylabel={nMSE}, 
		xmin=1, xmax = 44484, ymin = 0, ymax =2,
		y label style={yshift=-0.2cm},
		title = {joint 5},
		scaled ticks=false, tick label style={/pgf/number format/fixed}
		]
		\addplot[thick] table[x = steps_DLGP, y = e4_DLGP ]{\file};
		\addplot[thick, dashed] table[x = steps_DLGP500, y = e4_DLGP500 ]{\file};
		\addplot[red] table[x = steps_LGP, y = e4_LGP ]{\file};
		\addplot[red,dashed] table[x = steps_IP, y = e4_IP ]{\file};
		\addplot[blue] table[x = steps_Sparse,y = e4_Sparse ]{\file};
		\addplot[blue,dashed] table[x = steps_ISSGP,y = e4_ISSGP ]{\file};
		\addplot[green] table[x = steps_lwpr,y = e4_lwpr ]{\file};
		\addplot[green,dashed] table[x = steps_DLGP,y = e4_GPexact ]{\file};
		\end{axis}
		\end{tikzpicture}
	\end{minipage}
	\begin{minipage}{0.48\textwidth}
		\def\file{plots/Real_Sarcos_long_plot_data.txt}
		\begin{tikzpicture}
		\begin{axis}[grid=none,enlargelimits=false, axis on top,
		xlabel={$N$}, ylabel={nMSE},
		xmin=1, xmax = 44484, ymin = 0, ymax =2,
		y label style={yshift=-0.2cm},
		title = {joint 6},
		scaled ticks=false, tick label style={/pgf/number format/fixed}
		]
		\addplot[thick] table[x = steps_DLGP, y = e5_DLGP ]{\file};
		\addplot[thick, dashed] table[x = steps_DLGP500, y = e5_DLGP500 ]{\file};
		\addplot[red] table[x = steps_LGP, y = e5_LGP ]{\file};
		\addplot[red,dashed] table[x = steps_IP, y = e5_IP ]{\file};
		\addplot[blue] table[x = steps_Sparse,y = e5_Sparse ]{\file};
		\addplot[blue,dashed] table[x = steps_ISSGP,y = e5_ISSGP ]{\file};
		\addplot[green] table[x = steps_lwpr,y = e5_lwpr ]{\file};
		\addplot[green,dashed] table[x = steps_DLGP,y = e5_GPexact ]{\file};
		\end{axis}
		\end{tikzpicture}
	\end{minipage}

	\begin{minipage}{0.48\textwidth}
		\def\file{plots/Real_Sarcos_long_plot_data.txt}
		\begin{tikzpicture}
		\begin{axis}[grid=none,enlargelimits=false, axis on top,
		xlabel={$N$}, ylabel={nMSE},
		xmin=1, xmax = 44484, ymin = 0, ymax = 2.6,	
		y label style={yshift=-0.2cm},
		legend style={at={(1.02,0.5)},anchor=west},
		title = {joint 7},
		scaled ticks=false, tick label style={/pgf/number format/fixed}	]
		\addplot[thick] table[x = steps_Sparse, y = e6_DLGP  ]{\file};
		\addlegendentry{DLGP$_{100}$}
		\addplot[thick, dashed] table[x = steps_Sparse, y = e6_DLGP500  ]{\file};
		\addlegendentry{DLGP$_{500}$}
		\addplot[red] table[x = steps_Sparse, y = e6_LGP ]{\file};
		\addlegendentry{local GPs \cite{Nguyen-Tuong2009}}
		\addplot[red,dashed] table[x = steps_Sparse, y =e6_IP  ]{\file};
		\addlegendentry{SONIG \cite{Bijl2017}}
		\addplot[blue] table[x = steps_Sparse, y = e6_Sparse  ]{\file};
		\addlegendentry{DTC \cite{Schreiter2016}}
		\addplot[blue,dashed] table[x = steps_Sparse, y = e6_ISSGP  ]{\file};
		\addlegendentry{I-SSGP \cite{Gijsberts2013}}
		\addplot[green] table[x = steps_Sparse,y = e6_lwpr  ]{\file};
		\addlegendentry{LWPR \cite{Vijayakumar2000}}
		\addplot[green,dashed] table[x = steps_Sparse, y = e6_GPexact ]{\file};
		\addlegendentry{exact GP  \cite{Wang2019a}}
		\end{axis}
		\end{tikzpicture}
	\end{minipage}	
	\caption{Normalized mean squared errors depending on the number of training samples $N$ for 
		all joints of the SARCOS data set: DLGPs consistently outperform other online regression methods and exhibit identical
		learning behavior as exact GP regression. }
	\label{fig:nMSE_Sarcos}
\end{figure*}

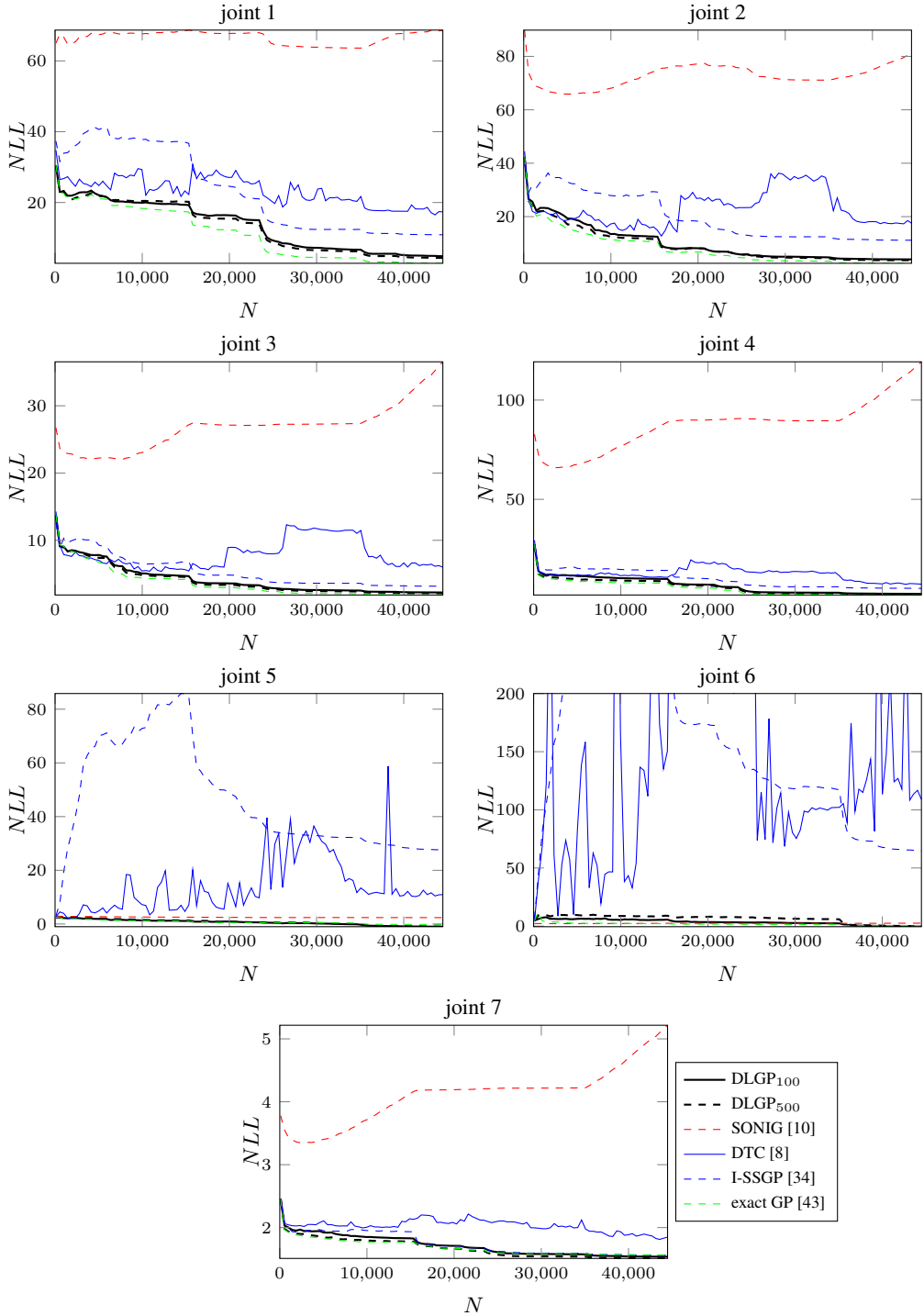
\begin{figure*}[p]
	\pgfplotsset{width=7.2cm, compat = 1.13, 
		height =74\textwidth /100, grid= major, 
		legend cell align = left, ticklabel style = {font=\scriptsize},
		every axis label/.append style={font=\small},
		legend style = {font=\scriptsize},title style={yshift=0pt, font = \small},
		every x tick scale label/.style={at={(xticklabel cs:1)},anchor=south west} }
	
	\center
	
	\begin{minipage}{0.48\textwidth}
		\def\file{plots/Real_Sarcos_longNLL_plot_data.txt}
		\begin{tikzpicture}
		\begin{axis}[grid=none,enlargelimits=false, axis on top,
		xlabel={$N$}, ylabel={$NLL$}, 
		xmin=1, xmax = 44484,
		y label style={yshift=-0.2cm},
		title = {joint 1},
		scaled ticks=false, tick label style={/pgf/number format/fixed}
		]
		\addplot[thick] table[x = steps_IP, y = aNLL0_DLGP ]{\file};
		\addplot[thick, dashed] table[x = steps_IP, y = aNLL0_DLGP500 ]{\file};
		\addplot[red,dashed] table[x = steps_IP, y = aNLL0_IP ]{\file};
		\addplot[blue] table[x = steps_Sparse,y = aNLL0_Sparse ]{\file};
		\addplot[blue,dashed] table[x = steps_ISSGP,y = aNLL0_ISSGP ]{\file};
		\addplot[green,dashed] table[x = steps_IP,y = aNLL0_GPexact ]{\file};
		\end{axis}
		\end{tikzpicture}
	\end{minipage}
	\begin{minipage}{0.48\textwidth}
		\def\file{plots/Real_Sarcos_longNLL_plot_data.txt}
		\begin{tikzpicture}
		\begin{axis}[grid=none,enlargelimits=false, axis on top,
		xlabel={$N$}, ylabel={$NLL$},
		xmin=1, xmax = 44484,
		y label style={yshift=-0.2cm},
		title = {joint 2},
		scaled ticks=false, tick label style={/pgf/number format/fixed}
		]
		\addplot[thick] table[x = steps_IP, y = aNLL1_DLGP ]{\file};
		\addplot[thick, dashed] table[x = steps_IP, y = aNLL1_DLGP500 ]{\file};
		\addplot[red,dashed] table[x = steps_IP, y = aNLL1_IP ]{\file};
		\addplot[blue] table[x = steps_Sparse,y = aNLL1_Sparse ]{\file};
		\addplot[blue,dashed] table[x = steps_ISSGP,y = aNLL1_ISSGP ]{\file};
		\addplot[green,dashed] table[x = steps_IP,y = aNLL1_GPexact ]{\file};
		\end{axis}
		\end{tikzpicture}
	\end{minipage}
	
	\begin{minipage}{0.48\textwidth}
		\def\file{plots/Real_Sarcos_longNLL_plot_data.txt}
		\begin{tikzpicture}
		\begin{axis}[grid=none,enlargelimits=false, axis on top,
		xlabel={$N$}, ylabel={$NLL$}, 
		xmin=1, xmax = 44484, 
		y label style={yshift=-0.2cm},
		title = {joint 3},
		scaled ticks=false, tick label style={/pgf/number format/fixed}
		]
		\addplot[thick] table[x = steps_IP, y = aNLL2_DLGP ]{\file};
		\addplot[thick, dashed] table[x = steps_IP, y = aNLL2_DLGP500 ]{\file};
		\addplot[red,dashed] table[x = steps_IP, y = aNLL2_IP ]{\file};
		\addplot[blue] table[x = steps_Sparse,y = aNLL2_Sparse ]{\file};
		\addplot[blue,dashed] table[x = steps_ISSGP,y = aNLL2_ISSGP ]{\file};
		\addplot[green,dashed] table[x = steps_IP,y = aNLL2_GPexact ]{\file};
		\end{axis}
		\end{tikzpicture}
	\end{minipage}
	\begin{minipage}{0.48\textwidth}
		\def\file{plots/Real_Sarcos_longNLL_plot_data.txt}
		\begin{tikzpicture}
		\begin{axis}[grid=none,enlargelimits=false, axis on top,
		xlabel={$N$}, ylabel={$NLL$},
		xmin=1, xmax = 44484,
		y label style={yshift=-0.2cm},
		title = {joint 4},
		scaled ticks=false, tick label style={/pgf/number format/fixed}
		]
		\addplot[thick] table[x = steps_IP, y = aNLL3_DLGP ]{\file};
		\addplot[thick, dashed] table[x = steps_IP, y = aNLL3_DLGP500 ]{\file};
		\addplot[red,dashed] table[x = steps_IP, y = aNLL3_IP ]{\file};
		\addplot[blue] table[x = steps_Sparse,y = aNLL3_Sparse ]{\file};
		\addplot[blue,dashed] table[x = steps_ISSGP,y = aNLL3_ISSGP ]{\file};
		\addplot[green,dashed] table[x = steps_IP,y = aNLL3_GPexact ]{\file};
		\end{axis}
		\end{tikzpicture}
	\end{minipage}
	
	\begin{minipage}{0.48\textwidth}
		\def\file{plots/Real_Sarcos_longNLL_plot_data.txt}
		\begin{tikzpicture}
		\begin{axis}[grid=none,enlargelimits=false, axis on top,
		xlabel={$N$}, ylabel={$NLL$}, 
		xmin=1, xmax = 44484, 
		y label style={yshift=-0.2cm},
		title = {joint 5},
		scaled ticks=false, tick label style={/pgf/number format/fixed}
		]
		\addplot[thick] table[x = steps_IP, y = aNLL4_DLGP ]{\file};
		\addplot[thick, dashed] table[x = steps_IP, y = aNLL4_DLGP500 ]{\file};
		\addplot[red,dashed] table[x = steps_IP, y = aNLL4_IP ]{\file};
		\addplot[blue] table[x = steps_Sparse,y = aNLL4_Sparse ]{\file};
		\addplot[blue,dashed] table[x = steps_ISSGP,y = aNLL4_ISSGP ]{\file};
		\addplot[green,dashed] table[x = steps_IP,y = aNLL4_GPexact ]{\file};
		\end{axis}
		\end{tikzpicture}
	\end{minipage}
	\begin{minipage}{0.48\textwidth}
		\def\file{plots/Real_Sarcos_longNLL_plot_data.txt}
		\begin{tikzpicture}
		\begin{axis}[grid=none,enlargelimits=false, axis on top,
		xlabel={$N$}, ylabel={$NLL$}, 
		xmin=1, xmax = 44484, ymin = -0.5, ymax =200,
		y label style={yshift=-0.2cm},
		title = {joint 6},
		scaled ticks=false, tick label style={/pgf/number format/fixed}
		]
		\addplot[thick] table[x = steps_IP, y = aNLL5_DLGP ]{\file};
		\addplot[thick, dashed] table[x = steps_IP, y = aNLL5_DLGP500 ]{\file};
		\addplot[red,dashed] table[x = steps_IP, y = aNLL5_IP ]{\file};
		\addplot[blue] table[x = steps_Sparse,y = aNLL5_Sparse ]{\file};
		\addplot[blue,dashed] table[x = steps_ISSGP,y = aNLL5_ISSGP ]{\file};
		\addplot[green,dashed] table[x = steps_IP, y = aNLL5_GPexact ]{\file};
		\end{axis}
		\end{tikzpicture}
	\end{minipage}

	\begin{minipage}{0.48\textwidth}
		\def\file{plots/Real_Sarcos_longNLL_plot_data.txt}
		\begin{tikzpicture}
		\begin{axis}[grid=none,enlargelimits=false, axis on top,
		xlabel={$N$}, ylabel={$NLL$}, 
		xmin=1, xmax = 44484,
		y label style={yshift=-0.2cm},
		legend style={at={(1.02,0.5)},anchor=west},
		title = {joint 7},
		scaled ticks=false, tick label style={/pgf/number format/fixed}	]
		\addplot[thick] table[x = steps_Sparse, y = aNLL6_DLGP  ]{\file};
		\addlegendentry{DLGP$_{100}$}
		\addplot[thick, dashed] table[x = steps_Sparse, y = aNLL6_DLGP500  ]{\file};
		\addlegendentry{DLGP$_{500}$}
		\addplot[red,dashed] table[x = steps_Sparse, y =aNLL6_IP  ]{\file};
		\addlegendentry{SONIG \cite{Bijl2017}}
		\addplot[blue] table[x = steps_Sparse, y = aNLL6_Sparse  ]{\file};
		\addlegendentry{DTC \cite{Schreiter2016}}
		\addplot[blue,dashed] table[x = steps_Sparse, y = aNLL6_ISSGP  ]{\file};
		\addlegendentry{I-SSGP \cite{Gijsberts2013}}
		\addplot[green,dashed] table[x = steps_Sparse, y = aNLL6_GPexact ]{\file};
		\addlegendentry{exact GP  \cite{Wang2019a}}
		\end{axis}
		\end{tikzpicture}
	\end{minipage}
	
	\caption{Average negative log-likelihood depending on the number of training samples $N$ for all joints of the SARCOS data set: 
		while some methods achieve reliable predictive distributions only for certain joints, DLGPs exhibit a performance similar to exact GPs.}
	\label{fig:NLL_Sarcos}
\end{figure*}

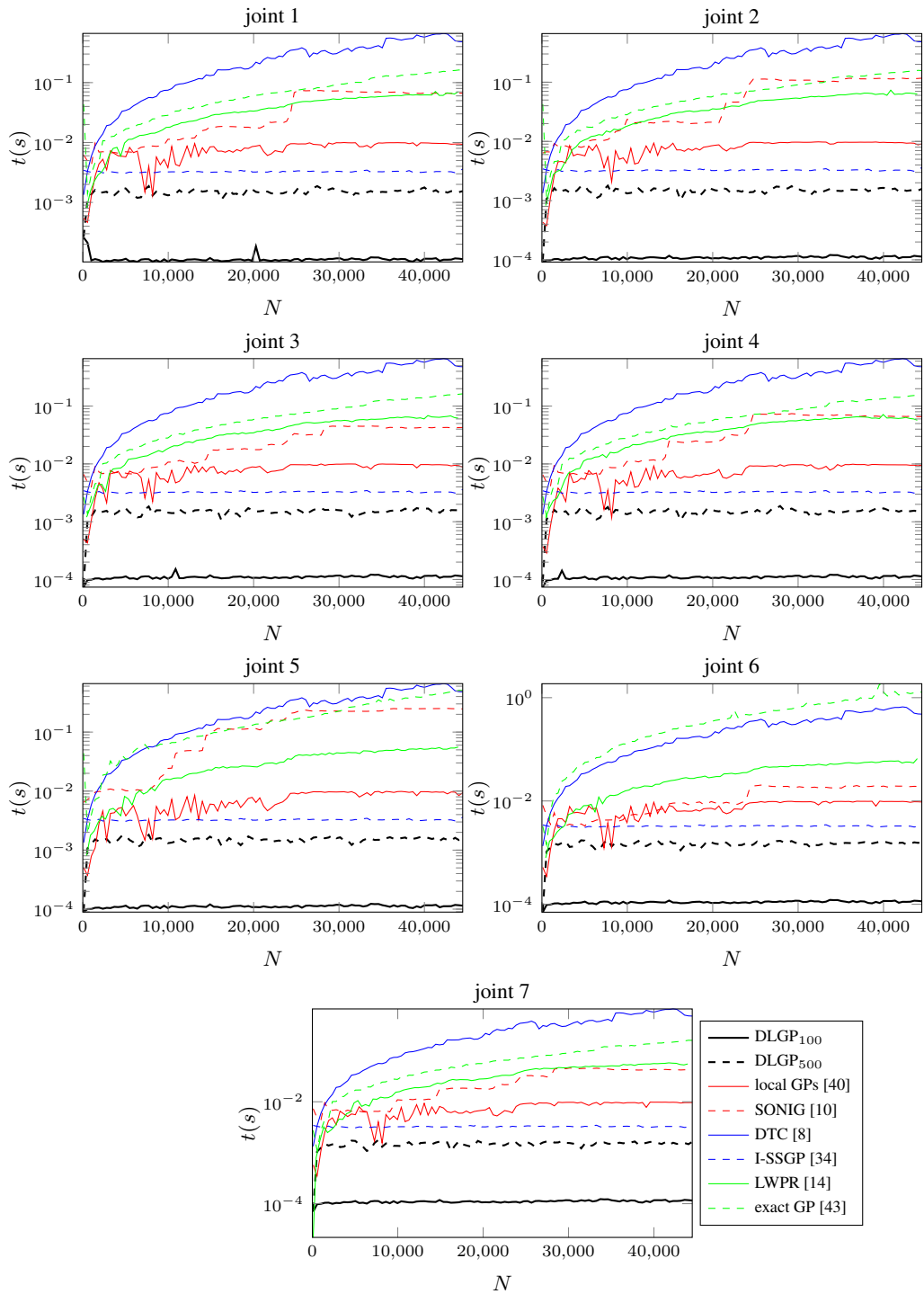
\begin{figure*}[p]
	\pgfplotsset{width=7.2cm, compat = 1.13, 
		height =74\textwidth /100, grid= major, 
		legend cell align = left, ticklabel style = {font=\scriptsize},
		every axis label/.append style={font=\small},
		legend style = {font=\scriptsize},title style={yshift=0pt, font = \small},
		every x tick scale label/.style={at={(xticklabel cs:1)},anchor=south west} }
	
	\center
	
	\begin{minipage}{0.48\textwidth}
		\def\file{plots/Real_Sarcos_long_plot_data.txt}
		\begin{tikzpicture}
		\begin{semilogyaxis}[grid=none,enlargelimits=false, axis on top,
		xlabel={$N$}, ylabel={$t(s)$}, 
		xmin=1, xmax = 44484,
		y label style={yshift=-0.2cm},
		title = {joint 1},
		scaled ticks=false, tick label style={/pgf/number format/fixed}
		]
		\addplot[thick] table[x = steps_IP, y = t_update0_DLGP ]{\file};
		\addplot[thick, dashed] table[x = steps_IP, y = t_update0_DLGP500 ]{\file};
		\addplot[red] table[x = steps_LGP, y =t_update0_LGP ]{\file};
		\addplot[red,dashed] table[x = steps_IP, y = t_update0_IP ]{\file};
		\addplot[blue] table[x = steps_Sparse,y = t_update0_Sparse ]{\file};
		\addplot[blue,dashed] table[x = steps_ISSGP,y = t_update0_ISSGP ]{\file};
		\addplot[green] table[x = steps_lwpr,y = t_update0_lwpr ]{\file};
		\addplot[green,dashed] table[x = steps_IP,y = t_update0_GPexact ]{\file};
		\end{semilogyaxis}
		\end{tikzpicture}
	\end{minipage}
	\begin{minipage}{0.48\textwidth}
		\def\file{plots/Real_Sarcos_long_plot_data.txt}
		\begin{tikzpicture}
		\begin{semilogyaxis}[grid=none,enlargelimits=false, axis on top,
		xlabel={$N$}, ylabel={$t(s)$},
		xmin=1, xmax = 44484, 
		y label style={yshift=-0.2cm},
		title = {joint 2},
		scaled ticks=false, tick label style={/pgf/number format/fixed}
		]
		\addplot[thick] table[x = steps_IP, y = t_update1_DLGP ]{\file};
		\addplot[thick, dashed] table[x = steps_IP, y = t_update1_DLGP500 ]{\file};
		\addplot[red] table[x = steps_LGP, y =t_update1_LGP ]{\file};
		\addplot[red,dashed] table[x = steps_IP, y = t_update1_IP ]{\file};
		\addplot[blue] table[x = steps_Sparse,y = t_update1_Sparse ]{\file};
		\addplot[blue,dashed] table[x = steps_ISSGP,y = t_update1_ISSGP ]{\file};
		\addplot[green] table[x = steps_lwpr,y = t_update1_lwpr ]{\file};
		\addplot[green,dashed] table[x = steps_IP,y = t_update1_GPexact ]{\file};
		\end{semilogyaxis}
		\end{tikzpicture}
	\end{minipage}
	
	\begin{minipage}{0.48\textwidth}
		\def\file{plots/Real_Sarcos_long_plot_data.txt}
		\begin{tikzpicture}
		\begin{semilogyaxis}[grid=none,enlargelimits=false, axis on top,
		xlabel={$N$}, ylabel={$t(s)$}, 
		xmin=1, xmax = 44484, 
		y label style={yshift=-0.2cm},
		title = {joint 3},
		scaled ticks=false, tick label style={/pgf/number format/fixed}
		]
		\addplot[thick] table[x = steps_IP, y = t_update2_DLGP ]{\file};
		\addplot[thick, dashed] table[x = steps_IP, y = t_update2_DLGP500 ]{\file};
		\addplot[red] table[x = steps_LGP, y = t_update2_LGP ]{\file};
		\addplot[red,dashed] table[x = steps_IP, y = t_update2_IP ]{\file};
		\addplot[blue] table[x = steps_Sparse,y = t_update2_Sparse ]{\file};
		\addplot[blue,dashed] table[x = steps_ISSGP,y = t_update2_ISSGP ]{\file};
		\addplot[green] table[x = steps_lwpr,y = t_update2_lwpr ]{\file};
		\addplot[green,dashed] table[x = steps_IP,y = t_update2_GPexact ]{\file};
		\end{semilogyaxis}
		\end{tikzpicture}
	\end{minipage}
	\begin{minipage}{0.48\textwidth}
		\def\file{plots/Real_Sarcos_long_plot_data.txt}
		\begin{tikzpicture}
		\begin{semilogyaxis}[grid=none,enlargelimits=false, axis on top,
		xlabel={$N$}, ylabel={$t(s)$}, 
		xmin=1, xmax = 44484, 
		y label style={yshift=-0.2cm},
		title = {joint 4},
		scaled ticks=false, tick label style={/pgf/number format/fixed}
		]
		\addplot[thick] table[x = steps_IP, y = t_update3_DLGP ]{\file};
		\addplot[thick, dashed] table[x = steps_IP, y = t_update3_DLGP500 ]{\file};
		\addplot[red] table[x = steps_LGP, y = t_update3_LGP ]{\file};
		\addplot[red,dashed] table[x = steps_IP, y = t_update3_IP ]{\file};
		\addplot[blue] table[x = steps_Sparse,y = t_update3_Sparse ]{\file};
		\addplot[blue,dashed] table[x = steps_ISSGP,y = t_update3_ISSGP ]{\file};
		\addplot[green] table[x = steps_lwpr,y = t_update3_lwpr ]{\file};
		\addplot[green,dashed] table[x = steps_IP,y = t_update3_GPexact ]{\file};
		\end{semilogyaxis}
		\end{tikzpicture}
	\end{minipage}
	
	\begin{minipage}{0.48\textwidth}
		\def\file{plots/Real_Sarcos_long_plot_data.txt}
		\begin{tikzpicture}
		\begin{semilogyaxis}[grid=none,enlargelimits=false, axis on top,
		xlabel={$N$}, ylabel={$t(s)$}, 
		xmin=1, xmax = 44484, 
		y label style={yshift=-0.2cm},
		title = {joint 5},
		scaled ticks=false, tick label style={/pgf/number format/fixed}
		]
		\addplot[thick] table[x = steps_IP, y = t_update4_DLGP ]{\file};
		\addplot[thick, dashed] table[x = steps_IP, y = t_update4_DLGP500 ]{\file};
		\addplot[red] table[x = steps_LGP, y = t_update4_LGP ]{\file};
		\addplot[red,dashed] table[x = steps_IP, y = t_update4_IP ]{\file};
		\addplot[blue] table[x = steps_Sparse,y = t_update4_Sparse ]{\file};
		\addplot[blue,dashed] table[x = steps_ISSGP,y = t_update4_ISSGP ]{\file};
		\addplot[green] table[x = steps_lwpr,y = t_update4_lwpr ]{\file};
		\addplot[green,dashed] table[x = steps_IP,y = t_update4_GPexact ]{\file};
		\end{semilogyaxis}
		\end{tikzpicture}
	\end{minipage}
	\begin{minipage}{0.48\textwidth}
		\def\file{plots/Real_Sarcos_long_plot_data.txt}
		\begin{tikzpicture}
		\begin{semilogyaxis}[grid=none,enlargelimits=false, axis on top,
		xlabel={$N$}, ylabel={$t(s)$}, 
		xmin=1, xmax = 44484,
		y label style={yshift=-0.2cm},
		title = {joint 6},
		scaled ticks=false, tick label style={/pgf/number format/fixed}
		]
		\addplot[thick] table[x = steps_IP, y =t_update5_DLGP ]{\file};
		\addplot[thick, dashed] table[x = steps_IP, y = t_update5_DLGP500 ]{\file};
		\addplot[red] table[x = steps_LGP, y = t_update5_LGP ]{\file};
		\addplot[red,dashed] table[x = steps_IP, y = t_update5_IP ]{\file};
		\addplot[blue] table[x = steps_Sparse,y = t_update5_Sparse ]{\file};
		\addplot[blue,dashed] table[x = steps_ISSGP,y = t_update5_ISSGP ]{\file};
		\addplot[green] table[x = steps_lwpr,y = t_update5_lwpr ]{\file};
		\addplot[green, dashed] table[x = steps_IP, y = t_update5_GPexact ]{\file};
		\end{semilogyaxis}
		\end{tikzpicture}
	\end{minipage}

	\begin{minipage}{0.48\textwidth}
		\def\file{plots/Real_Sarcos_long_plot_data.txt}
		\begin{tikzpicture}
		\begin{semilogyaxis}[grid=none,enlargelimits=false, axis on top,
		xlabel={$N$}, ylabel={$t(s)$},
		xmin=1, xmax = 44484,
		y label style={yshift=-0.2cm},
		legend style={at={(1.02,0.5)},anchor=west},
		title = {joint 7},
		scaled ticks=false, tick label style={/pgf/number format/fixed}	]
		\addplot[thick] table[x = steps_Sparse, y = t_update6_DLGP  ]{\file};
		\addlegendentry{DLGP$_{100}$}
		\addplot[thick, dashed] table[x = steps_Sparse, y = t_update6_DLGP500  ]{\file};
		\addlegendentry{DLGP$_{500}$}
		\addplot[red] table[x = steps_Sparse, y = t_update6_LGP ]{\file};
		\addlegendentry{local GPs \cite{Nguyen-Tuong2009}}
		\addplot[red,dashed] table[x = steps_Sparse, y =t_update6_IP  ]{\file};
		\addlegendentry{SONIG \cite{Bijl2017}}
		\addplot[blue] table[x = steps_Sparse, y = t_update6_Sparse  ]{\file};
		\addlegendentry{DTC \cite{Schreiter2016}}
		\addplot[blue,dashed] table[x = steps_Sparse, y = t_update6_ISSGP  ]{\file};
		\addlegendentry{I-SSGP \cite{Gijsberts2013}}
		\addplot[green] table[x = steps_lwpr,y = t_update6_lwpr  ]{\file};
		\addlegendentry{LWPR \cite{Vijayakumar2000}}
		\addplot[green,dashed] table[x = steps_Sparse, y = t_update6_GPexact ]{\file};
		\addlegendentry{exact GP  \cite{Wang2019a}}
		\end{semilogyaxis}
		\end{tikzpicture}
	\end{minipage}
	
	\caption{Average time necessary to update a model depending on the number of training samples $N$ for all joints 
		of the SARCOS data set: in contrast to other methods, the computational complexity of DLGPs does not depend on 
		the training targets, resulting in almost identical curves for the different joints.}
	\label{fig:t_update Sarcos}
\end{figure*}

\begin{figure*}[p]
	\pgfplotsset{width=7.2cm, compat = 1.13, 
		height =74\textwidth /100, grid= major, 
		legend cell align = left, ticklabel style = {font=\scriptsize},
		every axis label/.append style={font=\small},
		legend style = {font=\scriptsize},title style={yshift=0pt, font = \small},
		every x tick scale label/.style={at={(xticklabel cs:1)},anchor=south west} }
	
	\center
	
	\begin{minipage}{0.48\textwidth}
		\def\file{plots/Real_Sarcos_long_plot_data.txt}
		\begin{tikzpicture}
		\begin{semilogyaxis}[grid=none,enlargelimits=false, axis on top,
		xlabel={$N$}, ylabel={$t(s)$}, 
		xmin=1, xmax = 44484,
		y label style={yshift=-0.2cm},
		title = {joint 1},
		scaled ticks=false, tick label style={/pgf/number format/fixed}
		]
		\addplot[thick] table[x = steps_IP, y = t_pred0_DLGP ]{\file};
		\addplot[thick, dashed] table[x = steps_IP, y = t_pred0_DLGP500 ]{\file};
		\addplot[red] table[x = steps_IP,y = t_pred0_GPexact ]{\file};
		\addplot[red, dashed] table[x = steps_LGP, y = t_pred0_LGP ]{\file};
		\addplot[blue] table[x = steps_IP, y = t_pred0_IP ]{\file};
		\addplot[blue, dashed] table[x = steps_Sparse,y = t_pred0_Sparse ]{\file};
		\addplot[green] table[x = steps_ISSGP,y = t_pred0_ISSGP ]{\file};
		\addplot[green, dashed] table[x = steps_lwpr,y = t_pred0_lwpr ]{\file};
		\end{semilogyaxis}
		\end{tikzpicture}
	\end{minipage}
	\begin{minipage}{0.48\textwidth}
		\def\file{plots/Real_Sarcos_long_plot_data.txt}
		\begin{tikzpicture}
		\begin{semilogyaxis}[grid=none,enlargelimits=false, axis on top,
		xlabel={$N$}, ylabel={$t(s)$}, 
		xmin=1, xmax = 44484,
		y label style={yshift=-0.2cm},
		title = {joint 2},
		scaled ticks=false, tick label style={/pgf/number format/fixed}
		]
		\addplot[thick] table[x = steps_IP, y = t_pred1_DLGP ]{\file};
		\addplot[thick, dashed] table[x = steps_IP, y = t_pred1_DLGP500 ]{\file};
		\addplot[red] table[x = steps_IP,y = t_pred1_GPexact ]{\file};
		\addplot[red, dashed] table[x = steps_LGP, y = t_pred1_LGP ]{\file};
		\addplot[blue] table[x = steps_IP, y = t_pred1_IP ]{\file};
		\addplot[blue, dashed] table[x = steps_Sparse,y = t_pred1_Sparse ]{\file};
		\addplot[green] table[x = steps_ISSGP,y = t_pred1_ISSGP ]{\file};
		\addplot[green, dashed] table[x = steps_lwpr,y = t_pred1_lwpr ]{\file};
		\end{semilogyaxis}
		\end{tikzpicture}
	\end{minipage}
	
	\begin{minipage}{0.48\textwidth}
		\def\file{plots/Real_Sarcos_long_plot_data.txt}
		\begin{tikzpicture}
		\begin{semilogyaxis}[grid=none,enlargelimits=false, axis on top,
		xlabel={$N$}, ylabel={$t(s)$}, 
		xmin=1, xmax = 44484, 
		y label style={yshift=-0.2cm},
		title = {joint 3},
		scaled ticks=false, tick label style={/pgf/number format/fixed}
		]
		\addplot[thick] table[x = steps_IP, y = t_pred2_DLGP ]{\file};
		\addplot[thick, dashed] table[x = steps_IP, y = t_pred2_DLGP500 ]{\file};
		\addplot[red] table[x = steps_IP,y = t_pred2_GPexact ]{\file};
		\addplot[red, dashed] table[x = steps_LGP, y = t_pred2_LGP ]{\file};
		\addplot[blue] table[x = steps_IP, y = t_pred2_IP ]{\file};
		\addplot[blue, dashed] table[x = steps_Sparse,y = t_pred2_Sparse ]{\file};
		\addplot[green] table[x = steps_ISSGP,y = t_pred2_ISSGP ]{\file};
		\addplot[green, dashed] table[x = steps_lwpr,y = t_pred2_lwpr ]{\file};
		\end{semilogyaxis}
		\end{tikzpicture}
	\end{minipage}
	\begin{minipage}{0.48\textwidth}
		\def\file{plots/Real_Sarcos_long_plot_data.txt}
		\begin{tikzpicture}
		\begin{semilogyaxis}[grid=none,enlargelimits=false, axis on top,
		xlabel={$N$}, ylabel={$t(s)$},
		xmin=1, xmax = 44484,
		y label style={yshift=-0.2cm},
		title = {joint 4},
		scaled ticks=false, tick label style={/pgf/number format/fixed}
		]
		\addplot[thick] table[x = steps_IP, y = t_pred3_DLGP ]{\file};
		\addplot[thick, dashed] table[x = steps_IP, y = t_pred3_DLGP500 ]{\file};
		\addplot[red] table[x = steps_IP,y = t_pred3_GPexact ]{\file};
		\addplot[red, dashed] table[x = steps_LGP, y = t_pred3_LGP ]{\file};
		\addplot[blue] table[x = steps_IP, y = t_pred3_IP ]{\file};
		\addplot[blue, dashed] table[x = steps_Sparse,y = t_pred3_Sparse ]{\file};
		\addplot[green] table[x = steps_ISSGP,y = t_pred3_ISSGP ]{\file};
		\addplot[green, dashed] table[x = steps_lwpr,y = t_pred3_lwpr ]{\file};
		\end{semilogyaxis}
		\end{tikzpicture}
	\end{minipage}
	
	\begin{minipage}{0.48\textwidth}
		\def\file{plots/Real_Sarcos_long_plot_data.txt}
		\begin{tikzpicture}
		\begin{semilogyaxis}[grid=none,enlargelimits=false, axis on top,
		xlabel={$N$}, ylabel={$t(s)$},
		xmin=1, xmax = 44484,
		y label style={yshift=-0.2cm},
		title = {joint 5},
		scaled ticks=false, tick label style={/pgf/number format/fixed}
		]
		\addplot[thick] table[x = steps_IP, y = t_pred4_DLGP ]{\file};
		\addplot[thick, dashed] table[x = steps_IP, y = t_pred4_DLGP500 ]{\file};
		\addplot[red] table[x = steps_IP,y = t_pred4_GPexact ]{\file};
		\addplot[red, dashed] table[x = steps_LGP, y = t_pred4_LGP ]{\file};
		\addplot[blue] table[x = steps_IP, y = t_pred4_IP ]{\file};
		\addplot[blue, dashed] table[x = steps_Sparse,y = t_pred4_Sparse ]{\file};
		\addplot[green] table[x = steps_ISSGP,y = t_pred4_ISSGP ]{\file};
		\addplot[green, dashed] table[x = steps_lwpr,y = t_pred4_lwpr ]{\file};
		\end{semilogyaxis}
		\end{tikzpicture}
	\end{minipage}
	\begin{minipage}{0.48\textwidth}
		\def\file{plots/Real_Sarcos_long_plot_data.txt}
		\begin{tikzpicture}
		\begin{semilogyaxis}[grid=none,enlargelimits=false, axis on top,
		xlabel={$N$}, ylabel={$t(s)$},
		xmin=1, xmax = 44484, 
		y label style={yshift=-0.2cm},
		title = {joint 6},
		scaled ticks=false, tick label style={/pgf/number format/fixed}
		]
		\addplot[thick] table[x = steps_IP, y =t_pred5_DLGP ]{\file};
		\addplot[thick, dashed] table[x = steps_IP, y = t_pred5_DLGP500 ]{\file};
		\addplot[red] table[x = steps_IP, y = t_pred5_GPexact ]{\file};
		\addplot[red, dashed] table[x = steps_LGP, y = t_pred5_LGP ]{\file};
		\addplot[blue] table[x = steps_IP, y = t_pred5_IP ]{\file};
		\addplot[blue, dashed] table[x = steps_Sparse,y = t_pred5_Sparse ]{\file};
		\addplot[green] table[x = steps_ISSGP,y = t_pred5_ISSGP ]{\file};
		\addplot[green, dashed] table[x = steps_lwpr,y = t_pred5_lwpr ]{\file};
		\end{semilogyaxis}
		\end{tikzpicture}
	\end{minipage}

	\begin{minipage}{0.48\textwidth}
		\def\file{plots/Real_Sarcos_long_plot_data.txt}
		\begin{tikzpicture}
		\begin{semilogyaxis}[grid=none,enlargelimits=false, axis on top,
		xlabel={$N$}, ylabel={$t(s)$}, 
		xmin=1, xmax = 44484,
		y label style={yshift=-0.2cm},
		legend style={at={(1.02,0.5)},anchor=west},
		title = {joint 7},
		scaled ticks=false, tick label style={/pgf/number format/fixed}	]
		\addplot[thick] table[x = steps_Sparse, y = t_pred6_DLGP  ]{\file};
		\addlegendentry{DLGP$_{100}$}
		\addplot[thick, dashed] table[x = steps_Sparse, y = t_pred6_DLGP500  ]{\file};
		\addlegendentry{DLGP$_{500}$}
		
		\addplot[red] table[x = steps_Sparse, y = t_pred6_LGP ]{\file};
		\addlegendentry{local GPs \cite{Nguyen-Tuong2009}}
		\addplot[red, dashed] table[x = steps_Sparse, y =t_pred6_IP  ]{\file};
		\addlegendentry{SONIG \cite{Bijl2017}}
		\addplot[blue] table[x = steps_Sparse, y = t_pred6_Sparse  ]{\file};
		\addlegendentry{DTC \cite{Schreiter2016}}
		\addplot[blue, dashed] table[x = steps_Sparse, y = t_pred6_ISSGP  ]{\file};
		\addlegendentry{I-SSGP \cite{Gijsberts2013}}
		\addplot[green] table[x = steps_lwpr,y = t_pred6_lwpr ]{\file};
		\addlegendentry{LWPR \cite{Vijayakumar2000}}
		\addplot[green, dashed] table[x = steps_Sparse, y = t_pred6_GPexact ]{\file};
		\addlegendentry{exact GP  \cite{Wang2019a}}
		\end{semilogyaxis}
		\end{tikzpicture}
	\end{minipage}
	
	\caption{Average time necessary to compute a prediction depending on the number of training samples $N$ for all joints 
		of the SARCOS data set: DLGPs are faster than most other methods and exhibit only a slow, logarithmic increase in 
		computation time.}
	\label{fig:t_pred Sarcos}
\end{figure*}
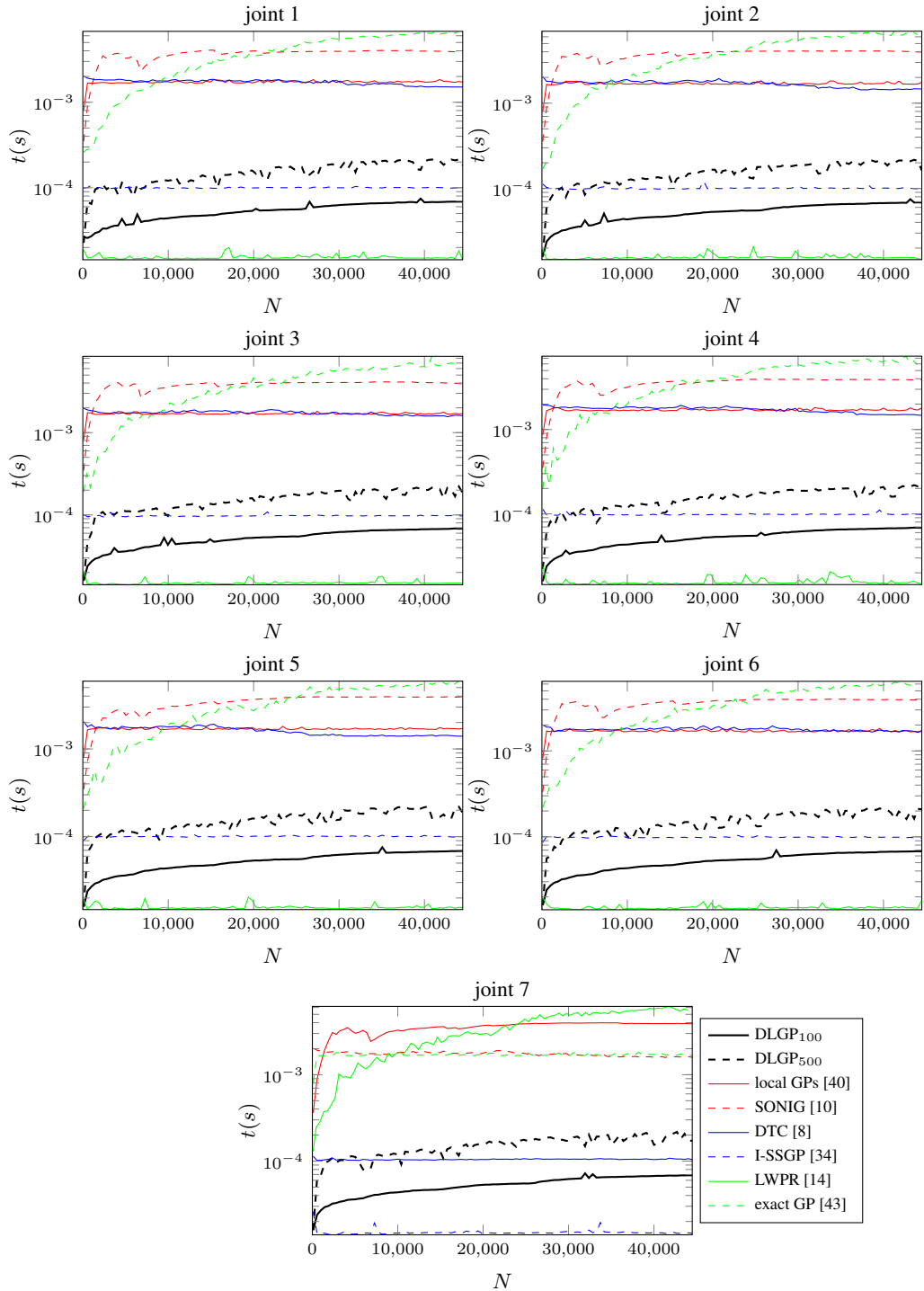

Figures \ref{fig:nMSE_Sarcos}-\ref{fig:t_pred Sarcos} depict the simulation results for the first scenario, which is 
evaluated on the SARCOS data set. The prediction error of the DLGPs, as illustrated in \cref{fig:nMSE_Sarcos} 
exhibits the same behavior for all joints: it is almost parallel to the prediction error curve of exact GPs. Furthermore, 
the DLGPs outperform the other real-time learning approaches on all joints. While some other methods also 
show good regression performance, their variance predictions can be highly unreliable, as shown in 
\cref{fig:NLL_Sarcos}. For some methods, the NLL is temporarily increasing despite of growing numbers of training samples, while
the NLL is continuously decreasing for DLGPs. In fact, the NLL of 
the DLGPs is barely higher than for exact GPs, which underlines the high quality of the predictive distributions of DLGPs. 
In addition to the advantages regarding the predictive distributions, 
DLGPs also strongly benefit from low average update times, which allow 
real-time learning, as shown in \cref{fig:t_update Sarcos}. For all joints, the update time of DLGPs with $\bar{N}=100$ 
is typically more than $10$ times faster than for state-of-the-art methods. Moreover, the prediction time of the DLGPs, 
as illustrated in \cref{fig:t_pred Sarcos}, exhibits a similar magnitude as the prediction time. Due to the slow growth of the 
prediction time, DLGPs allow similar prediction rates as many existing methods, while additionally updating the model with 
the same frequency.

\subsection{KUKA Flask Pushing Data Set}

Due to the different scenario, in which the KUKA flask pushing data set is learned, the performance measures are slightly
adjusted. The prediction error is investigated using the online nMSE as defined in \cite{Gijsberts2013}, which corresponds 
to the cumulative average normalized mean squared error. Analogously, we define the average online negative log-likelihood
as the cumulative average NLL. In order to avoid overly noisy computation time estimates arising from single predictions or 
updates, we apply a moving average filter with filter width $1,000$ to the update and prediction times.

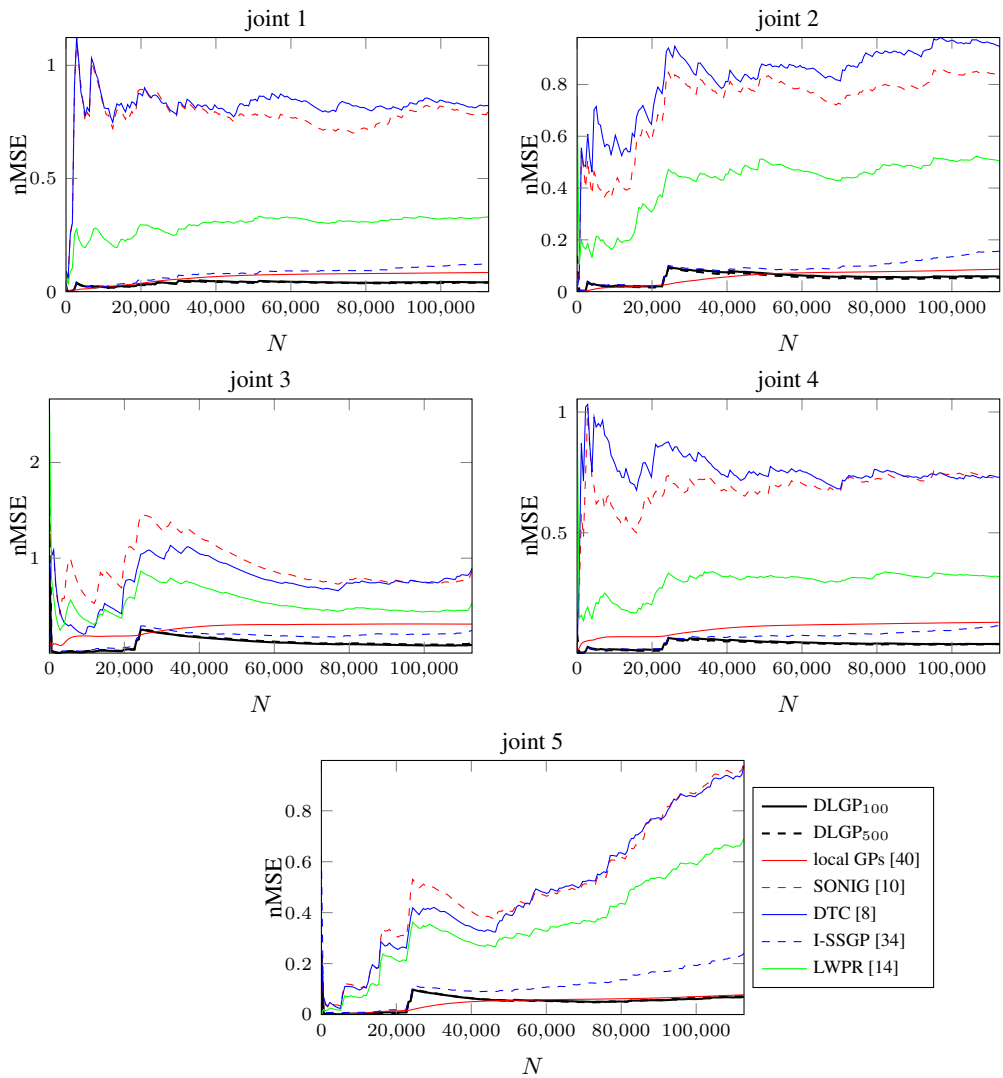
\begin{figure*}[p]
	\pgfplotsset{width=7.2cm, compat = 1.13, 
		height =74\textwidth /100, grid= major, 
		legend cell align = left, ticklabel style = {font=\scriptsize},
		every axis label/.append style={font=\small},
		legend style = {font=\scriptsize},title style={yshift=0pt, font = \small},
		every x tick scale label/.style={at={(xticklabel cs:1)},anchor=south west} }
	
	\center
	
	\begin{minipage}{0.48\textwidth}
		\def\file{plots/KUKA_flaskNLL_plot_data.txt}
		\begin{tikzpicture}
		\begin{axis}[grid=none,enlargelimits=false, axis on top,
		xlabel={$N$}, ylabel={nMSE},
		xmin=1, xmax = 112761,
		y label style={yshift=-0.2cm},
		title = {joint 1},
		scaled ticks=false, tick label style={/pgf/number format/fixed}
		]
		\addplot[thick] table[x = steps_IP, y = e0_DLGP ]{\file};
		\addplot[thick, dashed] table[x = steps_IP, y = e0_DLGP500 ]{\file};
		\addplot[red] table[x = steps_LGP, y = e0_LGP ]{\file};
		\addplot[red,dashed] table[x = steps_IP, y = e0_IP ]{\file};
		\addplot[blue] table[x = steps_Sparse,y = e0_Sparse ]{\file};
		\addplot[blue,dashed] table[x = steps_ISSGP,y = e0_ISSGP ]{\file};
		\addplot[green] table[x = steps_lwpr,y = e0_lwpr ]{\file};
		\end{axis}
		\end{tikzpicture}
	\end{minipage}
	\begin{minipage}{0.48\textwidth}
		\def\file{plots/KUKA_flaskNLL_plot_data.txt}
		\begin{tikzpicture}
		\begin{axis}[grid=none,enlargelimits=false, axis on top,
		xlabel={$N$}, ylabel={nMSE}, 
		xmin=1, xmax = 112761, 
		y label style={yshift=-0.2cm},
		title = {joint 2},
		scaled ticks=false, tick label style={/pgf/number format/fixed}
		]
		\addplot[thick] table[x = steps_IP, y = e1_DLGP ]{\file};
		\addplot[thick, dashed] table[x = steps_IP, y = e1_DLGP500 ]{\file};
		\addplot[red] table[x = steps_LGP, y = e1_LGP ]{\file};
		\addplot[red,dashed] table[x = steps_IP, y = e1_IP ]{\file};
		\addplot[blue] table[x = steps_Sparse,y = e1_Sparse ]{\file};
		\addplot[blue,dashed] table[x = steps_ISSGP,y = e1_ISSGP ]{\file};
		\addplot[green] table[x = steps_lwpr,y = e1_lwpr ]{\file};
		\end{axis}
		\end{tikzpicture}
	\end{minipage}
	
	\begin{minipage}{0.48\textwidth}
		\def\file{plots/KUKA_flaskNLL_plot_data.txt}
		\begin{tikzpicture}
		\begin{axis}[grid=none,enlargelimits=false, axis on top,
		xlabel={$N$}, ylabel={nMSE}, 
		xmin=1, xmax = 112761, 
		y label style={yshift=-0.2cm},
		title = {joint 3},
		scaled ticks=false, tick label style={/pgf/number format/fixed}
		]
		\addplot[thick] table[x = steps_IP, y = e2_DLGP ]{\file};
		\addplot[thick, dashed] table[x = steps_IP, y = e2_DLGP500 ]{\file};
		\addplot[red] table[x = steps_LGP, y = e2_LGP ]{\file};
		\addplot[red, dashed] table[x = steps_IP, y = e2_IP ]{\file};
		\addplot[blue] table[x = steps_Sparse,y = e2_Sparse ]{\file};
		\addplot[blue,dashed] table[x = steps_ISSGP,y = e2_ISSGP ]{\file};
		\addplot[green] table[x = steps_lwpr,y = e2_lwpr ]{\file};
		\end{axis}
		\end{tikzpicture}
	\end{minipage}
	\begin{minipage}{0.48\textwidth}
		\def\file{plots/KUKA_flaskNLL_plot_data.txt}
		\begin{tikzpicture}
		\begin{axis}[grid=none,enlargelimits=false, axis on top,
		xlabel={$N$}, ylabel={nMSE}, 
		xmin=1, xmax = 112761, 
		y label style={yshift=-0.2cm},
		title = {joint 4},
		scaled ticks=false, tick label style={/pgf/number format/fixed}
		]
		\addplot[thick] table[x = steps_IP, y = e3_DLGP ]{\file};
		\addplot[thick, dashed] table[x = steps_IP, y = e3_DLGP500 ]{\file};
		\addplot[red] table[x = steps_LGP, y = e3_LGP ]{\file};
		\addplot[red,dashed] table[x = steps_IP, y = e3_IP ]{\file};
		\addplot[blue] table[x = steps_Sparse,y = e3_Sparse ]{\file};
		\addplot[blue,dashed] table[x = steps_ISSGP,y = e3_ISSGP ]{\file};
		\addplot[green] table[x = steps_lwpr,y = e3_lwpr ]{\file};
		\end{axis}
		\end{tikzpicture}
	\end{minipage}
	
	\begin{minipage}{0.48\textwidth}
		\def\file{plots/KUKA_flaskNLL_plot_data.txt}
		\begin{tikzpicture}
		\begin{axis}[grid=none,enlargelimits=false, axis on top,
		xlabel={$N$}, ylabel={nMSE}, 
		xmin=1, xmax = 112761, 
		y label style={yshift=-0.2cm},
		legend style={at={(1.02,0.5)},anchor=west},
		title = {joint 5},
		scaled ticks=false, tick label style={/pgf/number format/fixed}
		]
		\addplot[thick] table[x = steps_IP, y = e4_DLGP ]{\file};
		\addlegendentry{DLGP$_{100}$}
		\addplot[thick, dashed] table[x = steps_IP, y = e4_DLGP500 ]{\file};
		\addlegendentry{DLGP$_{500}$}
		\addplot[red] table[x = steps_LGP, y = e4_LGP ]{\file};
		\addlegendentry{local GPs \cite{Nguyen-Tuong2009}}
		\addplot[red, dashed] table[x = steps_IP, y = e4_IP ]{\file};
		\addlegendentry{SONIG \cite{Bijl2017}}
		\addplot[blue] table[x = steps_Sparse,y = e4_Sparse ]{\file};
		\addlegendentry{DTC \cite{Schreiter2016}}
		\addplot[blue, dashed] table[x = steps_ISSGP,y = e4_ISSGP ]{\file};
		\addlegendentry{I-SSGP \cite{Gijsberts2013}}
		\addplot[green] table[x = steps_lwpr,y = e4_lwpr ]{\file};
		\addlegendentry{LWPR \cite{Vijayakumar2000}}
		\end{axis}
		\end{tikzpicture}
	\end{minipage}
	
	\caption{Online normalized mean squared errors for 
		all joints of the KUKA flask pushing data set: DLGPs exhibit low regression errors and are capable 
		of recovering from sudden environmental changes causing brief steps in the error.}
	\label{fig:nMSE_KUKA}
\end{figure*}

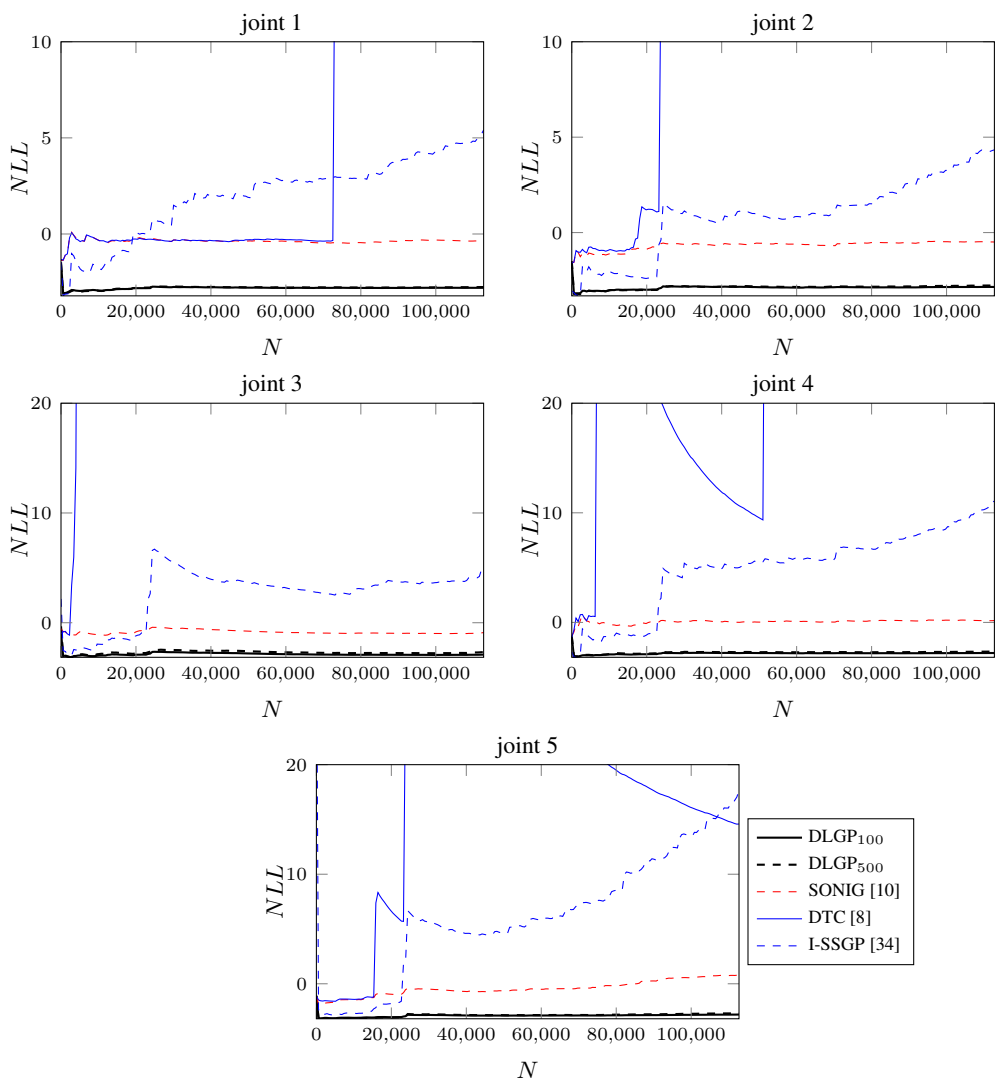
\begin{figure*}[p]
	\pgfplotsset{width=7.2cm, compat = 1.13, 
		height =74\textwidth /100, grid= major, 
		legend cell align = left, ticklabel style = {font=\scriptsize},
		every axis label/.append style={font=\small},
		legend style = {font=\scriptsize},title style={yshift=0pt, font = \small},
		every x tick scale label/.style={at={(xticklabel cs:1)},anchor=south west} }
	
	\center
	
	\begin{minipage}{0.48\textwidth}
		\def\file{plots/KUKA_flaskNLL_plot_data.txt}
		\begin{tikzpicture}
		\begin{axis}[grid=none,enlargelimits=false, axis on top,
		xlabel={$N$}, ylabel={$NLL$}, 
		xmin=1, xmax = 112761, ymax =10,
		y label style={yshift=-0.2cm},
		title = {joint 1}, 
		scaled ticks=false, tick label style={/pgf/number format/fixed}
		]
		\addplot[thick] table[x = steps_IP, y = aNLL0_DLGP ]{\file};
		\addplot[thick, dashed] table[x = steps_IP, y = aNLL0_DLGP500 ]{\file};
		\addplot[red,dashed] table[x = steps_IP, y = aNLL0_IP ]{\file};
		\addplot[blue] table[x = steps_Sparse,y = aNLL0_Sparse ]{\file};
		\addplot[blue,dashed] table[x = steps_ISSGP,y = aNLL0_ISSGP ]{\file};
		\end{axis}
		\end{tikzpicture}
	\end{minipage}
	\begin{minipage}{0.48\textwidth}
		\def\file{plots/KUKA_flaskNLL_plot_data.txt}
		\begin{tikzpicture}
		\begin{axis}[grid=none,enlargelimits=false, axis on top,
		xlabel={$N$}, ylabel={$NLL$}, 
		xmin=1, xmax = 112761, ymax =10, ,
		y label style={yshift=-0.2cm},
		title = {joint 2}, 
		scaled ticks=false, tick label style={/pgf/number format/fixed}
		]
		\addplot[thick] table[x = steps_IP, y = aNLL1_DLGP ]{\file};
		\addplot[thick, dashed] table[x = steps_IP, y = aNLL1_DLGP500 ]{\file};
		\addplot[red,dashed] table[x = steps_IP, y = aNLL1_IP ]{\file};
		\addplot[blue] table[x = steps_Sparse,y = aNLL1_Sparse ]{\file};
		\addplot[blue,dashed] table[x = steps_ISSGP,y = aNLL1_ISSGP ]{\file};
		\end{axis}
		\end{tikzpicture}
	\end{minipage}
	
	\begin{minipage}{0.48\textwidth}
		\def\file{plots/KUKA_flaskNLL_plot_data.txt}
		\begin{tikzpicture}
		\begin{axis}[grid=none,enlargelimits=false, axis on top,
		xlabel={$N$}, ylabel={$NLL$},
		xmin=1, xmax = 112761, ymax =20, 
		y label style={yshift=-0.2cm},
		title = {joint 3}, 
		scaled ticks=false, tick label style={/pgf/number format/fixed}
		]
		\addplot[thick] table[x = steps_IP, y = aNLL2_DLGP ]{\file};
		\addplot[thick, dashed] table[x = steps_IP, y = aNLL2_DLGP500 ]{\file};
		\addplot[red, dashed] table[x = steps_IP, y = aNLL2_IP ]{\file};
		\addplot[blue] table[x = steps_Sparse,y = aNLL2_Sparse ]{\file};
		\addplot[blue,dashed] table[x = steps_ISSGP,y = aNLL2_ISSGP ]{\file};
		\end{axis}
		\end{tikzpicture}
	\end{minipage}
	\begin{minipage}{0.48\textwidth}
		\def\file{plots/KUKA_flaskNLL_plot_data.txt}
		\begin{tikzpicture}
		\begin{axis}[grid=none,enlargelimits=false, axis on top,
		xlabel={$N$}, ylabel={$NLL$},
		xmin=1, xmax = 112761, ymax =20, 
		y label style={yshift=-0.2cm},
		title = {joint 4}, 
		scaled ticks=false, tick label style={/pgf/number format/fixed}
		]
		\addplot[thick] table[x = steps_IP, y = aNLL3_DLGP ]{\file};
		\addplot[thick, dashed] table[x = steps_IP, y = aNLL3_DLGP500 ]{\file};
		\addplot[red,dashed] table[x = steps_IP, y = aNLL3_IP ]{\file};
		\addplot[blue] table[x = steps_Sparse,y = aNLL3_Sparse ]{\file};
		\addplot[blue,dashed] table[x = steps_ISSGP,y = aNLL3_ISSGP ]{\file};
		\end{axis}
		\end{tikzpicture}
	\end{minipage}
	
	\begin{minipage}{0.48\textwidth}
		\def\file{plots/KUKA_flaskNLL_plot_data.txt}
		\begin{tikzpicture}
		\begin{axis}[grid=none,enlargelimits=false, axis on top,
		xlabel={$N$}, ylabel={$NLL$},
		xmin=1, xmax = 112761, ymax =20, 
		y label style={yshift=-0.2cm},
		legend style={at={(1.02,0.5)},anchor=west},
		title = {joint 5}, 
		scaled ticks=false, tick label style={/pgf/number format/fixed}
		]
		\addplot[thick] table[x = steps_IP, y = aNLL4_DLGP ]{\file};
		\addlegendentry{DLGP$_{100}$}
		\addplot[thick, dashed] table[x = steps_IP, y = aNLL4_DLGP500 ]{\file};
		\addlegendentry{DLGP$_{500}$}
		\addplot[red, dashed] table[x = steps_IP, y = aNLL4_IP ]{\file};
		\addlegendentry{SONIG \cite{Bijl2017}}
		\addplot[blue] table[x = steps_Sparse,y = aNLL4_Sparse ]{\file};
		\addlegendentry{DTC \cite{Schreiter2016}}
		\addplot[blue, dashed] table[x = steps_ISSGP,y = aNLL4_ISSGP ]{\file};
		\addlegendentry{I-SSGP \cite{Gijsberts2013}}
		\end{axis}
		\end{tikzpicture}
	\end{minipage}
	
	\caption{Average online negative log-likelihood for all joints of the KUKA flask 
		pushing data set: environmental changes causing steps in the regression error lead to a significant deterioration of the reliability of the 
		predictive distribution in state-of-the-art methods, while DLGPs are barely influenced.}
	\label{fig:NLL KUKA}
\end{figure*}

\begin{figure*}[p]
	\pgfplotsset{width=7.2cm, compat = 1.13, 
		height =74\textwidth /100, grid= major, 
		legend cell align = left, ticklabel style = {font=\scriptsize},
		every axis label/.append style={font=\small},
		legend style = {font=\scriptsize},title style={yshift=0pt, font = \small},
		every x tick scale label/.style={at={(xticklabel cs:1)},anchor=south west} }
	
	\center
	
	\begin{minipage}{0.48\textwidth}
		\def\file{plots/KUKA_flaskNLL_plot_data.txt}
		\begin{tikzpicture}
		\begin{semilogyaxis}[grid=none,enlargelimits=false, axis on top,
		xlabel={$N$}, ylabel={$t(s)$}, 
		xmin=1, xmax = 112761,
		y label style={yshift=-0.2cm},
		title = {joint 1},
		scaled ticks=false, tick label style={/pgf/number format/fixed}
		]
		\addplot[thick] table[x = steps_IP, y = t_pred0_DLGP ]{\file};
		\addplot[thick, dashed] table[x = steps_IP, y = t_pred0_DLGP500 ]{\file};
		\addplot[red] table[x = steps_LGP, y = t_pred0_LGP ]{\file};
		\addplot[red,dashed] table[x = steps_IP, y = t_pred0_IP ]{\file};
		\addplot[blue] table[x = steps_Sparse,y = t_pred0_Sparse ]{\file};
		\addplot[blue,dashed] table[x = steps_ISSGP,y = t_pred0_ISSGP ]{\file};
		\addplot[green] table[x = steps_lwpr,y = t_pred0_lwpr ]{\file};
		\end{semilogyaxis}
		\end{tikzpicture}
	\end{minipage}
	\begin{minipage}{0.48\textwidth}
		\def\file{plots/KUKA_flaskNLL_plot_data.txt}
		\begin{tikzpicture}
		\begin{semilogyaxis}[grid=none,enlargelimits=false, axis on top,
		xlabel={$N$}, ylabel={$t(s)$}, 
		xmin=1, xmax = 112761, 
		y label style={yshift=-0.2cm},
		title = {joint 2},
		scaled ticks=false, tick label style={/pgf/number format/fixed}
		]
		\addplot[thick] table[x = steps_IP, y = t_pred1_DLGP ]{\file};
		\addplot[thick, dashed] table[x = steps_IP, y = t_pred1_DLGP500 ]{\file};
		\addplot[red] table[x = steps_LGP, y = t_pred1_LGP ]{\file};
		\addplot[red,dashed] table[x = steps_IP, y = t_pred1_IP ]{\file};
		\addplot[blue] table[x = steps_Sparse,y = t_pred1_Sparse ]{\file};
		\addplot[blue,dashed] table[x = steps_ISSGP,y = t_pred1_ISSGP ]{\file};
		\addplot[green] table[x = steps_lwpr,y = t_pred1_lwpr ]{\file};
		\end{semilogyaxis}
		\end{tikzpicture}
	\end{minipage}
	
	\begin{minipage}{0.48\textwidth}
		\def\file{plots/KUKA_flaskNLL_plot_data.txt}
		\begin{tikzpicture}
		\begin{semilogyaxis}[grid=none,enlargelimits=false, axis on top,
		xlabel={$N$}, ylabel={$t(s)$}, 
		xmin=1, xmax = 112761,
		y label style={yshift=-0.2cm},
		title = {joint 3},
		scaled ticks=false, tick label style={/pgf/number format/fixed}
		]
		\addplot[thick] table[x = steps_IP, y = t_pred2_DLGP ]{\file};
		\addplot[thick, dashed] table[x = steps_IP, y = t_pred2_DLGP500 ]{\file};
		\addplot[red] table[x = steps_LGP, y = t_pred2_LGP ]{\file};
		\addplot[red, dashed] table[x = steps_IP, y = t_pred2_IP ]{\file};
		\addplot[blue] table[x = steps_Sparse,y = t_pred2_Sparse ]{\file};
		\addplot[blue,dashed] table[x = steps_ISSGP,y = t_pred2_ISSGP ]{\file};
		\addplot[green] table[x = steps_lwpr,y = t_pred2_lwpr ]{\file};
		\end{semilogyaxis}
		\end{tikzpicture}
	\end{minipage}
	\begin{minipage}{0.48\textwidth}
		\def\file{plots/KUKA_flaskNLL_plot_data.txt}
		\begin{tikzpicture}
		\begin{semilogyaxis}[grid=none,enlargelimits=false, axis on top,
		xlabel={$N$}, ylabel={$t(s)$}, 
		xmin=1, xmax = 112761, 
		y label style={yshift=-0.2cm},
		title = {joint 4},
		scaled ticks=false, tick label style={/pgf/number format/fixed}
		]
		\addplot[thick] table[x = steps_IP, y = t_pred3_DLGP ]{\file};
		\addplot[thick, dashed] table[x = steps_IP, y = t_pred3_DLGP500 ]{\file};
		\addplot[red] table[x = steps_LGP, y = t_pred3_LGP ]{\file};
		\addplot[red,dashed] table[x = steps_IP, y = t_pred3_IP ]{\file};
		\addplot[blue] table[x = steps_Sparse,y = t_pred3_Sparse ]{\file};
		\addplot[blue,dashed] table[x = steps_ISSGP,y = t_pred3_ISSGP ]{\file};
		\addplot[green] table[x = steps_lwpr,y = t_pred3_lwpr ]{\file};
		\end{semilogyaxis}
		\end{tikzpicture}
	\end{minipage}
	
	\begin{minipage}{0.48\textwidth}
		\def\file{plots/KUKA_flaskNLL_plot_data.txt}
		\begin{tikzpicture}
		\begin{semilogyaxis}[grid=none,enlargelimits=false, axis on top,
		xlabel={$N$}, ylabel={$t(s)$}, 
		xmin=1, xmax = 112761, 
		y label style={yshift=-0.2cm},
		legend style={at={(1.02,0.5)},anchor=west},
		title = {joint 5},
		scaled ticks=false, tick label style={/pgf/number format/fixed}
		]
		\addplot[thick] table[x = steps_IP, y = t_pred4_DLGP ]{\file};
		\addlegendentry{DLGP$_{100}$}
		\addplot[thick, dashed] table[x = steps_IP, y = t_pred4_DLGP500 ]{\file};
		\addlegendentry{DLGP$_{500}$}
		\addplot[red] table[x = steps_LGP, y = t_pred4_LGP ]{\file};
		\addlegendentry{local GPs \cite{Nguyen-Tuong2009}}
		\addplot[red, dashed] table[x = steps_IP, y = t_pred4_IP ]{\file};
		\addlegendentry{SONIG \cite{Bijl2017}}
		\addplot[blue] table[x = steps_Sparse,y = t_pred4_Sparse ]{\file};
		\addlegendentry{DTC \cite{Schreiter2016}}
		\addplot[blue, dashed] table[x = steps_ISSGP,y = t_pred4_ISSGP ]{\file};
		\addlegendentry{I-SSGP \cite{Gijsberts2013}}
		\addplot[green] table[x = steps_lwpr,y = t_pred4_lwpr ]{\file};
		\addlegendentry{LWPR \cite{Vijayakumar2000}}
		
		\end{semilogyaxis}
		\end{tikzpicture}
	\end{minipage}
	
	\caption{Average prediction time for all joints 
		of the KUKA flask pushing data set: Even with variance prediction, DLGPs are only 
		slightly slower than the fastest state-of-the-art method, and exhibit a prediction time to 
		the commonly used LWPR.}
	\label{fig:t_pred KUKA}
\end{figure*}
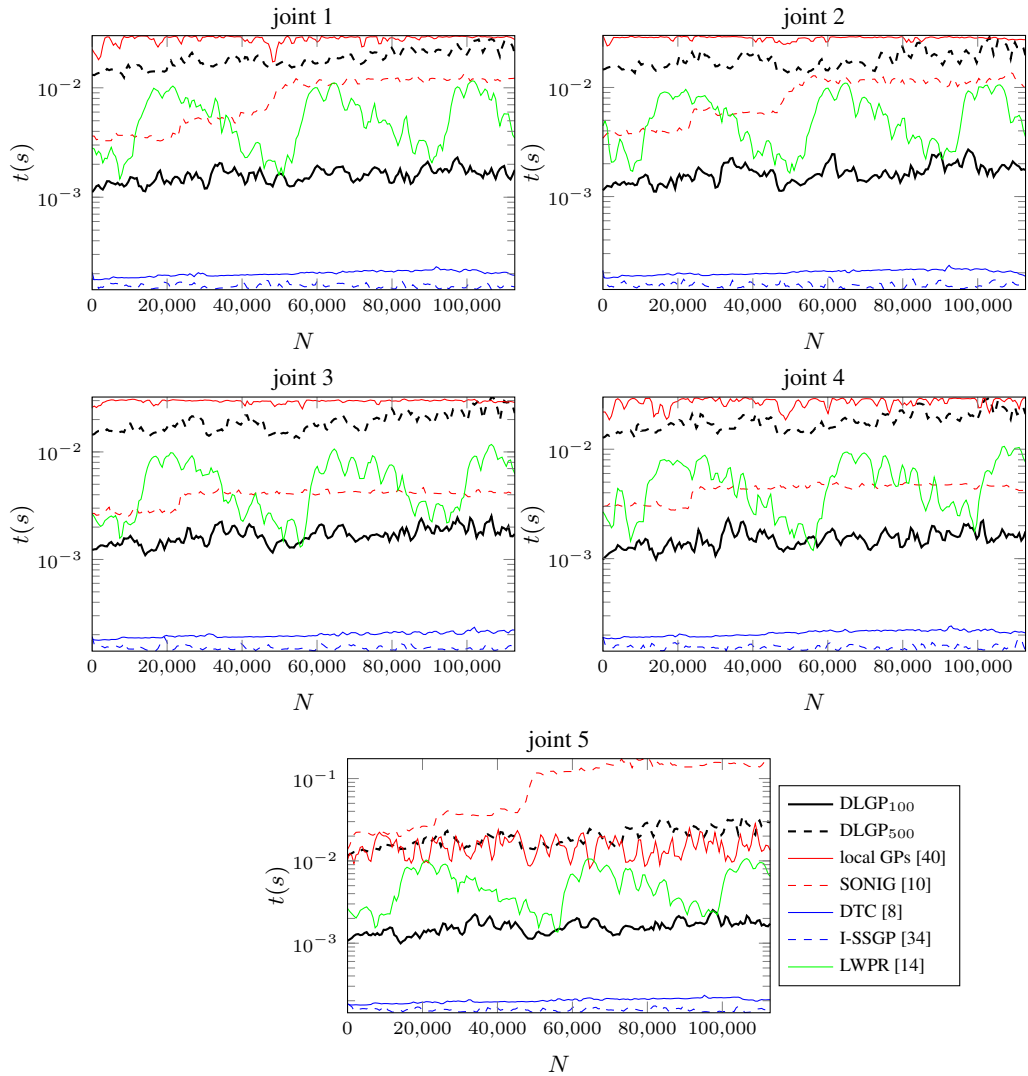

The simulation results of the second scenario on the KUKA flask pushing data set are depicted in 
Figs.~\ref{fig:nMSE_KUKA}-\ref{fig:t_pred KUKA}. As clearly shown in \cref{fig:nMSE_KUKA}, 
the prediction error of almost all methods suffers from a sudden increase after approximately $23,000$ 
data points of the online data for almost all joints, which is due to an outlier both in the targets and 
inputs. Local GPs do not exhibit a step in the error curves, but the error starts to increase after the 
outlier. While DLGPs suffer from a step similar to I-SSGPs, they convince through a continuously 
improving error afterwards, while the prediction error of other methods remains constant or even 
increases for some joints. The reason for this behavior is the division of local models, which 
ensures that an outlier can have an effect only in its neighborhood, which becomes smaller when more
data becomes available. As this effect is stronger, when each local model contains fewer samples, 
it is clear that the DLGP with $\bar{N}=100$ outperforms the DLGP with $\bar{N}=500$. The outlier has
an even stronger impact on the predictive distributions of most models, as indicated in 
\cref{fig:NLL KUKA}. While DLGPs exhibit a continuously low average NLL with merely 
a small increase after $23,000$ samples, methods such as I-SSGP or sparse online
GPs with DTC approximation exhibit rapid increases in the average NLL. Therefore, 
one of the main strengths of DLGPs is their capability of providing reliable predictive
distributions, while other methods are often over-confident. In addition, the slow 
logarithmic growth of the update complexity of DLGPs leads to an almost constant average 
update time for large numbers of training samples, which is lower than for most
other methods. This is depicted in \cref{fig:t_update KUKA}, where it should be noted
that all figures in the second scenario only display the results on the online data. Hence, $N=0$ 
in the figures corresponds to models, which have already been trained using $16,940$ offline samples. 
Although SONIG is faster than the DLGPs, this is caused by a very small active set, which in turn results 
in poor learning performance. Therefore, the DLGP with $\bar{N}=100$ samples is the only method 
providing fast update rates and good learning performance. Finally, the prediction times of DLGPs are 
comparable to many state-of-the-art approaches, as shown in \cref{fig:t_pred KUKA}. The logarithmic
growth in complexity merely causes a prediction time increase by a factor of approximately $2$, 
when adding more than $10^5$ training samples. Although other methods are faster in prediction, 
the prediction times of DLGPs could be easily reduced through a decrease of $\theta$ and $\bar{N}$
as discussed in \cref{sec:further Info}. Thereby, it is straightforward to tune both parameters 
under consideration of constraints on the prediction and update times.

\begin{figure*}[p]
	\pgfplotsset{width=7.2cm, compat = 1.13, 
		height =74\textwidth /100, grid= major, 
		legend cell align = left, ticklabel style = {font=\scriptsize},
		every axis label/.append style={font=\small},
		legend style = {font=\scriptsize},title style={yshift=0pt, font = \small},
		every x tick scale label/.style={at={(xticklabel cs:1)},anchor=south west} }
	
	\center
	
	\begin{minipage}{0.48\textwidth}
		\def\file{plots/KUKA_flaskNLL_plot_data.txt}
		\begin{tikzpicture}
		\begin{semilogyaxis}[grid=none,enlargelimits=false, axis on top,
		xlabel={$N$}, ylabel={$t(s)$}, 
		xmin=1, xmax = 112761,
		y label style={yshift=-0.2cm},ytick={0.001,1,1000},
		title = {joint 1},
		scaled ticks=false, tick label style={/pgf/number format/fixed}
		]
		\addplot[thick] table[x = steps_IP, y = t_update0_DLGP ]{\file};
		\addplot[thick, dashed] table[x = steps_IP, y = t_update0_DLGP500 ]{\file};
		\addplot[red] table[x = steps_LGP, y = t_update0_LGP ]{\file};
		\addplot[red,dashed] table[x = steps_IP, y = t_update0_IP ]{\file};
		\addplot[blue] table[x = steps_Sparse,y = t_update0_Sparse ]{\file};
		\addplot[blue,dashed] table[x = steps_ISSGP,y = t_update0_ISSGP ]{\file};
		\addplot[green] table[x = steps_lwpr,y = t_update0_lwpr ]{\file};
		\end{semilogyaxis}
		\end{tikzpicture}
	\end{minipage}
	\begin{minipage}{0.48\textwidth}
		\def\file{plots/KUKA_flaskNLL_plot_data.txt}
		\begin{tikzpicture}
		\begin{semilogyaxis}[grid=none,enlargelimits=false, axis on top,
		xlabel={$N$}, ylabel={$t(s)$},
		xmin=1, xmax = 112761, 
		y label style={yshift=-0.2cm},ytick={0.001,1,1000},
		title = {joint 2},
		scaled ticks=false, tick label style={/pgf/number format/fixed}
		]
		\addplot[thick] table[x = steps_IP, y = t_update1_DLGP ]{\file};
		\addplot[thick, dashed] table[x = steps_IP, y = t_update1_DLGP500 ]{\file};
		\addplot[red] table[x = steps_LGP, y = t_update1_LGP ]{\file};
		\addplot[red,dashed] table[x = steps_IP, y = t_update1_IP ]{\file};
		\addplot[blue] table[x = steps_Sparse,y = t_update1_Sparse ]{\file};
		\addplot[blue,dashed] table[x = steps_ISSGP,y = t_update1_ISSGP ]{\file};
		\addplot[green] table[x = steps_lwpr,y = t_update1_lwpr ]{\file};
		\end{semilogyaxis}
		\end{tikzpicture}
	\end{minipage}
	
	\begin{minipage}{0.48\textwidth}
		\def\file{plots/KUKA_flaskNLL_plot_data.txt}
		\begin{tikzpicture}
		\begin{semilogyaxis}[grid=none,enlargelimits=false, axis on top,
		xlabel={$N$}, ylabel={$t(s)$}, 
		xmin=1, xmax = 112761, 
		y label style={yshift=-0.2cm},ytick={0.001,1,1000},
		title = {joint 3},
		scaled ticks=false, tick label style={/pgf/number format/fixed}
		]
		\addplot[thick] table[x = steps_IP, y = t_update2_DLGP ]{\file};
		\addplot[thick, dashed] table[x = steps_IP, y = t_update2_DLGP500 ]{\file};
		\addplot[red] table[x = steps_LGP, y = t_update2_LGP ]{\file};
		\addplot[red, dashed] table[x = steps_IP, y = t_update2_IP ]{\file};
		\addplot[blue] table[x = steps_Sparse,y = t_update2_Sparse ]{\file};
		\addplot[blue,dashed] table[x = steps_ISSGP,y = t_update2_ISSGP ]{\file};
		\addplot[green] table[x = steps_lwpr,y = t_update2_lwpr ]{\file};
		\end{semilogyaxis}
		\end{tikzpicture}
	\end{minipage}
	\begin{minipage}{0.48\textwidth}
		\def\file{plots/KUKA_flaskNLL_plot_data.txt}
		\begin{tikzpicture}
		\begin{semilogyaxis}[grid=none,enlargelimits=false, axis on top,
		xlabel={$N$}, ylabel={$t(s)$}, 
		xmin=1, xmax = 112761, 
		y label style={yshift=-0.2cm},ytick={0.001,1,1000},
		title = {joint 4},
		scaled ticks=false, tick label style={/pgf/number format/fixed}
		]
		\addplot[thick] table[x = steps_IP, y = t_update3_DLGP ]{\file};
		\addplot[thick, dashed] table[x = steps_IP, y = t_update3_DLGP500 ]{\file};
		\addplot[red] table[x = steps_LGP, y = t_update3_LGP ]{\file};
		\addplot[red,dashed] table[x = steps_IP, y = t_update3_IP ]{\file};
		\addplot[blue] table[x = steps_Sparse,y = t_update3_Sparse ]{\file};
		\addplot[blue,dashed] table[x = steps_ISSGP,y = t_update3_ISSGP ]{\file};
		\addplot[green] table[x = steps_lwpr,y = t_update3_lwpr ]{\file};
		\end{semilogyaxis}
		\end{tikzpicture}
	\end{minipage}
	
	\begin{minipage}{0.48\textwidth}
		\def\file{plots/KUKA_flaskNLL_plot_data.txt}
		\begin{tikzpicture}
		\begin{semilogyaxis}[grid=none,enlargelimits=false, axis on top,
		xlabel={$N$}, ylabel={$t(s)$}, 
		xmin=1, xmax = 112761, 
		y label style={yshift=-0.2cm},ytick={0.001,1,1000},
		legend style={at={(1.02,0.5)},anchor=west},
		title = {joint 5},
		scaled ticks=false, tick label style={/pgf/number format/fixed}
		]
		\addplot[thick] table[x = steps_IP, y = t_update4_DLGP ]{\file};
		\addlegendentry{DLGP$_{100}$}
		\addplot[thick, dashed] table[x = steps_IP, y = t_update4_DLGP500 ]{\file};
		\addlegendentry{DLGP$_{500}$}
		\addplot[red] table[x = steps_LGP, y = t_update4_LGP ]{\file};
		\addlegendentry{local GPs \cite{Nguyen-Tuong2009}}
		\addplot[red, dashed] table[x = steps_IP, y = t_update4_IP ]{\file};
		\addlegendentry{SONIG \cite{Bijl2017}}
		\addplot[blue] table[x = steps_Sparse,y = t_update4_Sparse ]{\file};
		\addlegendentry{DTC \cite{Schreiter2016}}
		\addplot[blue, dashed] table[x = steps_ISSGP,y = t_update4_ISSGP ]{\file};
		\addlegendentry{I-SSGP \cite{Gijsberts2013}}
		\addplot[green] table[x = steps_lwpr,y = t_update4_lwpr ]{\file};
		\addlegendentry{LWPR \cite{Vijayakumar2000}}
		\end{semilogyaxis}
		\end{tikzpicture}
	\end{minipage}
	
	\caption{Average update time for all joints 
		of the KUKA flask pushing data set: despite of large training set sizes, training of the DLGP$_{100}$ model is faster than methods
		with a comparable regression performance.}
	\label{fig:t_update KUKA}
\end{figure*}
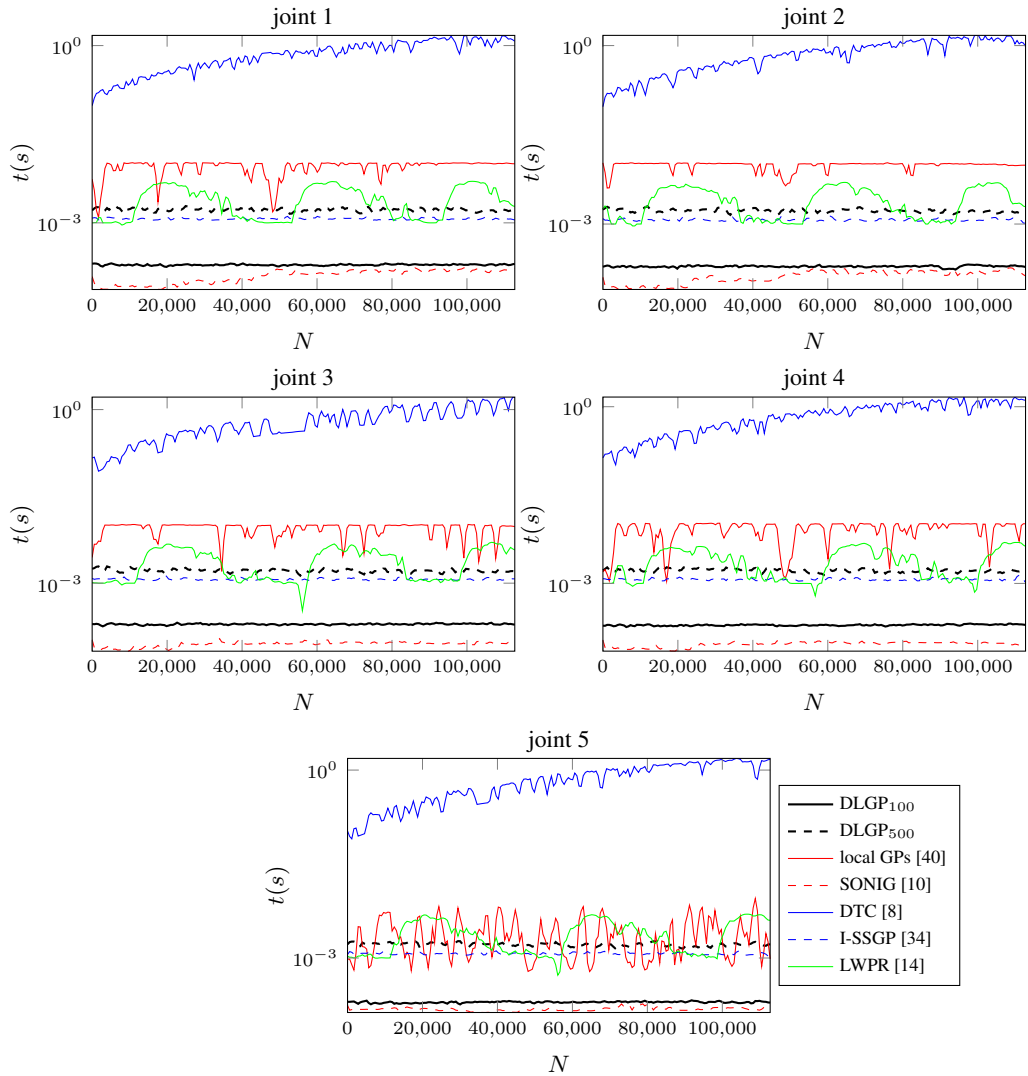

\end{document}